\pdfoutput=1

\documentclass[11pt]{article}
\usepackage{float}

\usepackage[a-1b]{pdfx}

\usepackage{framed}
\usepackage{mdwlist}
\usepackage{siunitx}
\usepackage{latexsym}
\usepackage{colortbl}
\usepackage{xcolor}
\usepackage{nicefrac}
\usepackage{booktabs}
\usepackage{fnpct}
\usepackage{amsfonts}
\usepackage[T1]{fontenc}
\usepackage{bold-extra}
\usepackage{amsmath}
\usepackage{amssymb}
\usepackage{bm}
\usepackage{graphicx}
\usepackage{mathtools}
\usepackage{microtype}
\usepackage{multirow}
\usepackage{multicol}
\usepackage{xpatch}
\usepackage{latexsym,comment}
\usepackage[normalem]{ulem}

\newcommand*{\missingreference}{{\Huge \colorbox{red}{?reference?}}}
\newcommand*{\missingcitation}{{\Huge \colorbox{red}{?citation?}}}

\makeatletter
\xpatchcmd{\@setref}{\bfseries}{\missingreference}{}{}
\def\@citex[#1]#2{\leavevmode
    \let\@citea\@empty
    \@cite{\@for\@citeb:=#2\do
        {\@citea\def\@citea{,\penalty\@m\ }%
            \edef\@citeb{\expandafter\@firstofone\@citeb\@empty}%
            \if@filesw\immediate\write\@auxout{\string\citation{\@citeb}}\fi
            \@ifundefined{b@\@citeb}{\hbox{\reset@font\missingcitation}%
                \G@refundefinedtrue
                \@latex@warning
                {Citation `\@citeb' on page \thepage \space undefined}}%
            {\@cite@ofmt{\csname b@\@citeb\endcsname}}}}{#1}}
\makeatother

\newcommand{\mm}[0]{\abr{llm}\xspace}

\newcommand{\dr}[0]{\textsc{DR}\xspace}
\newcommand{\ai}[0]{\abr{ai}\xspace}

\newcommand{\gem}[1]{\mbox{\textsc{gem}}}
\newcommand{\abr}[1]{\textsc{\uppercase{#1}}}

\newcommand{\hidetext}[1]{}
\newcommand{\ignore}[1]{}
\newif\ifcomment
\commenttrue

\ifcomment
    \newcommand{\pinaforecomment}[3]{\colorbox{#1}{\parbox{.8\linewidth}{#2: #3}}}

    \newcommand{\prtodo}[1]{\pinaforecomment{lightblue}{pr}{#1}}
    \newcommand{\prtodoi}[1]{\pinaforecomment{lightblue}{pr}{#1}}
\else
    \newcommand{\pinaforecomment}[3]{}
    \newcommand{\prtodo}[1]{}
    \newcommand{\prtodoi}[1]{}
\fi

\newcommand{\smallurl}[1]{ \begin{tiny}\url{#1}\end{tiny}}

\definecolor{lightblue}{HTML}{3cc7ea}
\definecolor{CUgold}{HTML}{CFB87C}
\definecolor{grey}{rgb}{0.95,0.95,0.95}
\definecolor{ceil}{rgb}{0.57, 0.63, 0.81}
\definecolor{UMDred}{HTML}{ed1c24}
\definecolor{UMDyellow}{HTML}{ffc20e}

\newcommand{\nlp}[0]{\abr{nlp}}

\usepackage[utf8]{inputenc}
\usepackage{pgfplots}
\usepackage{inconsolata}
\usepackage{booktabs}
\usepackage{multirow}
\usepackage{amsmath}
\usepackage{amsfonts}
\usepackage{pifont}
\usepackage{xspace}
\usepackage{graphicx}
\usepackage[percent]{overpic}
\usepackage{bm}
\usepackage[most]{tcolorbox}
\usepackage{csquotes}
\usepackage{anyfontsize}
\usepackage{comment}
\usepackage{float}
\usepackage{cuted}
\usepackage{tikz}
\usepackage{listings,multicol}
\usepackage{filecontents}
\usepackage{dsfont}
\DeclareUnicodeCharacter{2212}{−}
\usepgfplotslibrary{groupplots,dateplot}
\usetikzlibrary{patterns,shapes.arrows}
\pgfplotsset{compat=newest}

\usepackage{tikzscale}
\usepackage{amsmath}
\newcommand{\probP}{\text{I\kern-0.15em P}}

\newcommand{\model}{\textsc{MySQA}\xspace}
\newcommand{\modelFull}{\textsc{MyScholarQA}\xspace}
\newcommand{\scholarqa}{\textsc{ScholarQA}\xspace}
\newcommand{\scholarqaShort}{\textsc{SQA}\xspace}
\newcommand{\openscholarqa}{\textsc{OpenScholar}\xspace}
\newcommand{\storm}{\textsc{STORM}\xspace}

\newcommand{\metric}[1]{\hl{#1}}
\newcommand{\user}[1]{\textit{#1}}

\usepackage{soul}
\usepackage{etoolbox}

\newcommand{\authorSpacing}{0.5cm}

\usepackage{multirow, colortbl}

\usepackage{tabularx,booktabs}
\usepackage{makecell}

\usepackage[normalem]{ulem}

\usepackage{graphicx}

\usepackage[]{acl2023}

\usepackage{xspace}
\usepackage{multirow}
\usepackage{cancel}
\usepackage[utf8]{inputenc}
\usepackage{pgfplots}
\usepackage{dsfont}
\DeclareUnicodeCharacter{2212}{−}
\usepgfplotslibrary{groupplots,dateplot}
\usetikzlibrary{patterns,shapes.arrows}
\pgfplotsset{compat=newest}

\newcommand{\lessonsection}[2]{%
  \refstepcounter{subsection}%
  \subsection*{Lesson #1: #2}%
}

\definecolor{bggray}{rgb}{0.95, 0.95, 0.95}
\usepackage[%
    framemethod=tikz,
    skipbelow=\topskip,
    skipabove=\topskip
]{mdframed}
\mdfsetup{%
    leftmargin=0pt,
    rightmargin=0pt,
    backgroundcolor=bggray,
    middlelinecolor=black,
    roundcorner=3
}

\newtcolorbox[
  list inside=prompt,
  auto counter,
  number within=section
]{prompt}[1][]{%
  enhanced,
  float*=t,                 
  colbacktitle=black!60,
  fonttitle=\small,
  coltitle=white,
  fontupper=\footnotesize,
  boxsep=4pt,
  left=0pt, right=0pt, top=0pt, bottom=0pt,
  boxrule=1pt,
  width=\textwidth,          
  enlarge left by=0mm,
  enlarge right by=0mm,
  listing only,
  listing options={
    basicstyle=\ttfamily\footnotesize,
    breaklines=true,
    breakatwhitespace=true,
    language=json
  },
  #1,
}

\usepackage[%
    framemethod=tikz,
    skipbelow=\topskip,
    skipabove=\topskip
]{mdframed}
\mdfsetup{%
    leftmargin=0pt,
    rightmargin=0pt,
    backgroundcolor=white,
    middlelinecolor=black,
    roundcorner=3
}

\newtcolorbox[list inside=prompt,auto counter,number within=section]{summary}[1][]{
    colbacktitle=blue!10,
    fonttitle=\small,
    coltitle=black,
    fontupper=\footnotesize,
    boxsep=4pt,
    left=0pt,
    right=0pt,
    top=0pt,
    bottom=0pt,
    boxrule=1pt,
    width=\textwidth, 
    enlarge left by=0mm, 
    enlarge right by=0mm, 
    #1,
}

\usepackage[]{algpseudocode}
\usepackage[]{algorithm}
\usepackage{float}
\algtext*{EndFor}%
\algtext*{EndProcedure}%

\usepackage{booktabs}
\usepackage[normalem]{ulem}

\usepackage{times}
\usepackage{latexsym}
\usepackage{adjustbox}

\usepackage[T1]{fontenc}
\usepackage{amsmath}
\usepackage{amssymb}
\usepackage{booktabs}
\usepackage{multirow}
\usepackage{scalerel,xparse}
\usepackage[utf8]{inputenc}

\usepackage{cleveref}
\usepackage{microtype}
\usepackage{enumitem}
\crefformat{section}{\S#2#1#3}
\crefformat{subsection}{\S#2#1#3}
\crefformat{subsubsection}{\S#2#1#3}

\usepackage{soul} 
\usepackage{xcolor} 

\definecolor{lightgray}{RGB}{220,220,220}
\sethlcolor{lightgray}

\title{Language Models Don't Know What You Want:\\Evaluating Personalization in Deep Research Needs Real Users}

\author{
\textbf{Nishant Balepur}$^{1, 2, 3*}$\hspace{\authorSpacing}
\textbf{Malachi Hamada}$^{2}$ \hspace{\authorSpacing}
\textbf{Varsha Kishore}$^{2}$ \hspace{\authorSpacing}
\textbf{Sergey Feldman}$^{2}$ \\ \hspace{\authorSpacing}
\textbf{Amanpreet Singh}$^{2}$ \hspace{\authorSpacing} 
\textbf{Pao Siangliulue}$^{2}$ \hspace{\authorSpacing}
\textbf{Joseph Chee Chang}$^{2}$ \\ \hspace{\authorSpacing}
\textbf{Eunsol Choi}$^{3}$ \hspace{\authorSpacing}
\textbf{Jordan Boyd-Graber}$^{1}$ \hspace{\authorSpacing}
\textbf{Aakanksha Naik}$^{2}$ \\[0.5em]
$^{1}$University of Maryland \hspace{0.2cm}
$^{2}$Allen Institute for Artificial Intelligence \hspace{0.2cm}
$^{3}$New York University \\[0.25em]
\texttt{nbalepur@umd.edu} \hspace{0.5em} \texttt{aakankshan@allenai.org}\\[0.5em]
\raisebox{-0.75em}{\includegraphics[height=2em]{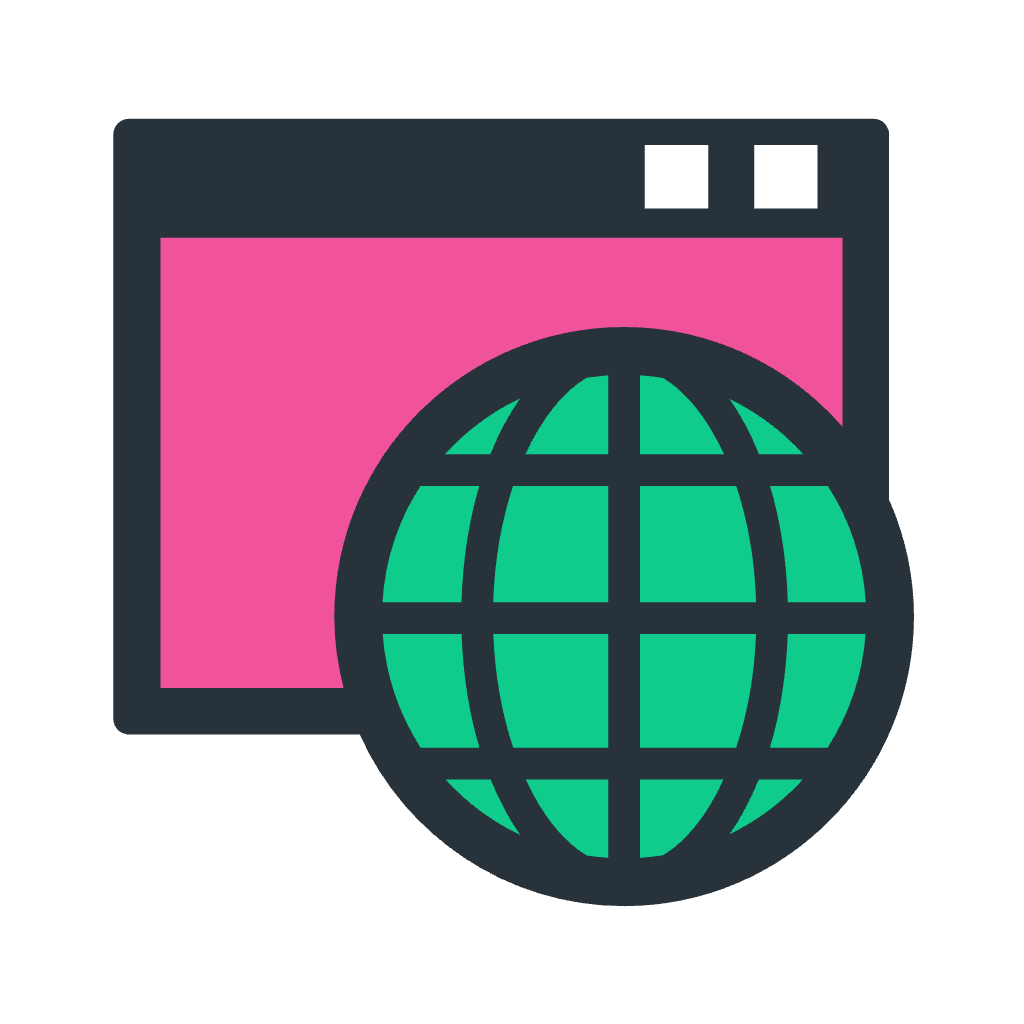}} \textbf{MyScholarQA:} \url{https://personalized-scholarqa.apps.allenai.org/}  
}

\begin{document}

\maketitle
\begingroup

\endgroup

\begin{abstract} {



Deep Research (\dr{}) systems help researchers cope with ballooning publishing counts.
Such tools synthesize scientific papers to answer research queries, but lack understanding of their users.
We address this with \modelFull{} (\model{}), a personalized \dr{} agent that: 
1)~infers a profile with a user's research interests; 
2) proposes personalized actions for a user's input query; 
and 3) writes a multi-section report for the query that follows user-approved actions.
We first test \model{} with \nlp{}'s standard protocol: we build a benchmark with synthetic~users and \mm{} judges, where \model{} beats baselines in citation metrics and personalized action-following.
However, we suspect this process does not cover all aspects of personalized \dr{} users value, so we interview~users~in an online version of \model{} to unmask~them.
We reveal nine nuanced errors~of personalized \dr{} undetectable by our \mm{} judges, and we study qualitative feedback to form lessons for future \dr{} design.
In all, we argue for a pillar of personalization that easy-to-use LLM judges can lead \nlp{} to overlook: real progress in personalization is only possible~with real users.\footnote{We release code and data at: \url{https://github.com/allenai/personalized-scholarqa-eval}} 
}
\end{abstract}

\begingroup
\renewcommand{\thefootnote}{\fnsymbol{footnote}}
\footnotetext[1]{Work primarily completed during internship at Ai2.}
\endgroup


\section{When Deep Research Gets to Know You} \label{section:introduction}


Scholars increasingly turn to \mm{}s to support their scientific research \cite{liao2024llms}, such as to learn new concepts \cite{august2023paper} or brainstorm ideas \cite{pu2025ideasynth}.
With publishing rates skyrocketing and literature becoming daunting to track~\cite{parolo2015attention}, a new use case of \mm{}s emerges: \textbf{Deep Research (\dr{})} tools that answer researchers' queries by retrieving, organizing, and synthesizing papers~into multi-section, attributed reports~\cite{Asai2024OpenScholarSS, huang2025deep}.

\dr{} has advanced in giving well-cited reports for queries~\cite{bragg2025astabench}, but few capture the individual needs of \textit{who} asks them, lacking \textbf{personalization}. 
By knowing a researcher's background, \dr{} could create more helpful reports: focusing on papers in certain domains, framing explanations in familiar terms, or showing how to use new ideas in users' ongoing work.
While decades of search engine research motivates personalization's benefits in \ai{} search tools~\cite{Teevan2005PersonalizingSV, Dou2007ALE}, work on personalized \dr{} remains sparse.


\begin{figure*}
    \centering

    \begin{overpic}[width=\linewidth]{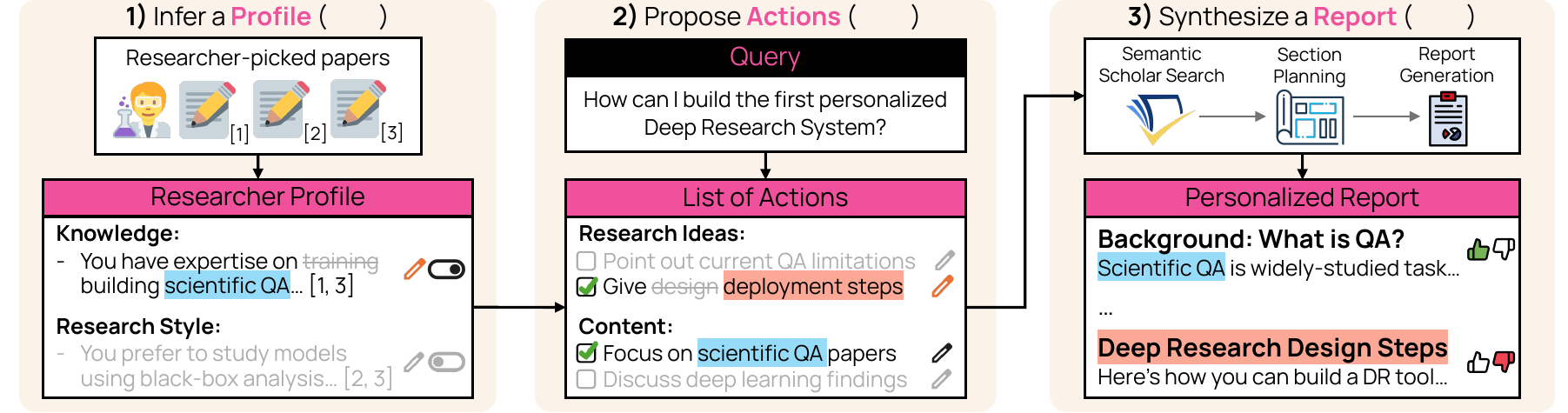}

    \put(20.8, 24.7){\scalebox{0.8}{\cref{subsection:mysqa_profile}}}
    \put(54.6, 24.7){\scalebox{0.8}{\cref{subsection:mysqa_plan}}}
    \put(90.13, 24.7){\scalebox{0.8}{\cref{subsection:mysqa_report}}}
    
    \end{overpic}
    \caption{\label{fig:model} Overview of \model{}: our three-step personalized Deep Research system. (1) Researchers upload papers from Semantic Scholar, from which an \mm{} infers a profile that captures their interests. (2) When the researcher asks a query, \model{} proposes a list of actions that could alter the report, tailored to the researcher's profile. (3) The system generates a report that answers the query and executes these actions through a multi-stage retrieval and \mm{} generation pipeline. To improve transparency and control, users can edit profiles, adjust actions, and view highlights in the report where \model{} personalizes content.  
}
\end{figure*}

We introduce \modelFull{} (\model{}): the first open-source personalized \dr{} tool that: 
1)~infers \emph{user profiles} via papers users pick to capture their interests (Fig~\ref{fig:model}, left);
2) suggests tool \emph{actions} tailored to the user's profile and query (Fig~\ref{fig:model}, center); and
3) writes a \textit{report} to answer the query and execute actions via a~multi-\mm{} system (Figure~\ref{fig:model}, right).
For transparency and control, our users can edit profiles in (1) so \model{} best captures them, adjust the actions that it executes in (2), and~review highlights where it personalizes content in (3).



A formative study showed \model{}'s promise but that the system was imperfect (\cref{subsection:formative}), leading us to ask: how should we evaluate \model{} to reveal common failures and track improvements?
In surveying recent \nlp{} work on personalization, offline benchmarks dominate \cite{voorhees1999trec}: for 31 ACL'25 works on personalization, all evaluate offline (18 with synthetic user datasets and 17 with \mm{} judges), but \textbf{only two run user studies with real end users }(Appendix~\ref{appendix:persona_survey}).
However, excelling in offline evaluation does not ensure the system is helpful~\cite{mozannar2025the}, as offline metrics can neglect what users value~\cite{NarayananVenkit2025SearchEI}.

To test what offline personalization evaluations miss, we first build a synthetic dataset pairing \dr{} queries with paper sets simulating users (\cref{subsection:dataset}), and 16 offline metrics (\cref{subsection:metrics}). 
Here, \model{}~excels: profiles accurately cite papers for inferred user interests, personalized actions more closely match user profiles versus generic ones, and our reports have higher quality and adherence to actions versus open-source and commercial \dr{} systems (\cref{subsection:offline_results}). 
To contrast this with online evaluations, we draw on human-computer interaction and use \model{} as~a probe~\cite{Hutchinson2003TechnologyPI} to unveil real users' needs in personalized \dr{}. 
In 90-minute interviews, 21 \dr{} users build profiles, pick query actions, and rate reports via a \model{}-backed interface~(\cref{subsection:interview}).
%






Participants perceive \model{} as helpful---73\% of its profiles, actions, and reports are satisfactory---but uncover nine personalization flaws our offline metrics miss: e.g., profiles overstating user expertise and tailored actions drifting from query intent (\cref{subsection:interview_metrics}).
To test if we could have flagged these~flaws offline, we use \mm{} judges to predict user satisfaction ratings (\cref{subsection:simulation_results}) with validation sets derived from interviews: they never beat majority class baselines, failing to capture user needs. 
Beyond finding metrics to advance, user feedback also yields lessons to inform personalized \dr{} design: make control~easy, let users digest personalization, use content beyond papers, and evaluate via mixed study designs~(\cref{section:insights}).


We aim to reinforce a pillar of personalization research: developing personalized tools needs feedback from real users.
Synthetic data is an alluring crutch and \mm{} judges seem reliable~\cite{Jiang2025KnowMR}, but these evaluations can miss what users actually value.
Thus, we urge their adoption as necessary but insufficient checks and instead advocate for user-centered evaluations to inform the design of personalized NLP systems. 
%

\newpage
\noindent Our contributions are:

\begin{enumerate}[leftmargin=*, itemsep=0pt]
    \item \modelFull{}, the first personalized Deep Research (\dr{}) system with an online demo.\footnote{https://personalized-scholarqa.apps.allenai.org/}

    \item A synthetic benchmark and \mm{} judge metrics as offline evaluations of personalized \dr{}.

    \item Discovery that \mm{} judges fail to predict nine failure modes of personalized \dr{}, advocating for user-centered personalization evaluations.

    \item The first formative and usability studies of personalized \dr{} with 26 active \dr{} users to inform future work on personalization design choices.
\end{enumerate}

\section{\model: Personalized Deep Research} \label{section:prototype}

Drawing on \citet{Brusilovsky1996MethodsAT}'s adaptive hypermedia, personalized Deep Research tools build a persistent user model of the researcher, then apply this model 
to adapt reports for user queries.
Our design goals are:~1) build a user model~to capture research interests; and
2) let users interpret and control how the system personalizes~\cite{Liu2024FilteringDR}, helping us learn the best ways to adapt outputs.

We realize this in \modelFull{}~(\model{}), a \dr{} tool (Fig~\ref{fig:model}) that infers profiles of users'~interests via papers (\cref{subsection:mysqa_profile}), plans actions for user~queries tailored to their profile (\cref{subsection:mysqa_plan}), and executes actions to adapt reports (\cref{subsection:mysqa_report}).
We now~describe~each step and evaluate its design in a formative study~(\cref{subsection:formative}).

\subsection{Inferring Researcher \underline{Profiles}} \label{subsection:mysqa_profile}

\model{} first infers a profile $\mathcal{P}$ from a user's papers $\mathcal{D}$ (Fig~\ref{fig:profile})---forming a persistent user model \cite{Brusilovsky1996MethodsAT}; we use papers, since \citet{Lin2024PaperCA} show they capture research interests.
Profiles can be biographies~\cite{Gao2024EndtoendTF} or keywords~\cite{10.1145/3539618.3591677}, but we use sentence-level inferences 
$\mathcal{P} = \{\mathcal{I}_1, \dots, \mathcal{I}_{n_1}\}$ about the user (e.g. ``\textit{Your papers argue evaluations should move beyond metrics to online studies.}''), similar to prior research in computer agents~and writing assistance \cite{Shaikh2025CreatingGU, Garbacea2025HyPerAlignIP}.


$\mathcal{P}$ has $n_1$ inferences evenly split over~five~aspects inspired by \citet{Tang2024StepBackPD}: knowledge (what they know), research style (how they do research), writing style (how they write), audience (whom they impact), and positions (what~they believe). 
Users find this structure organized~(\cref{subsection:formative}).  
$\mathcal{P}$ transparently cites snippets from the user's papers $\mathcal{D}$ for each inference $\mathcal{I}$ with an explanation.
We prompt \mm{}s~to create $\mathcal{P}$ from $\mathcal{D}$ (Appendix~\ref{appendix:prompt_outputs}).
Users can edit/disable any $\mathcal{I}$ to form a better profile $\mathcal{P}^*$.

\subsection{Proposing \underline{Actions} to Take} \label{subsection:mysqa_plan}

\begin{figure}[t]
    \centering
\includegraphics[width=\linewidth]{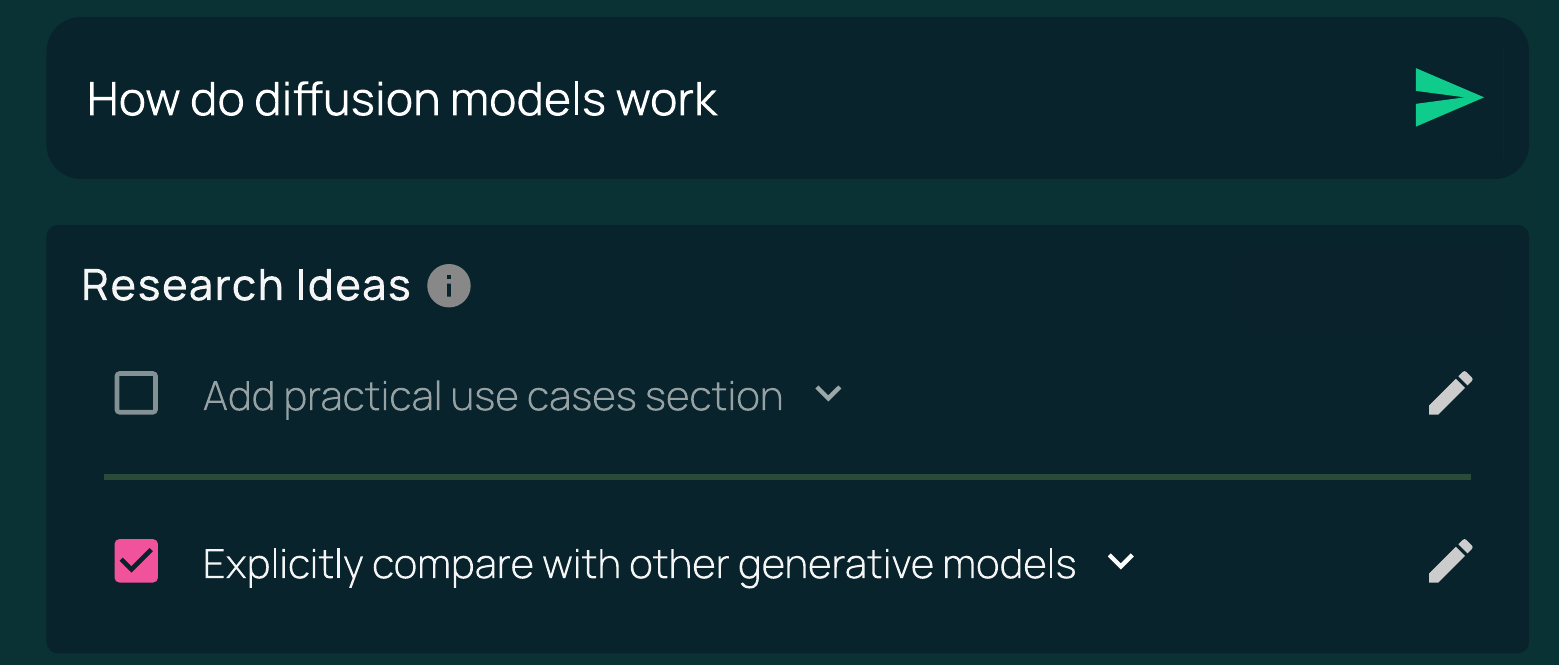}
    \caption{\label{fig:plan} Before answering a user's query, \model{} proposes actions that change how the report could be created, which users can adjust to customize their~report.}
\end{figure}

Equipped with a profile $\mathcal{P}^*$, users can ask a query~$q$ to get a report $\mathcal{R}$ adapted to $\mathcal{P}^*$~\cite{Brusilovsky1996MethodsAT}.  
Directly producing $\mathcal{R}$ limits customization;
instead, we draw on query clarification \cite{zhang2025modeling} and first return a list of actions $\mathcal{A} = \{a_1, \dots, a_{n_2}\}$ (Fig~\ref{fig:plan}) that \model{} could take when answering~$q$, such as ``\textit{find trivia QA papers}'' or ``\textit{add section on metrics}'' \citep{srikanth2026discotracerepresentingcomparinganswering}.
Exposing $\mathcal{A}$ lets users~steer execution and tell us how to adapt~$\mathcal{R}$---validated in a formative study (\cref{subsection:formative}).


$\mathcal{A}$ has $n_2$ actions split over four categories, based on how they will adapt $\mathcal{R}$: content (what $\mathcal{R}$ covers), style (how $\mathcal{R}$ explains), specificity (how $\mathcal{R}$ interprets $q$), and research ideas ($\mathcal{R}$'s proposed ideas the user can incorporate).
%
Actions can be \textit{personalized} $\mathcal{A}_{\text{person}}$ (conditioned on $q$ and $\mathcal{P}^*)$ or \textit{generic} $\mathcal{A}_{\text{gen}}$ (conditioned on $q$); $\mathcal{A}_{\text{gen}}$ ensures $a$ is~useful when $\mathcal{P}^*$ is unlike $q$ (e.g. if users ask about a new field).  

We prompt \mm{}s to create $\mathcal{A}_{\text{gen}}$ and $\mathcal{A}_{\text{person}}$ separately, merging them to form $\mathcal{A}$ (Appendix~\ref{appendix:prompt_outputs}); we add ways the \mm{} can adapt to $\mathcal{P}^*$ for $\mathcal{A}_{\text{person}}$'s prompt (e.g., skip basic terms for experts) but no strict rules, as we want to learn how to personalize reports 
\text{from} users (\cref{section:online}).
Users can edit/disable any action $a \in \mathcal{A}$ to form custom actions $\mathcal{A}^* \subseteq \mathcal{A}$.

\subsection{Synthesizing a Personalized \underline{Report}} \label{subsection:mysqa_report}

After the user submits actions $\mathcal{A}^*$,~\model{} writes a multi-section report $\mathcal{R} = \{\mathcal{S}_1,\dots , \mathcal{S}_{n_3}\}$~to answer query~$q$ while executing each $a \in \mathcal{A}^*$, personalizing the report via actions which were shaped by the profile.
We build on \scholarqa{} \cite[\scholarqaShort]{singh-etal-2025-ai2}---a \dr{} system with high-quality reports.
\scholarqaShort chains \mm{}s to retrieve papers from Semantic Scholar,\footnote{ https://www.semanticscholar.org/product/api} cluster them into sections, and iteratively generate well-cited sections to form $R$.

\begin{figure}[t]
    \centering
    \includegraphics[width=\linewidth]{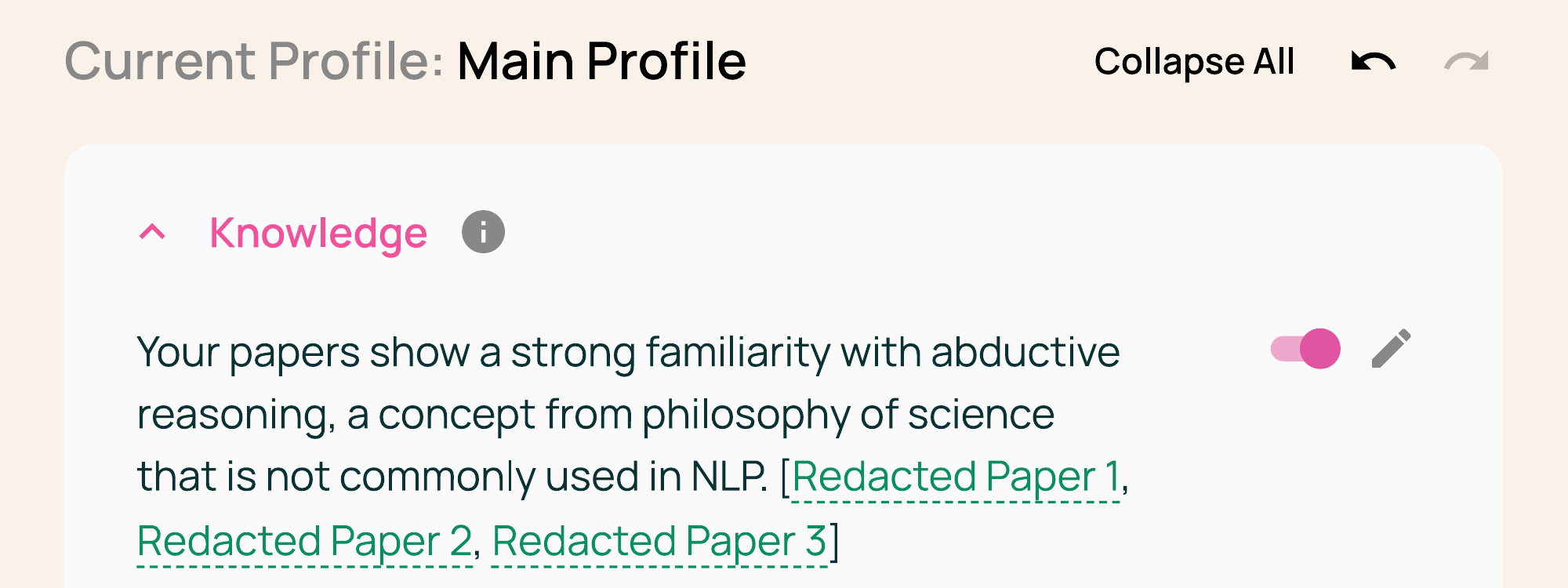}
    \caption{\label{fig:profile} After a user selects research papers,~\model{} infers an editable profile with inferences about the user, capturing their interests for personalizing reports.}
\end{figure}

\begin{figure}[t]
    \centering
\includegraphics[width=\linewidth]{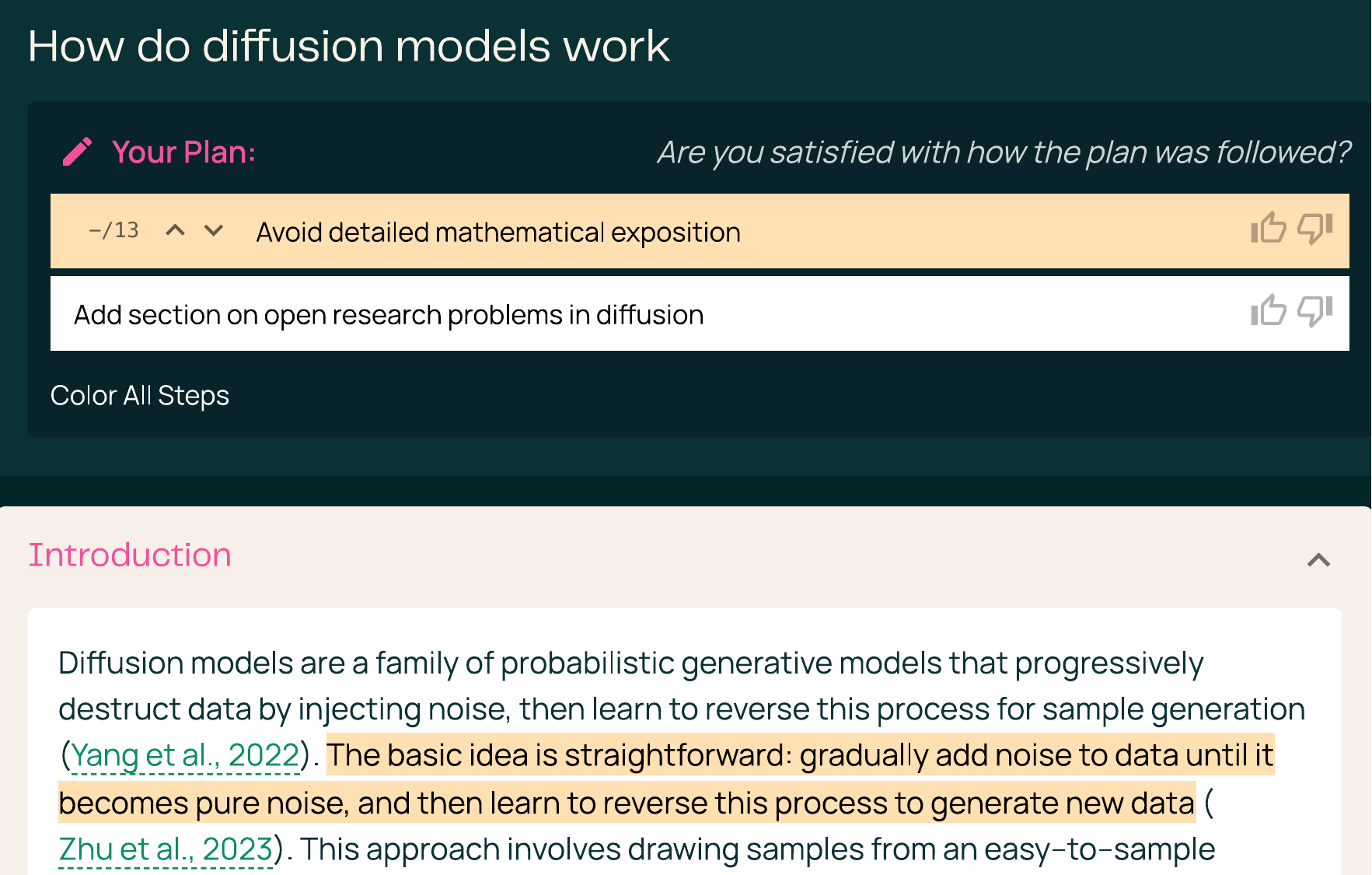}
    \caption{\label{fig:highlight} Reports in \model{} highlight personalized content, helping users navigate each action they select.}
\end{figure}


To equip \model{} to use $\mathcal{A}^*$, we tweak prompts in \scholarqaShort{}'s execution to also use $\mathcal{A}^*$ as input.
For example, the first step is a prompt converting $q$ to search terms for Semantic Scholar (i.e. ``convert $q$ to search terms''); we minimally modify it with instructions on how to also use $\mathcal{A}$ to create~search terms (i.e. ``convert $q$ to search terms while following these actions: $\mathcal{A}^*$'').
Most notably, we change: 1) the prompt for search terms---generating multiple search terms based on $q$ and $\mathcal{A}^*$, while \scholarqaShort{} originally generates just one; and 2) the report generation prompt---instructing the \mm{} to execute all $a \in \mathcal{A}^*$ and highlight parts of the text that relate to any action---one color per action (Fig~\ref{fig:highlight}).
(1)~tailors retrieval to $\mathcal{A}^*$, while (2) helps users see where \model{} personalizes $\mathcal{R}$ \cite[\cref{subsection:sensemaking}]{Kim2025IntentFlowIS}.  
Claude-4 Sonnet backs \model{},\footnote{https://www.anthropic.com/news/claude-4 \label{footnote:claude}} matching \scholarqaShort{}. 


\subsection{A Formative Study with \model{}} \label{subsection:formative}

To ensure our method matches users' expectations before investing in large-scale studies, we run a~formative study~\cite{Nielsen1993UsabilityE} via an early version of \model{} as a technology probe~\cite{Hutchinson2003TechnologyPI}.\footnote{Here, we use Gemini-2.5  Pro for \model{} profiles/plans.}
Five \dr{} users/CS students~bring papers of~their interest and two queries they asked \dr{} before.
We first discuss their \dr{} usage/personalization needs (\char`\~$15$ min).
Then they use \model{}---creating and editing profiles/actions to get personalized reports (\$35/hr; \char`\~$45$ min); later over email, they rate reports for their queries (detailed in~\ref{appendix:formative}).

Participants found using papers to infer personalized profiles/actions intuitive for personalized \dr{}; out of all potential user context, P1 noted ``\user{papers matter most}''. 
They also found our profile/action categories useful and comprehensive (\cref{subsection:mysqa_profile}, \cref{subsection:mysqa_plan}).
Many desired transparency, wanting it ``\user{as transparent as possible}'' (P4) and to ``\user{know how they made inferences about me}'' (P3).

Participants then rated their profiles, actions,~and reports from \model{}.
Profiles had surprising nuance and detail---``\user{It captures exactly what I have in mind but haven’t expressed}'' (P2)---but were~incomplete: ``\user{a really good starting point to review and refine}'' (P5).
Personalized actions were promising; some felt like ``\user{speaking with a colleague~who knew my work}'' (P2).
Still, participants liked generic/personalized ones similarly (\char`\~$60\%$), so we do not yet know when personalized \dr{} helps. 
Highlighting where reports tailored to actions made it ``\user{much easier to see custom parts at a glance}'' (P3) but such text could be ``\user{a bit general}''~(P5).
Lastly, they compared generic \scholarqaShort{} reports to personalized \model{} reports for their queries;~all favored the~latter, showing personalized \dr{}'s~utility.
In all, our study confirms \model{} is a solid~basis~for finding more aspects of personalized~\dr{}~users~value.

Post-study, we use participant feedback to refine \model{}.
We tweak prompts to try and fix common errors (e.g. ``make profiles specific'', \cref{subsection:mysqa_profile})~and host a UI for further online use, with a:
1) profile page to find~papers~and toggle/edit \mm{} inferences (Fig~\ref{fig:profile});  
2) home page to ask queries and toggle/edit \mm{} actions (Fig~\ref{fig:plan}); and  
3) report~page with highlights showing where each action was executed (Fig~\ref{fig:highlight}).
After adding these updates, we continue to larger-scale offline (\cref{section:offline}) and online (\cref{section:online}) evaluations.

\section{Offline Evaluation} \label{section:offline}

Our formative study showed \model{}’s promise (\cref{subsection:formative}) but did not thoroughly evaluate \dr{} output quality.
As~a next step, we adopt \nlp{} practices and test \model{} offline via simulated users (\cref{subsection:dataset}) and \mm{} judges (\cref{subsection:metrics}).
These analyses are cheap and popular \cite{Jiang2025KnowMR}, but may not capture users' personalization needs \cite{NarayananVenkit2025SearchEI}, so we use them as necessary but insufficient~checks to inform online, user-centered~evaluations~(\cref{subsection:offline_results}).

\subsection{Dataset Collection} \label{subsection:dataset}

No personalized \dr{} datasets exist of user~queries~$q$ and papers $\mathcal{D}$, so~we~make a synthetic one. We collect $q$ from ScholarQA-CS2 \cite{bragg2025astabench}---a research agent benchmark with 200 \dr{} $q$ (100 dev/100 test). We attach synthetic users to each $q$ with low, medium, and high expertise for $q$. We simulate synthetic users based on papers in CS-PaperSum~\cite{Liu2025CSPaperSumAL}: CS conference papers grouped by first author (with three or more papers).
We compute~expertise~via cosine similarity\footnote{$[0,0.2]$ (low), $(0.2,0.35]$ (medium), and $(0.35,1.0]$ (high); dropping ranges if no $\mathcal{D}$ matches} between GRIT-LM embeddings~\cite{Muennighoff2024GenerativeRI} of $q$ and user papers $\mathcal{D}$.
%
We get 281 and 291 $(q,\mathcal{D})$ pairs for dev and test splits, respectively.

\subsection{Metric Implementation} \label{subsection:metrics}

On our dataset (\cref{subsection:dataset}), we now study \model{}~profiles, actions, and reports with objective metrics~for clearly undesired errors, deferring a study of subjective~metrics (\cref{subsection:simulation_results}).
Gemini-2.5 Flash \cite{Comanici2025Gemini2P} is our \mm{} judge \cite{Zheng2023JudgingLW} for metrics (human agreement in Appendix~\ref{appendix:metrics}). 

For each \textbf{profile} inference $\mathcal{I}$, we evaluate: 1) \metric{category accuracy}, if $\mathcal{I}$ is in the correct category (e.g. knowledge); 
2) \metric{inference accuracy}, if $\mathcal{I}$ contradicts cited papers, like summary faithfulness~\cite{Kryscinski2019EvaluatingTF}; 
3) \metric{citation relevance},~the proportion of cited papers that support any part of $\mathcal{I}$ to penalize overciting; and 
4) \metric{specificity}, a 1--5 score for how much $\mathcal{I}$ differs~among researchers. As confounders, we show mean cited paper count and words in $\mathcal{I}$.  


We compare personalized and generic \textbf{actions} with: 1) \metric{win rate}, how~often~a~judge picks~$\mathcal{A}_{\text{person}}$ vs~$\mathcal{A}_{\text{gen}}$ given $\mathcal{P}$, a consistency check from \citet{balepur-etal-2025-whose}; 2) \metric{coherence},~how~often each $a \in \mathcal{A}$ does not contradict~$q$ (e.g. $q=$``what is QA'', $a=$``focus on NLI'' is a conflict); and 
3) \metric{uniqueness}, the~proportion of $a \in \mathcal{A}$ that a system without $\mathcal{A}$ would~\underline{not} already follow.
We check (3) via how often \scholarqaShort{} prompted with just $q$ executes $a \in \mathcal{A}$ in $\mathcal{R}$, via the ``action adherence'' report metric below.


In \textbf{reports}, we assess how well $\mathcal{R}$~uses~$q$~and~$\mathcal{A}$. 
We use four report quality metrics from~ScholarQA-CS2 \cite{bragg2025astabench}: 1) \metric{answer coverage}, how many elements a correct answer for $q$ must include are covered in $\mathcal{R}$; 2) \metric{answer precision}, how directly $\mathcal{R}$ answers $q$; 3) \metric{citation precision}, $\mathcal{R}$'s citation accuracy; and 4) \metric{citation recall}, how often $\mathcal{R}$'s~claims are cited. 
We add \metric{action adherence}---how often~$\mathcal{R}$ follows actions in $\mathcal{A}$ at any point \cite{Qin2024InFoBenchEI}.

\subsection{Baselines} \label{subsection:baseline}


For profiles/actions, we assess \mm{}s to pick one for \model{}.  
\textbf{Profiles} infer over long contexts \cite{Kuratov2024BABILongTT} only once, so we evaluate reasoning \mm{}s: Claude-4~Sonnet+think, o3 \footnote{https://openai.com/index/introducing-o3-and-o4-mini/}, Gemini-2.5 Pro \cite{Comanici2025Gemini2P}, and DeepSeek-r1 \cite{DeepSeekAI2025DeepSeekR1IR}.
\textbf{Actions} are per-query, so we test fast \mm{}s: Gemini-2.5 Flash, GPT-4.1, Claude-4 Sonnet, and DeepSeek-V3 \cite{DeepSeekAI2024DeepSeekV3TR}.
In \textbf{reports}, we compare~\model{} to open-source/commercial \dr{} that take concatenations of $\langle q \cdot \mathcal{A}\rangle$ as prompts: \scholarqa{} \cite{singh-etal-2025-ai2}, \openscholarqa{} \cite{Asai2024OpenScholarSS}, 
\storm{} \cite{Shao2024AssistingIW}, Perplexity's~Sonar\footnote{https://sonar.perplexity.ai/} and OpenAI o3 \dr{}\footnote{https://openai.com/index/introducing-deep-research/}
(details in~Appendix~\ref{appendix:setup}). 

\begin{table}[t]
\small
\centering
\setlength{\tabcolsep}{2pt}
\begin{tabular}{@{}lcccccc@{}}
\toprule
\textbf{Model} & Inf. Acc & Cit. Rel. & Cat. Acc. & Spec. & \# Cite & \# Words \\ \midrule
G-Pro     & \textbf{97.1}           & \textbf{97.4}            & 99.4           & 3.73         & \textbf{2.45}     & 23.2    \\
Sonnet         & 92.5           & 97.4            & 99.1           & 4.12         & 1.91      & 21.6    \\
OAI-o3      & 88.6           & 91.8            & \textbf{99.8}           & \textbf{4.20}         &  2.03     & 18.9    \\
DS-r1    & 77.8           & 80.7            & 97.2           & 3.56         & 1.89      & 9.9    \\ \bottomrule
\end{tabular}
\caption{Profile ($n_1=25$) inference accuracy, relevance, and specificity across \mm{}s. Highest scores are \textbf{bold}. Reasoning \mm{}s infer accurate, relevant user profiles. \label{table:profile_offline}}
\end{table}
\begin{table}[t]
\small
\centering
\setlength{\tabcolsep}{2pt}
\begin{tabular}{@{}lccccc@{}}
\toprule
\textbf{Model} & W.R. & $p_{\text{gen}}$ Coh. & $p_{\text{person}}$ Coh. & $p_{\text{gen}}$ Uniq. & $p_{\text{person}}$ Uniq. \\ \midrule
G-Flash        & 91.3 & 93.2                  & 84.0                     & 42.9                   & 60.7                   \\
Sonnet         & 93.5 & 91.8                  & 82.0                     & \textbf{50.4}                   & \textbf{68.2}                   \\
GPT-4.1        & \textbf{94.7} & \textbf{93.5}                 & \textbf{84.1}                     & 49.1                   & 66.3                   \\
DS-V3          & 94.3 & 89.2                  & 72.3                     & 54.5                   & 72.2                   \\ \bottomrule
\end{tabular}
\caption{Personalized/generic action ($n_2=16$) win rate, coherence, and uniqueness. The highest score is \textbf{bold}. \mm{} judges that condition on synthetic users' papers are more likely to select personalized actions, indicating that they can personalize actions based on user profiles. \label{table:plan_offline}}
\end{table}
\begin{table}[t]
\small
\centering
\setlength{\tabcolsep}{2.5pt}
\begin{tabular}{@{}lccccc@{}}
\toprule
\textbf{Model}   & An. Cov.     & An. Prec.    & Cit. Prec.    & Cit. Rec. & $\mathcal{A}$ Adh.    \\ \midrule
\model{}         & \textbf{91.4} & 89.9          & \textbf{91.8} & \textbf{81.4} & {\text{83.2}}    \\
\textsc{SQA}        & 88.9          & 89.1          & {\text{90.5}}    & {\text{76.9}}    & 81.3          \\
\textsc{OpenSc.}    & 77.2          & \textbf{97.4} & 82.5          & 60.4          & 82.5          \\
\storm{}         & 72.0          & {\text{92.2}}    & 73.3          & 64.7          & 74.4          \\
Sonar \dr{} & 81.0          & 82.9          & 64.3          & 46.3          & 75.0          \\
o3 \dr{}        & {\text{89.1}}    & 90.2          & 79.2          & 56.7          & \textbf{93.8} \\ \bottomrule
\end{tabular}
\caption{\dr{} report quality and action adherence scores for query $q$ and eight actions in $\mathcal{A}_{\text{gen}} \cup \mathcal{A}_{\text{person}}$. The best score is \textbf{bold}. \model{} surpasses every \dr{} baseline in three out of five metrics.
We only evaluate o3 \dr{} using ten examples due to latency and cost limitations. \label{table:report_offline}}
\end{table}

\subsection{Offline Results} \label{subsection:offline_results}

With our dataset (\cref{subsection:dataset}), metrics (\cref{subsection:metrics}), and baselines (\cref{subsection:baseline}) set, we now test each step in \model{}.
In \textbf{profiles}, all models but DS-r1 create accurately cited inferences (Table~\ref{table:profile_offline}) and 
all~accurately categorize inferences.
More citations reduces specificity---a common personalization/generalization tradeoff \cite{Han2022SplitGPAB}.
For \model{}'s profiles, we use Gemini-Pro, scoring the highest in three metrics. 

All models give personalized \textbf{actions} with higher win rates and uniqueness over generic ones, indicating such actions are tailored to profiles (Table~\ref{table:plan_offline}).
Yet, personalized ones conflict queries more; Appendix~\ref{appendix:tension} shows tension between tailored actions and answering queries, often for papers~with low query similarity (\cref{subsection:dataset}), so~personalization may not always help \cite{Zhang2013ToPO}.  
Actions generally had modest adoption rates in our formative (\cref{subsection:formative}), so we rotate all 4 models in \model{} for diversity.

In \textbf{reports}, \model{} has the best or second-best scores 4/5 times---the most of any \dr{} system (Table~\ref{table:report_offline}). It also always beats its base system \scholarqaShort{}: our modifications (\cref{subsection:mysqa_report}) improve its report quality.

Having shown \model{}'s output quality offline, we can now deploy the tool online to find flaws and insights our simulation-based data may miss~(\cref{section:online}).

\begin{table*}[]
\small
\centering
\setlength{\tabcolsep}{4pt}
\begin{tabular}{@{}cclc@{}}
\toprule
\multicolumn{1}{l}{\textbf{Output}} & \textbf{Aspect}    & \textbf{Description}                                                                                      & \textbf{Freq.} \\ \midrule
\multirow{4}{*}{Profile}       & DOMAIN     & Uses terms, definitions, or details that do not capture the user's domain of research.      & 27.6\%      \\
                               & OVERCLAIM  & Claims something applies to the user broadly, but only applies to some/parts of papers.    & 17.9\%      \\
                               & CONVENTION & Infers a generic convention of the user's field (e.g. "You enumerate contributions").     & 12.8\%      \\
                               & CONTRAST   & Has a contrast that misrepresents the user (e.g. "You are X, not Y", but the user is Y). & 12.2\%      \\ 
                               \midrule
\multirow{2}{*}{Action}          & NARROW     & The action is too specific and would overly constrain the information coverage.           & 43.8\%      \\
                               & OFFTOPIC   & The action deviates too far from the query, distracting from the user's goal/intent.         & 23.6\%      \\
                               \midrule
\multirow{3}{*}{Report}        & UNINFORM  & The content is too vague/high-level to be informative; the user wanted more details.       & 38.0\%      \\
                               & PRESENT    & The user wanted the content presented in a different style/format (e.g. bullet points).    & 25.3\%      \\
                               & IGNORE    & One or more implicit/explicit requirements in the action was ignored.     & 22.8\%      \\ \bottomrule
\end{tabular}
\caption{The nine most common personalization errors in \model{}'s outputs discovered from our interviews, missed in our offline evaluations. All twenty metrics we discover are in Table~\ref{table:codes_full}. \label{table:codes_selected}}
\end{table*}

\section{Moving \model{} Offline to Online} \label{section:online}

\model{} excels on synthetic datasets (\cref{section:offline}),~but this may not mirror real user needs \cite{Saxon2024BenchmarksAM} and even 10+ metrics might not cover all aspects of personalized \dr{} users value~\cite{Venkit2025DeepTRACEAD}.  


We thus run a user study (\cref{subsection:interview}) to answer open questions:
\textbf{RQ1}---What personalization errors do offline metrics neglect?~(\cref{subsection:interview_metrics}) and can off-the-shelf \mm{} judges evaluate them?~(\cref{subsection:simulation_results}). \textbf{RQ2}---What lessons can guide future personalized \dr{}?~(\cref{section:insights}) 



\subsection{Interviewing Active Deep Research Users} \label{subsection:interview}
  
We interview\footnote{Our internal review board approved our study (\cref{section:ethics}).} 21 active \dr{} users (CS researchers on Upwork, \$30-40/hr) for 90 minutes who show \dr{} familiarity in a pilot survey; 19 use OpenAI \dr{} (Appendix~\ref{appendix:survey}).  
Each~bring:~1) five papers~of interest;\footnote{Papers they have written, wish they had written, inspire them, or are relevant to a current project.} and 2) three queries to ask \dr{}.
While using \model{}, they screen-record and ``think out loud''~\cite{Danks1984ProtocolAV}.  
We transcribe recordings for open thematic analysis \cite{Boyatzis1998TransformingQI}.


In \model{}'s UI (\cref{subsection:formative}), participants review their profiles (\char`\~$45$ min.) and disable/edit poor inferences.  
Next, they review every query's proposed actions (\char`\~$5$ min.), picking or editing which ones \model{} should follow.
They submit actions to tailor reports, rating its quality and ability to follow actions while verbalizing reasoning for their ratings~(\char`\~$15$ min.).

\begin{figure}[t]
    \centering
    \includegraphics[width=\linewidth,trim=0 10 0 10,clip]{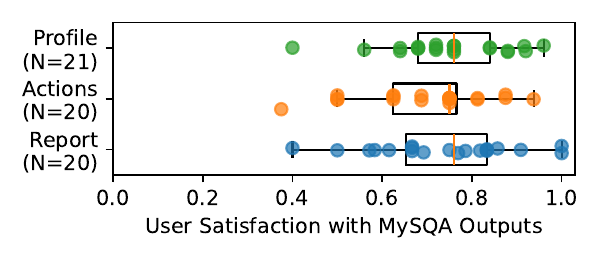}
    \caption{User satisfaction on \model{} profiles, actions, and reports. Users are satisfied with $73\%$ of them.\footnotemark \label{fig:satisfaction}}
\end{figure}
\footnotetext{U10 spent the interview just viewing the~profile ($n=20$).}

\begin{figure*}[t]
    \centering
\includegraphics[width=\linewidth, trim=0 0 0 10, clip]{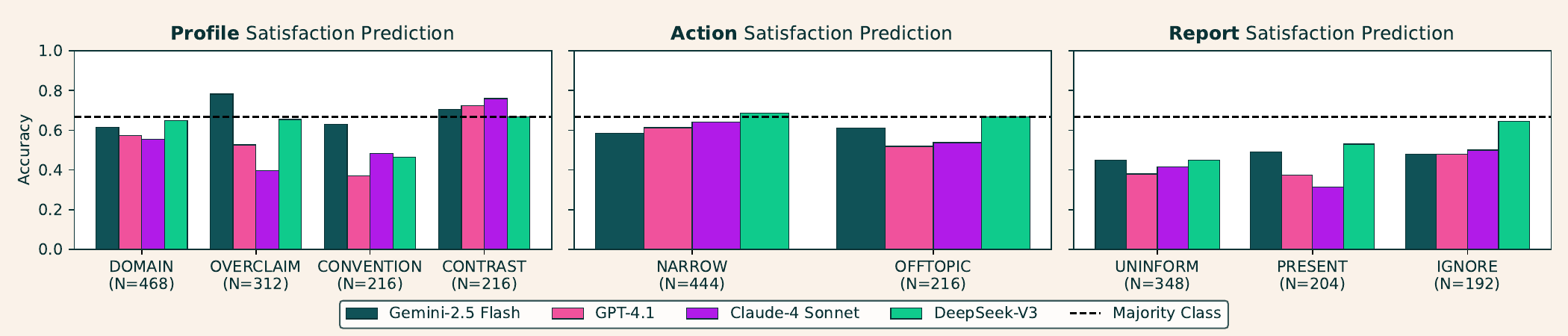}
    \caption{\label{fig:simulation} \mm{} judge accuracy for predicting users' satisfaction of personalized profiles, actions, and reports.
    No model ever beats a majority class baseline \cite[$\alpha=0.05$ Binomial test with Bonferroni correction]{dror-etal-2018-hitchhikers}.}
    \vspace{-1ex}
\end{figure*}

\subsection{RQ1: What our Offline Evaluation Missed} \label{subsection:interview_metrics}

Participants liked $73\%$ of profiles, actions, and action executions in reports (Fig~\ref{fig:satisfaction}), so \model{} was useful.
We now analyze the~remaining $27\%$ to uncover issues, guiding~future evaluation: an author reviewed all 1044~judgments and qualitatively labeled where participants were dissatisfied;~another verified themes on $60$ rows~(with 90\%~agreement).

We find nine common issues (Table~\ref{table:codes_selected}).
Most~often, profiles mistake technical terms (DOMAIN), actions are restrictive (NARROW), and reports~lack detail (UNINFORM). 
Many are open \nlp{}~problems: factuality~\cite[DOMAIN]{Wang2024FactualityOL}, style~\cite[PRESENT]{jin2022deep}, or expertise calibration~\cite[OVERCLAIM, CONVENTION]{Joshi2025ELIWhyET}.  
Some issues in the full list (Table~\ref{table:codes_full}) seem untraceable offline---e.g., accurate but unimportant profile inferences (UNIMPORT) and rejected actions due to mistrust in \model{}'s ability (TRUST)---only detectable with online feedback.

Crucially, every issue is invisible to our offline metrics (\cref{subsection:offline_results})---despite covering 10+ reasonable aspects of personalization---so synthetic datasets can overlook what users value in personalization.

\subsection{LLMs Don't Know What \dr{} Users Want} \label{subsection:simulation_results}

Having found personalized \dr{} aspects we~missed offline (\cref{subsection:interview_metrics}), we now see if we \textit{could~have} 
evaluated them via \mm{} judges.
We answer this~with a classifier:
given an aspect, were users satisfied by profiles, actions, and action executions in reports for that aspect. 
For each aspect, we label~outputs users dislike as $l=1$ paired with two negatives ($l=0$): a random liked output and a ``hard'' negative (the most stylistically-similar liked output)---both from the user.  
\mm{}s access the same context (e.g. papers, actions, highlights) as users, six~few-shot examples, and definitions of each aspect\footnote{e.g., for DOMAIN, we prompt ``Would the user be satisfied with the technical terminology in this profile inference?''} to predict $\hat{l}$: would the user like or dislike the output?  

Our four \mm{} judges predict~user satisfaction no better than majority-class baselines (Figure~\ref{fig:simulation}), so strong, off-the-shelf judges~struggle to~capture these issues (\cref{subsection:offline_results}).
As offline evaluations~can miss what users value---metrics that \mm{} judges may not predict sans extensive engineering effort---we advocate shifting away from just simulation-based evaluations towards user feedback~\cite{10.1145/3170427.3188506}.


\section{RQ2: Lessons for Personalized \dr{}} \label{section:insights}


Beyond surfacing limitations in offline evaluations (\cref{subsection:simulation_results}), we now show how online user feedback also offers richer insights for \dr{}.
We distill these into four lessons to help NLP and HCI researchers design improved personalized \dr{} tools for users.

\lessonsection{1}{Balance User Control and Effort} \label{subsection:control}

Adding control to interactive tools helps, but such effort can exceed what users want to invest~\cite{Shneiderman1983DirectMA, jin2017different}. 
\dr{} interactions are different: as reports take~minutes to write, our pilot survey (Appendix~\ref{appendix:survey}) reveals users~are willing to dedicate more effort up front to avoid subpar outputs.
Most \dr{} tools add this control~via clarification or follow-up questions \cite{jiang2024into}, but surveyed users disliked this feature: they must rearticulate their personal needs for every query.



Instead, \model{} creates a persistent profile~for all queries---matching users who often ``\user{want [the] system to know about me \underline{once} then act accordingly}'' (U21)---and has users pick among actions---easier than answering follow-up queries (Appendix~\ref{appendix:survey}).
Actions still had ample control; U1 noted actions ``\user{are not complex but also very granular which I really like}'' and U4 felt they let them ``\user{do what I cannot express in a perfect way.}'' 
U6 felt actions transparently showed how the system answered queries while U17 felt it kept them more actively engaged than existing tools---helping them ``\user{brainstorm to get what I want}''.
Many believed this process could save time, for U18, ``\user{on the order of days}'', since it avoids the frustration of re-prompting/re-answering generic systems that ``\user{keep missing the point}'' (U3).


\model{} has more control than most \dr{} tools, which made users realize they wanted more, even at~the cost of more effort.  
Some wanted to see~more actions per query (U13, U19), add new actions (U3, U16), or ``emphasize'' certain actions (U4).  
Others wanted to also steer \dr{} via paper filters (U2, U11), multi-turn dialogue (U5, U6), and monitoring the tool as it learned their preferences over time (U15)---updating its memory of the user \cite{yuan-etal-2025-personalized}.

While more personalized \dr{} control seems~helpful, it could risk reinforcing filter bubbles~\cite{Zhang2024SeeWT}---keeping users away from new~ideas. Control also may not benefit all query~types~\cite{Dou2007ALE}: U7 desired a toggle to control when \model{} proposes actions.
Future work must study not just ways to control personalized \dr{}, but \textit{when} it should be controllable and \textit{to what extent}.

\lessonsection{2}{Make Personalization Easy to Digest} \label{subsection:sensemaking}

\dr{} inputs and outputs can be hard to navigate, but \model{}'s structure made content easy to digest.
In most \dr{} tools, users must engineer long queries to express personalized needs (U2, U12): a difficult process for \dr{} users to manage alone \cite{zamfirescu2023johnny}.
Instead, \model{} decomposes tailored query writing via two structured phases---profiles and actions---which helped users efficiently organize their needs (U1, U5, U15); U3 noted: ``\user{I have used 3-4 AI tools and none of them have such steps exactly like that in such a structured~way.}''

While commercial \dr{} masks if and how personalization occurs (\cref{subsection:related_work_personalization}), our report highlights~kept personalization visible/easy to skim: they ``\user{helped focus on text}'' (U12) and were ``\user{easy to cognitively process}'' (U5).  
Still, highlights were tough to get right.
Preferences on highlight frequency~varied---some wanted full sections (U6), others only ``\user{key numbers}'' (U5) or ``\user{evidence}'' (U21)---and for more personalized reports with more actions, highlights were overwhelming (U6, U12).
We also saw signs of over-trust \cite{Lee2004TrustIA}; when no content was highlighted for an action, some assumed the content did not exist, rather than an error from our system.  
\dr{} tools must offer trustworthy sensemaking aids \cite{10.1145/1357054.1357161}, simplifying personalization without overwhelming or misleading~users. 

\lessonsection{3}{Dream Bigger than Just Papers} \label{subsection:modality}

While \model{} infers interests just from chosen papers, participants were surprised by profiles' detail and nuance, often noting ``\user{very important positions [they] would claim}'' (U3) and ``\user{digging stuff [they] didn't think it can detect}'' (U16).~This spurred users to ideate further signals for personalizing profiles, like active project materials and prior queries (Appendix~\ref{appendix:survey}) and new personalization use cases, like collaborator search (U9), biography writing (U3), paper reading (U11), and programming (U12).  


Participants also wanted modalities beyond text and citations in reports, pointing to a wide variety based on their personal needs, like code snippets for frequent coders (U1, U4) and LaTeX formulas for theory-focused researchers (U2, U4), while others generally desired tables/figures (U5, U7) and interactive visualizations \citep[U20]{mondal-etal-2024-presentations}.
By~capturing content preferences, participants~felt reports would give a ``\user{better view in a shorter time}'' (U12) or a ``\user{holistic view of entire papers}'' (U17).

Going beyond papers in personalized \dr{} is a clear win, but has challenges: reasoning over arbitrary user inputs to construct profiles \cite{Zhao2025KnollCA, Shaikh2025CreatingGU} and routing to the best modality to convey content tailored to a user's background \cite{Chen2025GenerativeIF}. 
Future work should study not just how to implement these accurately, but to make them helpful for individual researchers.

\lessonsection{4}{Evaluation Isn't One-Size-Fits-All} \label{subsection:feedback}

While we show offline metrics miss aspects of~personalized \dr{} (\cref{subsection:interview_metrics}, \cref{subsection:simulation_results}), we caveat online studies are not perfect, final fixes \cite{hosking2024human};~for example, edits on action evince not just what users \textit{want} \dr{} to do but also what they \textit{think}~it~can~do---U6 skipped complex actions until they felt it ``\user{understood the basics}.''
Similarly, users need to predict utility \cite{levy1992introduction}.
For instance, participants noted \model{} ``\user{can be a great timesaver}'' (U17) and found points they ``\user{miss while writing research papers}'' (U7),~but such benefits~are~hard to predict \citep{balepur-etal-2024-smart, balepur-etal-2025-good}.
Two directions can bolster personalized \dr{} evaluation: longitudinal studies to account for learning effects \cite{jahani2024generative}, and richer signals to evaluate downstream helpfulness (e.g., time taken, citations clicked).

We thus advocate for mixed evaluations in personalized \dr{}.
For \model{}, formative studies confirmed the workflow met user needs (\cref{subsection:formative}) in early stages, while offline metrics formed a scalable way to check baseline report quality, like citation accuracy and action following (\cref{subsection:offline_results}).
Online evaluation then found what we missed offline---issues to fix in metrics (\cref{subsection:simulation_results}) and how to make \dr{} more useful (\cref{section:insights}).
Overall, evaluations support distinct parts and stages of personalized system design, showing the need for \nlp{} research in personalization to move beyond offline metrics towards real user feedback~\cite{Saxon2024BenchmarksAM, balepur2025these}.



\section{Related Work} \label{section:related_work}

Given our paper's focus on personalized Deep Research (\dr{}), we review personalization generally in \nlp{}   (\cref{subsection:related_work_personalization}) and the advent of \dr{} tools (\cref{subsection:deep_research}).

\subsection{Personalization in \nlp} \label{subsection:related_work_personalization}

Personalization tailors models to user-specific context \cite{Zhang2024PersonalizationOL, Sorensen2024ART}, which can aid engagement \cite{kumar2019understanding}, satisfaction \cite{liang2006personalized}, and learning \cite{Leong2024PuttingTI}---making its usefulness in \dr{} clear.

Most personalized methods \citep{zhang2025modeling}: 1) gather data to form user models; and 2) adapt~outputs to (1).  
User models have used demographics \cite{kirk2024prism}, preferences~\cite{ryan2025synthesizeme}, and user history \cite{Salemi2023LaMPWL},  while adaptation has retrieved user contexts \cite{sun-etal-2025-persona}, prompted via personas \cite{lee-etal-2023-p5}, and tuned parameters or reward models~\cite{tan-etal-2024-democratizing, chen2025pal}.  
\model{} infers user~models from papers and adapts reports through prompting.  

Our survey of 31 ACL'25 papers (Appendix~\ref{appendix:persona_survey}) shows \nlp{} stays offline to test methods; 14 see~if models match outputs from real users \cite{Salemi2023LaMPWL}, but this assumes \mm{}s cannot create outputs better than users. 17 works instead tailor models~to simulated user chats \cite{liu-etal-2025-persona} scored by \mm{}s, but only ten check judges for reliability.
Even when reliable, the ``user'' does not exist \cite{kim2025cupid}, so such judges may not reflect true user needs.
We thus assess personalization online with real users---which only two ACL'25 papers do \cite{flicke2025scholar}---revealing aspects~of personalized \dr{} \mm{} judges struggle to capture. 

\subsection{Deep Research Systems} \label{subsection:deep_research}

Scientific Deep Research (\dr{}) systems synthesize long-form reports via documents to aid scientific~research \cite{Java2025CharacterizingDR}.  
Such systems can create Wikipedia articles \cite{sauper2009automatically, liu2018generating}, expository text \cite{balepur-etal-2023-expository, jiang2025archidocgen}, and surveys \cite{hu2024taxonomy, yan2025surveyforge}.  
\dr{} often employs a mix of retrieval \cite{Lewis2020RetrievalAugmentedGF} and AI agents \cite{yao2023react}, chaining these models to find, organize, and condense scientific papers \cite{wang2024autosurvey}.

As users find \dr{} useful \cite{shen2023beyond}, work has started deploying them online to aid researchers and crowdsource user feedback~\cite{zhao2025sciarena}. 
Open-source UIs/systems include~\storm{} \cite{Shao2024AssistingIW}, \scholarqa{} \cite{singh-etal-2025-ai2}, and \openscholarqa{} \cite{Asai2024OpenScholarSS}, but as our survey reveals (Appendix~\ref{appendix:survey}), most exposure to \dr{} is commercial (e.g. OpenAI \dr{}, Perplexity).

Few \dr{} tools are personalized. 
Exceptions~are Co-\textsc{STORM} \cite{jiang2024into}~and products like OpenAI/Gemini's \dr{} which ask users follow-up queries, but it is unclear if they tailor to~persistent user models.
Concurrently, \citet{liang2025towards}~construct a personalized \dr{} benchmark with synthetic data and \mm{} judges like us (\cref{subsection:dataset}), but do not~release a new system or run online studies.
Instead, \model{} is an open-source personalized \dr{} tool based on adaptive hypermedia design \cite{Brusilovsky1996MethodsAT}: building a model of the user from their papers and using it to tailor actions for input queries.   

\section{Conclusion} \label{section:conclusion}

Synthetic evaluations rule personalization in \nlp{}, but 
our paper on personalized Deep Research (\dr{}) shows how online studies with real users help: they reveal errors that \mm{} judges miss (\cref{subsection:simulation_results}) and guiding lessons for future systems (\cref{section:insights}).
Earlier search engine research used real users~to~push personalization \cite{joachims2002optimizing}, but \nlp{} has recently stopped at easy-to-use \mm{} judges \cite{Zheng2023JudgingLW} claimed to simulate users \cite{Binz2024CentaurAF}, rarely adopting online studies.
We urge readers to fight~the~temptation: simulated benchmarks are preliminary~tests, but real progress in~personalization requires real users \citep{seshadri2026lost}.

As evidence, our online study points to new open questions in personalized \dr{}, as \model{} is imperfect.
These include \nlp{} modeling efforts in factuality (\cref{subsection:interview_metrics}) and multi-modality (\cref{subsection:modality}), and \text{HCI} studies for effortless~control (\cref{subsection:control}) and digestible personalization (\cref{subsection:sensemaking}).
Tackling these will be~cyclical: new methods/evaluations will decide when~\dr{} meets basic needs, while online studies will~reveal what lacks in features/design to~truly~help~users. 

\section*{Acknowledgments}

We would like to thank the Allen Institute for Artificial Intelligence (Ai2) and the CLIP lab at the University of Maryland.
In particular, we thank Connor Baumler and Paiheng Xu for reviewing earlier versions of our interface and study design.
We appreciate discussions with Doug Downey, Dan Weld, Tal August, Rachel Rudinger, Shi Feng, Matt Latzke, Jonathan Bragg, and Jay DeYoung on earlier versions of our prototype and paper.
We thank Dang Nguyen, Navita Goyal, Joy Wongkamjan, Connor Baumler, Paiheng Xu, Yapei Chang, Atrey Desai, and Hyojung Han for voting on candidate titles.
Nishant is especially~grateful to Yapei Chang, Hita Kambhamettu, Federica Bologna, Peiling Jiang, Xinran Zhao, Michael Noukhovitch,~Anej~Svete, Alexis Ross, Akhila Yerukola, and Amanda Bertsch for making the summer in Seattle memorable.

This material is based upon work supported by the National Science Foundation under Grant No. \abr{dge}-2236417 (Balepur) and \abr{iis}-2403436 (Boyd-Graber).
Any opinions, findings, and conclusions or recommendations expressed in this material are those of the author(s) and do not necessarily reflect the views of the National Science Foundation.

\section{Limitations} \label{section:limitations}

Due to resource constraints, we could only deploy our \model{} system online, so our findings on personalization are most directly tied to our setting of Deep Research and execution in \model{}.
However, many of our insights may apply generally; our uncovered issues in \model{}'s outputs (\cref{section:online}, e.g., do not over-claim what applies to a user)
and suggestions for future work (\cref{section:insights}, e.g., extend modalities) apply in general personalization settings.
Further, we are not asserting that these are the only issues to focus on when building personalized systems; as argued in the paper, it is useful to study feedback from end users of your system \cite{Saxon2024BenchmarksAM}.


We also note that our \mm{} judge experiments cannot cover all parameter configurations (\cref{subsection:simulation_results}); although we follow best practices in prompt engineering \cite{Schulhoff2024ThePR}---such as adding few-shot examples and giving clear definitions for each metric---perhaps there is another prompt and \mm{} judge that leads to higher accuracy.
In Appendix~\ref{appendix:simulation}, we show that changing the number of few-shot examples does not largely improve \mm{} prediction accuracy, meaning that it may escape off-the-self models, but could be attainable by training reward models on more data \cite{Liu2024RMBenchBR}. 

\model{} can be slow---generating profiles takes $3$ minutes and reports take $5$ minutes---which can harm user experience \cite{Shneiderman1984ResponseTA}.
While faster than some commercial \dr{} tools (e.g., OpenAI \dr{} took 8 hours for 10 queries), improving efficiency would make \model{} more useful.
To manage delays, we let users view jokes, fun facts, a Chrome Dino Game,\footnote{https://nbalepur.github.io/ai2-trex-runner/} and execution progress in our interface (Appendix~\ref{appendix:interface}), which many users enjoyed.
Future work can focus on reducing latency with smaller, efficient models \cite{Gou2020KnowledgeDA} or better using other waiting time, like pre-computing reports while users spend time customizing actions.

\section{Ethical Considerations} \label{section:ethics}

While the \model{} framework poses no risks in theory, we found that in the profile inference stage, even aligned \mm{}s \cite{Bai2022TrainingAH} can generate potentially offensive or insensitive inferences.
This did not happen with users, but while annotating profiles on our benchmark for agreement, we viewed one such inference: ``\textit{Your papers sometimes contain slightly awkward phrasing or non-standard terminology, suggesting a potential non-native English writing background or direct translation of concepts}''.
This shows that even strong \mm{}s are susceptible to harmful stereotyping in personalization \cite{Kantharuban2024StereotypeOP}, motivating the need for future safeguards to defend against these issues.

We attended each interview with participants to mitigate any risks with our system.
Further, our entire formative and interview study designs were approved by our organization's internal review board.
We collect and release no PII in our data and ensure participants were fairly compensated between \$30--\$40/hr, well above our region's minimum wage. In
Appendix~\ref{appendix:recruitment}, we detail our recruitment protocol.

We use GenAI for UI design, revisions, and analysis; we detail this in Appendix~\ref{appendix:genai} for transparency.

\bibliography{custom}

@inproceedings{lewis2011affective,
  title={Affective computational priming and creativity},
  author={Lewis, Sheena and Dontcheva, Mira and Gerber, Elizabeth},
  booktitle={Proceedings of the SIGCHI Conference on Human Factors in Computing Systems},
  pages={735--744},
  year={2011}
}

@article{Java2025CharacterizingDR,
  title={Characterizing Deep Research: A Benchmark and Formal Definition},
  author={Abhinav Java and Ashmit Khandelwal and Sukruta Prakash Midigeshi and Aaron Halfaker and Amit Deshpande and Navin Goyal and Ankur Gupta and Nagarajan Natarajan and Amit Sharma},
  journal={ArXiv},
  year={2025},
  volume={abs/2508.04183},
  url={https://api.semanticscholar.org/CorpusID:280536600}
}

@inproceedings{Dou2007ALE,
  title={A large-scale evaluation and analysis of personalized search strategies},
  author={Zhicheng Dou and Ruihua Song and Ji-Rong Wen},
  booktitle={The Web Conference},
  year={2007},
  url={https://api.semanticscholar.org/CorpusID:1257668}
}

@inproceedings{Teevan2005PersonalizingSV,
  title={Personalizing search via automated analysis of interests and activities},
  author={Jaime Teevan and Susan T. Dumais and Eric Horvitz},
  booktitle={Annual International ACM SIGIR Conference on Research and Development in Information Retrieval},
  year={2005},
  url={https://api.semanticscholar.org/CorpusID:316030}
}

@article{liang2025towards,
  title={Towards Personalized Deep Research: Benchmarks and Evaluations},
  author={Liang, Yuan and Li, Jiaxian and Wang, Yuqing and Wang, Piaohong and Tian, Motong and Liu, Pai and Qiao, Shuofei and Fang, Runnan and Zhu, He and Zhang, Ge and others},
  journal={arXiv preprint arXiv:2509.25106},
  year={2025}
}

@inproceedings{
bragg2025astabench,
title={AstaBench: Rigorous Benchmarking of {AI} Agents with a Scientific Research Suite},
author={Jonathan Bragg and Mike D'Arcy and Nishant Balepur and Dan Bareket and Bhavana Dalvi Mishra and Sergey Feldman and Dany Haddad and Jena D. Hwang and Peter Jansen and Varsha Kishore and Bodhisattwa Prasad Majumder and Aakanksha Naik and Sigal Rahamimov and Kyle Richardson and Amanpreet Singh and Harshit Surana and Aryeh Tiktinsky and Rosni Vasu and Guy Wiener and Chloe Anastasiades and Stefanus Candra and Jason Dunkelberger and Daniel Emery and Rob Evans and Malachi Hamada and Regan Huff and Rodney Kinney and Matt Latzke and Jaron Lochner and Ruben Lozano-Aguilera and Ngoc-Uyen Nguyen and Smita Rao and Amber Tanaka and Brooke Vlahos and Peter Clark and Doug Downey and Yoav Goldberg and Ashish Sabharwal and Daniel S Weld},
booktitle={The Fourteenth International Conference on Learning Representations},
year={2026},
url={https://openreview.net/forum?id=M7TNf5J26u}
}

@proceedings{acl-ws-2025-long,
    title = "Proceedings of the 63rd Annual Meeting of the Association for Computational Linguistics (Volume 1: Long Papers)",
    editor = "Che, Wanxiang  and
      Nabende, Joyce  and
      Shutova, Ekaterina  and
      Pilehvar, Mohammad Taher",
    month = jul,
    year = "2025",
    address = "Vienna, Austria",
    publisher = "Association for Computational Linguistics",
    url = "https://aclanthology.org/2025.acl-long.0/",
    doi = "10.18653/v1/2025.acl-long.0",
    ISBN = "979-8-89176-251-0"
}

@inproceedings{liu-etal-2025-persona,
    title = "A Persona-Aware {LLM}-Enhanced Framework for Multi-Session Personalized Dialogue Generation",
    author = "Liu, Dongshuo  and
      Wu, Zhijing  and
      Song, Dandan  and
      Huang, Heyan",
    editor = "Che, Wanxiang  and
      Nabende, Joyce  and
      Shutova, Ekaterina  and
      Pilehvar, Mohammad Taher",
    booktitle = "Findings of the Association for Computational Linguistics: ACL 2025",
    month = jul,
    year = "2025",
    address = "Vienna, Austria",
    publisher = "Association for Computational Linguistics",
    url = "https://aclanthology.org/2025.findings-acl.5/",
    doi = "10.18653/v1/2025.findings-acl.5",
    pages = "103--123",
    ISBN = "979-8-89176-256-5"
}

@article{flicke2025scholar,
  title={Scholar Inbox: Personalized Paper Recommendations for Scientists},
  author={Flicke, Markus and Angrabeit, Glenn and Iyengar, Madhav and Protsenko, Vitalii and Shakun, Illia and Cicvaric, Jovan and Kargi, Bora and He, Haoyu and Schuler, Lukas and Scholz, Lewin and others},
  journal={arXiv preprint arXiv:2504.08385},
  year={2025}
}

@inproceedings{Shao2024AssistingIW,
  title={Assisting in Writing Wikipedia-like Articles From Scratch with Large Language Models},
  author={Yijia Shao and Yucheng Jiang and Theodore A. Kanell and Peter Xu and Omar Khattab and Monica S. Lam},
  booktitle={North American Chapter of the Association for Computational Linguistics},
  year={2024},
  url={https://api.semanticscholar.org/CorpusID:267782917}
}

@article{parolo2015attention,
  title={Attention decay in science},
  author={Parolo, Pietro Della Briotta and Pan, Raj Kumar and Ghosh, Rumi and Huberman, Bernardo A and Kaski, Kimmo and Fortunato, Santo},
  journal={Journal of Informetrics},
  volume={9},
  number={4},
  pages={734--745},
  year={2015},
  publisher={Elsevier}
}

@article{huang2025deep,
  title={Deep Research Agents: A Systematic Examination And Roadmap},
  author={Huang, Yuxuan and Chen, Yihang and Zhang, Haozheng and Li, Kang and Fang, Meng and Yang, Linyi and Li, Xiaoguang and Shang, Lifeng and Xu, Songcen and Hao, Jianye and others},
  journal={arXiv preprint arXiv:2506.18096},
  year={2025}
}

@inproceedings{pu2025ideasynth,
  title={Ideasynth: Iterative research idea development through evolving and composing idea facets with literature-grounded feedback},
  author={Pu, Kevin and Feng, KJ Kevin and Grossman, Tovi and Hope, Tom and Dalvi Mishra, Bhavana and Latzke, Matt and Bragg, Jonathan and Chang, Joseph Chee and Siangliulue, Pao},
  booktitle={Proceedings of the 2025 CHI Conference on Human Factors in Computing Systems},
  pages={1--31},
  year={2025}
}

@inproceedings{singh-etal-2025-ai2,
    title = "Ai2 Scholar {QA}: Organized Literature Synthesis with Attribution",
    author = "Singh, Amanpreet  and
      Chang, Joseph Chee  and
      Haddad, Dany  and
      Naik, Aakanksha  and
      Hwang, Jena D.  and
      Kinney, Rodney  and
      Weld, Daniel S  and
      Downey, Doug  and
      Feldman, Sergey",
    editor = "Mishra, Pushkar  and
      Muresan, Smaranda  and
      Yu, Tao",
    booktitle = "Proceedings of the 63rd Annual Meeting of the Association for Computational Linguistics (Volume 3: System Demonstrations)",
    month = jul,
    year = "2025",
    address = "Vienna, Austria",
    publisher = "Association for Computational Linguistics",
    url = "https://aclanthology.org/2025.acl-demo.49/",
    doi = "10.18653/v1/2025.acl-demo.49",
    pages = "513--523",
    ISBN = "979-8-89176-253-4"
}

@article{Asai2024OpenScholarSS,
  title={OpenScholar: Synthesizing Scientific Literature with Retrieval-augmented LMs},
  author={Akari Asai and Jacqueline He and Rulin Shao and Weijia Shi and Amanpreet Singh and Joseph Chee Chang and Kyle Lo and Luca Soldaini and Sergey Feldman and Mike D'Arcy and David Wadden and Matt Latzke and Minyang Tian and Pan Ji and Shengyan Liu and Hao Tong and Bohao Wu and Yanyu Xiong and Luke S. Zettlemoyer and Graham Neubig and Dan Weld and Doug Downey and Wen-tau Yih and Pang Wei Koh and Hanna Hajishirzi},
  journal={ArXiv},
  year={2024},
  volume={abs/2411.14199},
  url={https://api.semanticscholar.org/CorpusID:274166189}
}

@article{DeepSeekAI2024DeepSeekV3TR,
  title={DeepSeek-V3 Technical Report},
  author={DeepSeek-AI and Aixin Liu and Bei Feng and Bing Xue and Bing-Li Wang and Bochao Wu and Chengda Lu and Chenggang Zhao and Chengqi Deng and Chenyu Zhang and Chong Ruan and Damai Dai and Daya Guo and Dejian Yang and Deli Chen and Dong-Li Ji and Erhang Li and Fangyun Lin and Fucong Dai and Fuli Luo and Guangbo Hao and Guanting Chen and Guowei Li and H. Zhang and Han Bao and Hanwei Xu and Haocheng Wang and Haowei Zhang and Honghui Ding and Huajian Xin and Huazuo Gao and Hui Li and Hui Qu and J. L Cai and Jian Liang and Jianzhong Guo and Jiaqi Ni and Jiashi Li and Jiawei Wang and Jin Chen and Jingchang Chen and Jingyang Yuan and Junjie Qiu and Junlong Li and Jun-Mei Song and Kai Dong and Kai Hu and Kaige Gao and Kang Guan and Kexin Huang and Kuai Yu and Lean Wang and Lecong Zhang and Lei Xu and Leyi Xia and Liang Zhao and Litong Wang and Liyue Zhang and Meng Li and Miaojun Wang and Mingchuan Zhang and Minghua Zhang and Minghui Tang and Mingming Li and Ning Tian and Panpan Huang and Peiyi Wang and Peng Zhang and Qiancheng Wang and Qihao Zhu and Qinyu Chen and Qiushi Du and R. J. Chen and Ruiqi Jin and Ruiqi Ge and Ruisong Zhang and Ruizhe Pan and Runji Wang and Runxin Xu and Ruoyu Zhang and Ruyi Chen and S. S. Li and Shanghao Lu and Shangyan Zhou and Shanhuang Chen and Shao-Ping Wu and Shengfeng Ye and Shirong Ma and Shiyu Wang and Shuang Zhou and Shuiping Yu and Shunfeng Zhou and Shuting Pan and T. Wang and Tao Yun and Tian Pei and Tianyu Sun and Wangding Xiao and Wangding Zeng and Wanjia Zhao and Wei An and Wen Liu and Wenfeng Liang and Wenjun Gao and Wen-Xuan Yu and Wentao Zhang and X. Q. Li and Xiangyu Jin and Xianzu Wang and Xiaoling Bi and Xiaodong Liu and Xiaohan Wang and Xi-Cheng Shen and Xi-aokang Chen and Xiaokang Zhang and Xiaosha Chen and Xiaotao Nie and Xiaowen Sun and Xiaoxiang Wang and Xin Cheng and Xin Liu and Xin Xie and Xingchao Liu and Xingkai Yu and Xinnan Song and Xinxia Shan and Xinyi Zhou and Xinyu Yang and Xinyuan Li and Xuecheng Su and Xuheng Lin and Y. K. Li and Y. Q. Wang and Y. X. Wei and Y. X. Zhu and Yang Zhang and Yanhong Xu and Yanping Huang and Yao Li and Yao Zhao and Yaofeng Sun and Yao Li and Yaohui Wang and Yi Yu and Yi Zheng and Yichao Zhang and Yifan Shi and Yi Xiong and Ying He and Ying Tang and Yishi Piao and Yisong Wang and Yixuan Tan and Yi-Bing Ma and Yiyuan Liu and Yongqiang Guo and Yu Wu and Yuan Ou and Yuchen Zhu and Yuduan Wang and Yue Gong and Yuheng Zou and Yujia He and Yukun Zha and Yunfan Xiong and Yunxiang Ma and Yuting Yan and Yu-Wei Luo and Yu-mei You and Yuxuan Liu and Yuyang Zhou and Z. F. Wu and Zehui Ren and Zehui Ren and Zhangli Sha and Zhe Fu and Zhean Xu and Zhen Huang and Zhen Zhang and Zhenda Xie and Zhen-guo Zhang and Zhewen Hao and Zhibin Gou and Zhicheng Ma and Zhigang Yan and Zhihong Shao and Zhipeng Xu and Zhiyu Wu and Zhongyu Zhang and Zhuoshu Li and Zihui Gu and Zijia Zhu and Zijun Liu and Zi-An Li and Ziwei Xie and Ziyang Song and Ziyi Gao and Zizheng Pan},
  journal={ArXiv},
  year={2024},
  volume={abs/2412.19437},
  url={https://api.semanticscholar.org/CorpusID:275118643}

}

@inproceedings{jin2017different,
  title={How do different levels of user control affect cognitive load and acceptance of recommendations?},
  author={Jin, Yucheng and Cardoso, Bruno and Verbert, Katrien},
  booktitle={Jin, Y., Cardoso, B. and Verbert, K., 2017, August. How do different levels of user control affect cognitive load and acceptance of recommendations?. In Proceedings of the 4th Joint Workshop on Interfaces and Human Decision Making for Recommender Systems co-located with ACM Conference on Recommender Systems (RecSys 2017)},
  volume={1884},
  pages={35--42},
  year={2017},
  organization={CEUR Workshop Proceedings}
}

@article{jin2022deep,
  title={Deep learning for text style transfer: A survey},
  author={Jin, Di and Jin, Zhijing and Hu, Zhiting and Vechtomova, Olga and Mihalcea, Rada},
  journal={Computational Linguistics},
  volume={48},
  number={1},
  pages={155--205},
  year={2022},
  publisher={MIT Press One Broadway, 12th Floor, Cambridge, Massachusetts 02142, USA~…}
}

@inproceedings{balepur-etal-2024-smart,
    title = "A {SMART} Mnemonic Sounds like ``Glue Tonic'': Mixing {LLM}s with Student Feedback to Make Mnemonic Learning Stick",
    author = "Balepur, Nishant  and
      Shu, Matthew  and
      Hoyle, Alexander  and
      Robey, Alison  and
      Feng, Shi  and
      Goldfarb-Tarrant, Seraphina  and
      Boyd-Graber, Jordan Lee",
    editor = "Al-Onaizan, Yaser  and
      Bansal, Mohit  and
      Chen, Yun-Nung",
    booktitle = "Proceedings of the 2024 Conference on Empirical Methods in Natural Language Processing",
    month = nov,
    year = "2024",
    address = "Miami, Florida, USA",
    publisher = "Association for Computational Linguistics",
    url = "https://aclanthology.org/2024.emnlp-main.786/",
    doi = "10.18653/v1/2024.emnlp-main.786",
    pages = "14202--14225"
}

@inproceedings{balepur-etal-2025-good,
    title = "A Good Plan is Hard to Find: Aligning Models with Preferences is Misaligned with What Helps Users",
    author = "Balepur, Nishant  and
      Shu, Matthew  and
      Sung, Yoo Yeon  and
      Goldfarb-Tarrant, Seraphina  and
      Feng, Shi  and
      Yang, Fumeng  and
      Rudinger, Rachel  and
      Boyd-Graber, Jordan Lee",
    editor = "Christodoulopoulos, Christos  and
      Chakraborty, Tanmoy  and
      Rose, Carolyn  and
      Peng, Violet",
    booktitle = "Proceedings of the 2025 Conference on Empirical Methods in Natural Language Processing",
    month = nov,
    year = "2025",
    address = "Suzhou, China",
    publisher = "Association for Computational Linguistics",
    url = "https://aclanthology.org/2025.emnlp-main.585/",
    doi = "10.18653/v1/2025.emnlp-main.585",
    pages = "11568--11595",
    ISBN = "979-8-89176-332-6"
}

@inproceedings{
seshadri2026lost,
title={Lost in Simulation: {LLM}-Simulated Users are Unreliable Proxies for Human Users in Agentic Evaluations},
author={Preethi Seshadri and Samuel Cahyawijaya and Ayomide Odumakinde and Sameer Singh and Seraphina Goldfarb-Tarrant},
booktitle={Algorithmic Fairness Across Alignment Procedures and Agentic Systems},
year={2026},
url={https://openreview.net/forum?id=m57vJLBHxA}
}

@inproceedings{10.1145/3170427.3188506,
author = {Du, Fan and Malik, Sana and Theocharous, Georgios and Koh, Eunyee},
title = {Personalizable and Interactive Sequence Recommender System},
year = {2018},
isbn = {9781450356213},
publisher = {Association for Computing Machinery},
address = {New York, NY, USA},
url = {https://doi.org/10.1145/3170427.3188506},
doi = {10.1145/3170427.3188506},
booktitle = {Extended Abstracts of the 2018 CHI Conference on Human Factors in Computing Systems},
pages = {1–6},
numpages = {6},
location = {Montreal QC, Canada},
series = {CHI EA '18}
}

@article{DeepSeekAI2025DeepSeekR1IR,
  title={DeepSeek-R1: Incentivizing Reasoning Capability in LLMs via Reinforcement Learning},
  author={DeepSeek-AI and Daya Guo and Dejian Yang and Haowei Zhang and Jun-Mei Song and Ruoyu Zhang and Runxin Xu and Qihao Zhu and Shirong Ma and Peiyi Wang and Xiaoling Bi and Xiaokang Zhang and Xingkai Yu and Yu Wu and Z. F. Wu and Zhibin Gou and Zhihong Shao and Zhuoshu Li and Ziyi Gao and Aixin Liu and Bing Xue and Bing-Li Wang and Bochao Wu and Bei Feng and Chengda Lu and Chenggang Zhao and Chengqi Deng and Chenyu Zhang and Chong Ruan and Damai Dai and Deli Chen and Dong-Li Ji and Erhang Li and Fangyun Lin and Fucong Dai and Fuli Luo and Guangbo Hao and Guanting Chen and Guowei Li and H. Zhang and Han Bao and Hanwei Xu and Haocheng Wang and Honghui Ding and Huajian Xin and Huazuo Gao and Hui Qu and Hui Li and Jianzhong Guo and Jiashi Li and Jiawei Wang and Jingchang Chen and Jingyang Yuan and Junjie Qiu and Junlong Li and Jiong Cai and Jiaqi Ni and Jian Liang and Jin Chen and Kai Dong and Kai Hu and Kaige Gao and Kang Guan and Kexin Huang and Kuai Yu and Lean Wang and Lecong Zhang and Liang Zhao and Litong Wang and Liyue Zhang and Lei Xu and Leyi Xia and Mingchuan Zhang and Minghua Zhang and M. Tang and Meng Li and Miaojun Wang and Mingming Li and Ning Tian and Panpan Huang and Peng Zhang and Qiancheng Wang and Qinyu Chen and Qiushi Du and Ruiqi Ge and Ruisong Zhang and Ruizhe Pan and Runji Wang and R. J. Chen and Ruiqi Jin and Ruyi Chen and Shanghao Lu and Shangyan Zhou and Shanhuang Chen and Shengfeng Ye and Shiyu Wang and Shuiping Yu and Shunfeng Zhou and Shuting Pan and S. S. Li and Shuang Zhou and Shao-Kang Wu and Tao Yun and Tian Pei and Tianyu Sun and T. Wang and Wangding Zeng and Wanjia Zhao and Wen Liu and Wenfeng Liang and Wenjun Gao and Wen-Xia Yu and Wentao Zhang and Wangding Xiao and Wei An and Xiaodong Liu and Xiaohan Wang and Xi-aokang Chen and Xiaotao Nie and Xin Cheng and Xin Liu and Xin Xie and Xingchao Liu and Xinyu Yang and Xinyuan Li and Xuecheng Su and Xuheng Lin and X. Q. Li and Xiangyu Jin and Xi-Cheng Shen and Xiaosha Chen and Xiaowen Sun and Xiaoxiang Wang and Xinnan Song and Xinyi Zhou and Xianzu Wang and Xinxia Shan and Y. K. Li and Y. Q. Wang and Y. X. Wei and Yang Zhang and Yanhong Xu and Yao Li and Yao Zhao and Yaofeng Sun and Yaohui Wang and Yi Yu and Yichao Zhang and Yifan Shi and Yi Xiong and Ying He and Yishi Piao and Yisong Wang and Yixuan Tan and Yiyang Ma and Yiyuan Liu and Yongqiang Guo and Yuan Ou and Yuduan Wang and Yue Gong and Yu-Jing Zou and Yujia He and Yunfan Xiong and Yu-Wei Luo and Yu-mei You and Yuxuan Liu and Yuyang Zhou and Y. X. Zhu and Yanping Huang and Yao Li and Yi Zheng and Yuchen Zhu and Yunxiang Ma and Ying Tang and Yukun Zha and Yuting Yan and Zehui Ren and Zehui Ren and Zhangli Sha and Zhe Fu and Zhean Xu and Zhenda Xie and Zhen-guo Zhang and Zhewen Hao and Zhicheng Ma and Zhigang Yan and Zhiyu Wu and Zihui Gu and Zijia Zhu and Zijun Liu and Zi-An Li and Ziwei Xie and Ziyang Song and Zizheng Pan and Zhen Huang and Zhipeng Xu and Zhongyu Zhang and Zhen Zhang},
  journal={ArXiv},
  year={2025},
  volume={abs/2501.12948},
  url={https://api.semanticscholar.org/CorpusID:275789950}
}

@inproceedings{balepur-etal-2023-expository,
    title = "Expository Text Generation: Imitate, Retrieve, Paraphrase",
    author = "Balepur, Nishant  and
      Huang, Jie  and
      Chang, Kevin",
    editor = "Bouamor, Houda  and
      Pino, Juan  and
      Bali, Kalika",
    booktitle = "Proceedings of the 2023 Conference on Empirical Methods in Natural Language Processing",
    month = dec,
    year = "2023",
    address = "Singapore",
    publisher = "Association for Computational Linguistics",
    url = "https://aclanthology.org/2023.emnlp-main.729/",
    doi = "10.18653/v1/2023.emnlp-main.729",
    pages = "11896--11919"
}

@inproceedings{lambert-etal-2025-rewardbench,
    title = "{R}eward{B}ench: Evaluating Reward Models for Language Modeling",
    author = "Lambert, Nathan  and
      Pyatkin, Valentina  and
      Morrison, Jacob  and
      Miranda, LJ  and
      Lin, Bill Yuchen  and
      Chandu, Khyathi  and
      Dziri, Nouha  and
      Kumar, Sachin  and
      Zick, Tom  and
      Choi, Yejin  and
      Smith, Noah A.  and
      Hajishirzi, Hannaneh",
    editor = "Chiruzzo, Luis  and
      Ritter, Alan  and
      Wang, Lu",
    booktitle = "Findings of the Association for Computational Linguistics: NAACL 2025",
    month = apr,
    year = "2025",
    address = "Albuquerque, New Mexico",
    publisher = "Association for Computational Linguistics",
    url = "https://aclanthology.org/2025.findings-naacl.96/",
    doi = "10.18653/v1/2025.findings-naacl.96",
    pages = "1755--1797",
    ISBN = "979-8-89176-195-7"
}

@article{ryan2025synthesizeme,
  title={SynthesizeMe! Inducing Persona-Guided Prompts for Personalized Reward Models in LLMs},
  author={Ryan, Michael J and Shaikh, Omar and Bhagirath, Aditri and Frees, Daniel and Held, William and Yang, Diyi},
  journal={arXiv preprint arXiv:2506.05598},
  year={2025}
}

@article{kim2025cupid,
  title={CUPID: Evaluating Personalized and Contextualized Alignment of LLMs from Interactions},
  author={Kim, Tae Soo and Lee, Yoonjoo and Park, Yoonah and Kim, Jiho and Kim, Young-Ho and Kim, Juho},
  journal={arXiv preprint arXiv:2508.01674},
  year={2025}
}

@inproceedings{lee-etal-2023-p5,
    title = "P5: Plug-and-Play Persona Prompting for Personalized Response Selection",
    author = "Lee, Joosung  and
      Oh, Minsik  and
      Lee, Donghun",
    editor = "Bouamor, Houda  and
      Pino, Juan  and
      Bali, Kalika",
    booktitle = "Proceedings of the 2023 Conference on Empirical Methods in Natural Language Processing",
    month = dec,
    year = "2023",
    address = "Singapore",
    publisher = "Association for Computational Linguistics",
    url = "https://aclanthology.org/2023.emnlp-main.1031/",
    doi = "10.18653/v1/2023.emnlp-main.1031",
    pages = "16571--16582"
}

@inproceedings{
chen2025pal,
title={{PAL}: Sample-Efficient Personalized Reward Modeling for Pluralistic Alignment},
author={Daiwei Chen and Yi Chen and Aniket Rege and Zhi Wang and Ramya Korlakai Vinayak},
booktitle={The Thirteenth International Conference on Learning Representations},
year={2025},
url={https://openreview.net/forum?id=1kFDrYCuSu}
}

@inproceedings{sun-etal-2025-persona,
    title = "Persona-{DB}: Efficient Large Language Model Personalization for Response Prediction with Collaborative Data Refinement",
    author = "Sun, Chenkai  and
      Yang, Ke  and
      Gangi Reddy, Revanth  and
      Fung, Yi  and
      Chan, Hou Pong  and
      Small, Kevin  and
      Zhai, ChengXiang  and
      Ji, Heng",
    editor = "Rambow, Owen  and
      Wanner, Leo  and
      Apidianaki, Marianna  and
      Al-Khalifa, Hend  and
      Eugenio, Barbara Di  and
      Schockaert, Steven",
    booktitle = "Proceedings of the 31st International Conference on Computational Linguistics",
    month = jan,
    year = "2025",
    address = "Abu Dhabi, UAE",
    publisher = "Association for Computational Linguistics",
    url = "https://aclanthology.org/2025.coling-main.20/",
    pages = "281--296"
}

@inproceedings{tan-etal-2024-democratizing,
    title = "Democratizing Large Language Models via Personalized Parameter-Efficient Fine-tuning",
    author = "Tan, Zhaoxuan  and
      Zeng, Qingkai  and
      Tian, Yijun  and
      Liu, Zheyuan  and
      Yin, Bing  and
      Jiang, Meng",
    editor = "Al-Onaizan, Yaser  and
      Bansal, Mohit  and
      Chen, Yun-Nung",
    booktitle = "Proceedings of the 2024 Conference on Empirical Methods in Natural Language Processing",
    month = nov,
    year = "2024",
    address = "Miami, Florida, USA",
    publisher = "Association for Computational Linguistics",
    url = "https://aclanthology.org/2024.emnlp-main.372/",
    doi = "10.18653/v1/2024.emnlp-main.372",
    pages = "6476--6491"
}

@article{kirk2024prism,
  title={The PRISM alignment dataset: What participatory, representative and individualised human feedback reveals about the subjective and multicultural alignment of large language models},
  author={Kirk, Hannah Rose and Whitefield, Alexander and Rottger, Paul and Bean, Andrew M and Margatina, Katerina and Mosquera-Gomez, Rafael and Ciro, Juan and Bartolo, Max and Williams, Adina and He, He and others},
  journal={Advances in Neural Information Processing Systems},
  volume={37},
  pages={105236--105344},
  year={2024}
}

@article{Leong2024PuttingTI,
  title={Putting Things into Context: Generative AI-Enabled Context Personalization for Vocabulary Learning Improves Learning Motivation},
  author={Joanne Leong and Pat Pataranutaporn and Valdemar Danry and Florian Perteneder and Yaoli Mao and Pattie Maes},
  journal={Proceedings of the 2024 CHI Conference on Human Factors in Computing Systems},
  year={2024},
  url={https://api.semanticscholar.org/CorpusID:269750914}
}

@article{kumar2019understanding,
  title={Understanding the role of artificial intelligence in personalized engagement marketing},
  author={Kumar, Vipin and Rajan, Bharath and Venkatesan, Rajkumar and Lecinski, Jim},
  journal={California management review},
  volume={61},
  number={4},
  pages={135--155},
  year={2019},
  publisher={SAGE Publications Sage CA: Los Angeles, CA}
}

@article{liang2006personalized,
  title={Personalized content recommendation and user satisfaction: Theoretical synthesis and empirical findings},
  author={Liang, Ting-Peng and Lai, Hung-Jen and Ku, Yi-Cheng},
  journal={Journal of Management Information Systems},
  volume={23},
  number={3},
  pages={45--70},
  year={2006},
  publisher={Taylor \& Francis}
}

@article{Sorensen2024ART,
  title={A Roadmap to Pluralistic Alignment},
  author={Taylor Sorensen and Jared Moore and Jillian R. Fisher and Mitchell Gordon and Niloofar Mireshghallah and Christopher Rytting and Andre Ye and Liwei Jiang and Ximing Lu and Nouha Dziri and Tim Althoff and Yejin Choi},
  journal={ArXiv},
  year={2024},
  volume={abs/2402.05070},
  url={https://api.semanticscholar.org/CorpusID:267523348}
}

@article{jiang2024into,
  title={Into the unknown unknowns: Engaged human learning through participation in language model agent conversations},
  author={Jiang, Yucheng and Shao, Yijia and Ma, Dekun and Semnani, Sina J and Lam, Monica S},
  journal={arXiv preprint arXiv:2408.15232},
  year={2024}
}

@inproceedings{jiang2025archidocgen,
  title={ArchiDocGen: Multi-Agent Framework for Expository Document Generation in the Architectural Industry},
  author={Jiang, Junjie and Wu, Haodong and Zhang, Yongqi and Guo, Songyue and Liu, Bingcen and Cao, Caleb Chen and Shao, Ruizhe and Guan, Chao and Xu, Peng and Chen, Lei},
  booktitle={Proceedings of the 63rd Annual Meeting of the Association for Computational Linguistics (Volume 6: Industry Track)},
  pages={605--618},
  year={2025}
}

@article{zhao2025sciarena,
  title={SciArena: An Open Evaluation Platform for Foundation Models in Scientific Literature Tasks},
  author={Zhao, Yilun and Zhang, Kaiyan and Hu, Tiansheng and Wu, Sihong and Bras, Ronan Le and Anderson, Taira and Bragg, Jonathan and Chang, Joseph Chee and Dodge, Jesse and Latzke, Matt and others},
  journal={arXiv preprint arXiv:2507.01001},
  year={2025}
}

@article{shen2023beyond,
  title={Beyond summarization: Designing ai support for real-world expository writing tasks},
  author={Shen, Zejiang and August, Tal and Siangliulue, Pao and Lo, Kyle and Bragg, Jonathan and Hammerbacher, Jeff and Downey, Doug and Chang, Joseph Chee and Sontag, David},
  journal={arXiv preprint arXiv:2304.02623},
  year={2023}
}

@article{wang2024autosurvey,
  title={Autosurvey: Large language models can automatically write surveys},
  author={Wang, Yidong and Guo, Qi and Yao, Wenjin and Zhang, Hongbo and Zhang, Xin and Wu, Zhen and Zhang, Meishan and Dai, Xinyu and Wen, Qingsong and Ye, Wei and others},
  journal={Advances in neural information processing systems},
  volume={37},
  pages={115119--115145},
  year={2024}
}

@article{yan2025surveyforge,
  title={Surveyforge: On the outline heuristics, memory-driven generation, and multi-dimensional evaluation for automated survey writing},
  author={Yan, Xiangchao and Feng, Shiyang and Yuan, Jiakang and Xia, Renqiu and Wang, Bin and Zhang, Bo and Bai, Lei},
  journal={arXiv preprint arXiv:2503.04629},
  year={2025}
}

@article{hu2024taxonomy,
  title={Taxonomy Tree Generation from Citation Graph},
  author={Hu, Yuntong and Li, Zhuofeng and Zhang, Zheng and Ling, Chen and Kanjiani, Raasikh and Zhao, Boxin and Zhao, Liang},
  journal={arXiv preprint arXiv:2410.03761},
  year={2024}
}

@article{liu2018generating,
  title={Generating wikipedia by summarizing long sequences},
  author={Liu, Peter J and Saleh, Mohammad and Pot, Etienne and Goodrich, Ben and Sepassi, Ryan and Kaiser, Lukasz and Shazeer, Noam},
  journal={arXiv preprint arXiv:1801.10198},
  year={2018}
}

@inproceedings{sauper2009automatically,
  title={Automatically generating wikipedia articles: A structure-aware approach},
  author={Sauper, Christina and Barzilay, Regina},
  booktitle={Proceedings of the Joint Conference of the 47th Annual Meeting of the ACL and the 4th International Joint Conference on Natural Language Processing of the AFNLP},
  pages={208--216},
  year={2009}
}

@article{Lewis2020RetrievalAugmentedGF,
  title={Retrieval-Augmented Generation for Knowledge-Intensive NLP Tasks},
  author={Patrick Lewis and Ethan Perez and Aleksandara Piktus and Fabio Petroni and Vladimir Karpukhin and Naman Goyal and Heinrich Kuttler and Mike Lewis and Wen-tau Yih and Tim Rockt{\"a}schel and Sebastian Riedel and Douwe Kiela},
  journal={ArXiv},
  year={2020},
  volume={abs/2005.11401},
  url={https://api.semanticscholar.org/CorpusID:218869575}
}

@inproceedings{yao2023react,
  title={React: Synergizing reasoning and acting in language models},
  author={Yao, Shunyu and Zhao, Jeffrey and Yu, Dian and Du, Nan and Shafran, Izhak and Narasimhan, Karthik and Cao, Yuan},
  booktitle={International Conference on Learning Representations (ICLR)},
  year={2023}
}

@article{Binz2024CentaurAF,
  title={Centaur: a foundation model of human cognition},
  author={Marcel Binz and Elif Akata and Matthias Bethge and Franziska Brandle and Frederick Callaway and Julian Coda-Forno and Peter Dayan and Can Demircan and Maria K. Eckstein and No'emi 'EltetHo and Thomas L. Griffiths and Susanne Haridi and Akshay Kumar Jagadish and Ji-An Li and Alex Kipnis and Sreejan Kumar and Tobias Ludwig and Marvin Mathony and Marcelo G. Mattar and Alireza Modirshanechi and Surabhi S. Nath and Joshua C. Peterson and Milena Rmu{\v{s}} and Evan M. Russek and Tankred Saanum and Natalia Scharfenberg and Johannes A. Schubert and Luca M. Schulze Buschoff and Nishad Singhi and Xin Sui and Mirko Thalmann and Fabian J. Theis and Vuong Truong and Vishaal Udandarao and Konstantinos Voudouris and Robert C. Wilson and Kristin Witte and Shuchen Wu and Dirk Wulff and Huadong Xiong and Eric Schulz},
  journal={ArXiv},
  year={2024},
  volume={abs/2410.20268},
  url={https://api.semanticscholar.org/CorpusID:273654604}
}

@article{Zhang2024PersonalizationOL,
  title={Personalization of Large Language Models: A Survey},
  author={Zhehao Zhang and Ryan A. Rossi and Branislav Kveton and Yijia Shao and Diyi Yang and Hamed Zamani and Franck Dernoncourt and Joe Barrow and Tong Yu and Sungchul Kim and Ruiyi Zhang and Jiuxiang Gu and Tyler Derr and Hongjie Chen and Ju-Ying Wu and Xiang Chen and Zichao Wang and Subrata Mitra and Nedim Lipka and Nesreen K. Ahmed and Yu Wang},
  journal={ArXiv},
  year={2024},
  volume={abs/2411.00027},
  url={https://api.semanticscholar.org/CorpusID:273798244}
}

@article{Brusilovsky1996MethodsAT,
  title={Methods and techniques of adaptive hypermedia},
  author={Peter Brusilovsky},
  journal={User Modeling and User-Adapted Interaction},
  year={1996},
  volume={6},
  pages={87-129},
  url={https://api.semanticscholar.org/CorpusID:16808655}
}

@article{Kuratov2024BABILongTT,
  title={BABILong: Testing the Limits of LLMs with Long Context Reasoning-in-a-Haystack},
  author={Yuri Kuratov and Aydar Bulatov and Petr Anokhin and Ivan Rodkin and Dmitry Sorokin and Artyom Y. Sorokin and Mikhail Burtsev},
  journal={ArXiv},
  year={2024},
  volume={abs/2406.10149},
  url={https://api.semanticscholar.org/CorpusID:270521583}
}

@inproceedings{10.1145/3539618.3591677,
author = {Mysore, Sheshera and Jasim, Mahmood and Mccallum, Andrew and Zamani, Hamed},
title = {Editable User Profiles for Controllable Text Recommendations},
year = {2023},
isbn = {9781450394086},
publisher = {Association for Computing Machinery},
address = {New York, NY, USA},
url = {https://doi.org/10.1145/3539618.3591677},
doi = {10.1145/3539618.3591677},
booktitle = {Proceedings of the 46th International ACM SIGIR Conference on Research and Development in Information Retrieval},
pages = {993–1003},
numpages = {11},
location = {Taipei, Taiwan},
series = {SIGIR '23}
}

@article{
mozannar2025the,
title={The RealHumanEval: Evaluating Large Language Models{\textquoteright} Abilities to Support Programmers},
author={Hussein Mozannar and Valerie Chen and Mohammed Alsobay and Subhro Das and Sebastian Zhao and Dennis Wei and Manish Nagireddy and Prasanna Sattigeri and Ameet Talwalkar and David Sontag},
journal={Transactions on Machine Learning Research},
issn={2835-8856},
year={2025},
url={https://openreview.net/forum?id=hGaWq5Buj7},
note={Expert Certification}
}

@article{levy1992introduction,
  title={An introduction to prospect theory},
  author={Levy, Jack S},
  journal={Political psychology},
  pages={171--186},
  year={1992},
  publisher={JSTOR}
}

@inproceedings{
hosking2024human,
title={Human Feedback is not Gold Standard},
author={Tom Hosking and Phil Blunsom and Max Bartolo},
booktitle={The Twelfth International Conference on Learning Representations},
year={2024},
url={https://openreview.net/forum?id=7W3GLNImfS}
}

@article{jahani2024generative,
  title={As generative models improve, people adapt their prompts},
  author={Jahani, Eaman and Manning, Benjamin and Zhang, Joe and Tu Ye, Hong Yi and Alsobay, Mohammed and Nicolaides, Christos and Suri, Siddharth and Holtz, David},
  journal={People Adapt Their Prompts*(July 18, 2024)},
  year={2024}
}

@inproceedings{Chen2025GenerativeIF,
  title={Generative Interfaces for Language Models},
  author={Jiaqi Chen and Yanzhe Zhang and Yutong Zhang and Yijia Shao and Diyi Yang},
  year={2025},
  url={https://api.semanticscholar.org/CorpusID:280869842}
}

@article{Zhao2025KnollCA,
  title={Knoll: Creating a Knowledge Ecosystem for Large Language Models},
  author={Dora Zhao and Diyi Yang and Michael S. Bernstein},
  journal={ArXiv},
  year={2025},
  volume={abs/2505.19335},
  url={https://api.semanticscholar.org/CorpusID:278904985}
}

@article{Liu2024RMBenchBR,
  title={RM-Bench: Benchmarking Reward Models of Language Models with Subtlety and Style},
  author={Yantao Liu and Zijun Yao and Rui Min and Yixin Cao and Lei Hou and Juanzi Li},
  journal={ArXiv},
  year={2024},
  volume={abs/2410.16184},
  url={https://api.semanticscholar.org/CorpusID:273507377}
}

@article{Schulhoff2024ThePR,
  title={The Prompt Report: A Systematic Survey of Prompting Techniques},
  author={Sander Schulhoff and Michael Ilie and Nishant Balepur and Konstantine Kahadze and Amanda Liu and Chenglei Si and Yinheng Li and Aayush Gupta and HyoJung Han and Sevien Schulhoff and Pranav Sandeep Dulepet and Saurav Vidyadhara and Dayeon Ki and Sweta Agrawal and Chau Minh Pham and Gerson C. Kroiz and Feileen Li and Hudson Tao and Ashay Srivastava and Hevander Da Costa and Saloni Gupta and Megan L. Rogers and Inna Goncearenco and Giuseppe Sarli and Igor Galynker and Denis Peskoff and Marine Carpuat and Jules White and Shyamal Anadkat and Alexander Miserlis Hoyle and Philip Resnik},
  journal={ArXiv},
  year={2024},
  volume={abs/2406.06608},
  url={https://api.semanticscholar.org/CorpusID:270380093}
}

@article{Gou2020KnowledgeDA,
  title={Knowledge Distillation: A Survey},
  author={Jianping Gou and B. Yu and Stephen J. Maybank and Dacheng Tao},
  journal={International Journal of Computer Vision},
  year={2020},
  volume={129},
  pages={1789 - 1819},
  url={https://api.semanticscholar.org/CorpusID:219559263}
}

@article{Shneiderman1984ResponseTA,
  title={Response time and display rate in human performance with computers},
  author={Ben Shneiderman},
  journal={ACM Comput. Surv.},
  year={1984},
  volume={16},
  pages={265-285},
  url={https://api.semanticscholar.org/CorpusID:445900}
}

@article{Kantharuban2024StereotypeOP,
  title={Stereotype or Personalization? User Identity Biases Chatbot Recommendations},
  author={Anjali Kantharuban and Jeremiah Milbauer and Emma Strubell and Graham Neubig},
  journal={ArXiv},
  year={2024},
  volume={abs/2410.05613},
  url={https://api.semanticscholar.org/CorpusID:273228304}
}

@article{Bai2022TrainingAH,
  title={Training a Helpful and Harmless Assistant with Reinforcement Learning from Human Feedback},
  author={Yuntao Bai and Andy Jones and Kamal Ndousse and Amanda Askell and Anna Chen and Nova Dassarma and Dawn Drain and Stanislav Fort and Deep Ganguli and T. J. Henighan and Nicholas Joseph and Saurav Kadavath and John Kernion and Tom Conerly and Sheer El-Showk and Nelson Elhage and Zac Hatfield-Dodds and Danny Hernandez and Tristan Hume and Scott Johnston and Shauna Kravec and Liane Lovitt and Neel Nanda and Catherine Olsson and Dario Amodei and Tom B. Brown and Jack Clark and Sam McCandlish and Chris Olah and Benjamin Mann and Jared Kaplan},
  journal={ArXiv},
  year={2022},
  volume={abs/2204.05862},
  url={https://api.semanticscholar.org/CorpusID:248118878}
}

@article{Shneiderman1983DirectMA,
  title={Direct Manipulation: A Step Beyond Programming Languages},
  author={Ben Shneiderman},
  journal={Computer},
  year={1983},
  volume={16},
  pages={57-69},
  url={https://api.semanticscholar.org/CorpusID:14942172}
}

@article{Lee2004TrustIA,
  title={Trust in Automation: Designing for Appropriate Reliance},
  author={John D. Lee and Katrina A. See},
  journal={Human Factors: The Journal of Human Factors and Ergonomics Society},
  year={2004},
  volume={46},
  pages={50 - 80},
  url={https://api.semanticscholar.org/CorpusID:5210390}
}

@article{Zhang2024SeeWT,
  title={See Widely, Think Wisely: Toward Designing a Generative Multi-agent System to Burst Filter Bubbles},
  author={Yu Zhang and Jingwei Sun and Li Feng and Cen Yao and Mingming Fan and Liuxin Zhang and Qianying Wang and Xin Geng and Yong Rui},
  journal={Proceedings of the 2024 CHI Conference on Human Factors in Computing Systems},
  year={2024},
  url={https://api.semanticscholar.org/CorpusID:269746858}
}

@inproceedings{yuan-etal-2025-personalized,
    title = "Personalized Large Language Model Assistant with Evolving Conditional Memory",
    author = "Yuan, Ruifeng  and
      Sun, Shichao  and
      Li, Yongqi  and
      Wang, Zili  and
      Cao, Ziqiang  and
      Li, Wenjie",
    editor = "Rambow, Owen  and
      Wanner, Leo  and
      Apidianaki, Marianna  and
      Al-Khalifa, Hend  and
      Eugenio, Barbara Di  and
      Schockaert, Steven",
    booktitle = "Proceedings of the 31st International Conference on Computational Linguistics",
    month = jan,
    year = "2025",
    address = "Abu Dhabi, UAE",
    publisher = "Association for Computational Linguistics",
    url = "https://aclanthology.org/2025.coling-main.254/",
    pages = "3764--3777"
}

@inproceedings{dror-etal-2018-hitchhikers,
    title = "The Hitchhiker{'}s Guide to Testing Statistical Significance in Natural Language Processing",
    author = "Dror, Rotem  and
      Baumer, Gili  and
      Shlomov, Segev  and
      Reichart, Roi",
    editor = "Gurevych, Iryna  and
      Miyao, Yusuke",
    booktitle = "Proceedings of the 56th Annual Meeting of the Association for Computational Linguistics (Volume 1: Long Papers)",
    month = jul,
    year = "2018",
    address = "Melbourne, Australia",
    publisher = "Association for Computational Linguistics",
    url = "https://aclanthology.org/P18-1128/",
    doi = "10.18653/v1/P18-1128",
    pages = "1383--1392"
}

@article{Liu2025CSPaperSumAL,
  title={CS-PaperSum: A Large-Scale Dataset of AI-Generated Summaries for Scientific Papers},
  author={Javin Liu and Aryan Vats and Zihao He},
  journal={ArXiv},
  year={2025},
  volume={abs/2502.20582},
  url={https://api.semanticscholar.org/CorpusID:276724947}
}

@inproceedings{Wang2024FactualityOL,
  title={Factuality of Large Language Models: A Survey},
  author={Yuxia Wang and Minghan Wang and Muhammad Arslan Manzoor and Fei Liu and Georgi N. Georgiev and Rocktim Jyoti Das and Preslav Nakov},
  booktitle={Conference on Empirical Methods in Natural Language Processing},
  year={2024},
  url={https://api.semanticscholar.org/CorpusID:267412450}
}

@inproceedings{Joshi2025ELIWhyET,
  title={ELI-Why: Evaluating the Pedagogical Utility of Language Model Explanations},
  author={Brihi Joshi and Keyu He and Sahana Ramnath and Sadra Sabouri and Kaitlyn Zhou and Souti Chattopadhyay and Swabha Swayamdipta and Xiang Ren},
  booktitle={Annual Meeting of the Association for Computational Linguistics},
  year={2025},
  url={https://api.semanticscholar.org/CorpusID:279410747}
}

@inproceedings{10.1145/1357054.1357161,
author = {Zhang, Xiaolong and Qu, Yan and Giles, C. Lee and Song, Piyou},
title = {CiteSense: supporting sensemaking of research literature},
year = {2008},
isbn = {9781605580111},
publisher = {Association for Computing Machinery},
address = {New York, NY, USA},
url = {https://doi.org/10.1145/1357054.1357161},
doi = {10.1145/1357054.1357161},
booktitle = {Proceedings of the SIGCHI Conference on Human Factors in Computing Systems},
pages = {677–680},
numpages = {4},
location = {Florence, Italy},
series = {CHI '08}
}

@inproceedings{Danks1984ProtocolAV,
  title={Protocol Analysis: Verbal Reports as Data},
  author={Joseph H. Danks and K. Anders Ericsson and Herbert A. Simon},
  year={1984},
  url={https://api.semanticscholar.org/CorpusID:59628641}
}

@inproceedings{Nielsen1993UsabilityE,
  title={Usability engineering},
  author={Jakob Nielsen},
  booktitle={The Computer Science and Engineering Handbook},
  year={1993},
  url={https://api.semanticscholar.org/CorpusID:260423343}
}

@inproceedings{Venkit2025DeepTRACEAD,
  title={DeepTRACE: Auditing Deep Research AI Systems for Tracking Reliability Across Citations and Evidence},
  author={Pranav Narayanan Venkit and Philippe Laban and Yilun Zhou and Kung-Hsiang Huang and Yixin Mao and Chien-Sheng Wu},
  year={2025},
  url={https://api.semanticscholar.org/CorpusID:281194921}
}

@article{NarayananVenkit2025SearchEI,
  title={Search Engines in the AI Era: A Qualitative Understanding to the False Promise of Factual and Verifiable Source-Cited Responses in LLM-based Search},
  author={Pranav Narayanan Venkit and Philippe Laban and Yilun Zhou and Yixin Mao and Chien-Sheng Wu},
  journal={Proceedings of the 2025 ACM Conference on Fairness, Accountability, and Transparency},
  year={2025},
  url={https://api.semanticscholar.org/CorpusID:279469433}
}

@article{Zhang2013ToPO,
  title={To personalize or not: a risk management perspective},
  author={Weinan Zhang and Jun Wang and Bowei Chen and Xiaoxue Zhao},
  journal={Proceedings of the 7th ACM conference on Recommender systems},
  year={2013},
  url={https://api.semanticscholar.org/CorpusID:571694}
}

@article{Han2022SplitGPAB,
  title={SplitGP: Achieving Both Generalization and Personalization in Federated Learning},
  author={Dong-Jun Han and Do-Yeon Kim and Minseok Choi and Christopher G. Brinton and Jaekyun Moon},
  journal={IEEE INFOCOM 2023 - IEEE Conference on Computer Communications},
  year={2022},
  pages={1-10},
  url={https://api.semanticscholar.org/CorpusID:254823183}
}

@article{Comanici2025Gemini2P,
  title={Gemini 2.5: Pushing the Frontier with Advanced Reasoning, Multimodality, Long Context, and Next Generation Agentic Capabilities},
  author={Gheorghe Comanici and others},
  journal={ArXiv},
  year={2025},
  volume={abs/2507.06261},
  url={https://api.semanticscholar.org/CorpusID:280151524}
}

@inproceedings{joachims2002optimizing,
  title={Optimizing search engines using clickthrough data},
  author={Joachims, Thorsten},
  booktitle={Proceedings of the eighth ACM SIGKDD international conference on Knowledge discovery and data mining},
  pages={133--142},
  year={2002}
}

@misc{srikanth2026discotracerepresentingcomparinganswering,
      title={DiscoTrace: Representing and Comparing Answering Strategies of Humans and LLMs in Information-Seeking Question Answering}, 
      author={Neha Srikanth and Jordan Boyd-Graber and Rachel Rudinger},
      year={2026},
      eprint={2604.15140},
      archivePrefix={arXiv},
      primaryClass={cs.CL},
      url={https://arxiv.org/abs/2604.15140}, 
}

@inproceedings{mondal-etal-2024-presentations,
    title = "Presentations by the Humans and For the Humans: Harnessing {LLM}s for Generating Persona-Aware Slides from Documents",
    author = "Mondal, Ishani  and
      S, Shwetha  and
      Natarajan, Anandhavelu  and
      Garimella, Aparna  and
      Bandyopadhyay, Sambaran  and
      Boyd-Graber, Jordan",
    editor = "Graham, Yvette  and
      Purver, Matthew",
    booktitle = "Proceedings of the 18th Conference of the European Chapter of the Association for Computational Linguistics (Volume 1: Long Papers)",
    month = mar,
    year = "2024",
    address = "St. Julian{'}s, Malta",
    publisher = "Association for Computational Linguistics",
    url = "https://aclanthology.org/2024.eacl-long.163/",
    doi = "10.18653/v1/2024.eacl-long.163",
    pages = "2664--2684"
}

@inproceedings{balepur2025these,
  title={Which of these best describes multiple choice evaluation with llms? a) forced b) flawed c) fixable d) all of the above},
  author={Balepur, Nishant and Rudinger, Rachel and Boyd-Graber, Jordan Lee},
  booktitle={Proceedings of the 63rd Annual Meeting of the Association for Computational Linguistics (Volume 1: Long Papers)},
  pages={3394--3418},
  year={2025}
}

@inproceedings{balepur-etal-2025-whose,
    title = "Whose Boat Does it Float? Improving Personalization in Preference Tuning via Inferred User Personas",
    author = "Balepur, Nishant  and
      Padmakumar, Vishakh  and
      Yang, Fumeng  and
      Feng, Shi  and
      Rudinger, Rachel  and
      Boyd-Graber, Jordan Lee",
    editor = "Che, Wanxiang  and
      Nabende, Joyce  and
      Shutova, Ekaterina  and
      Pilehvar, Mohammad Taher",
    booktitle = "Proceedings of the 63rd Annual Meeting of the Association for Computational Linguistics (Volume 1: Long Papers)",
    month = jul,
    year = "2025",
    address = "Vienna, Austria",
    publisher = "Association for Computational Linguistics",
    url = "https://aclanthology.org/2025.acl-long.168/",
    doi = "10.18653/v1/2025.acl-long.168",
    pages = "3371--3393",
    ISBN = "979-8-89176-251-0"
}

@inproceedings{Kryscinski2019EvaluatingTF,
  title={Evaluating the Factual Consistency of Abstractive Text Summarization},
  author={Wojciech Kryscinski and Bryan McCann and Caiming Xiong and Richard Socher},
  booktitle={Conference on Empirical Methods in Natural Language Processing},
  year={2019},
  url={https://api.semanticscholar.org/CorpusID:204976362}
}

@article{Zheng2023JudgingLW,
  title={Judging LLM-as-a-judge with MT-Bench and Chatbot Arena},
  author={Lianmin Zheng and Wei-Lin Chiang and Ying Sheng and Siyuan Zhuang and Zhanghao Wu and Yonghao Zhuang and Zi Lin and Zhuohan Li and Dacheng Li and Eric P. Xing and Haotong Zhang and Joseph E. Gonzalez and Ion Stoica},
  journal={ArXiv},
  year={2023},
  volume={abs/2306.05685},
  url={https://api.semanticscholar.org/CorpusID:259129398}
}

@article{Qin2024InFoBenchEI,
  title={InFoBench: Evaluating Instruction Following Ability in Large Language Models},
  author={Yiwei Qin and Kaiqiang Song and Yebowen Hu and Wenlin Yao and Sangwoo Cho and Xiaoyang Wang and Xuansheng Wu and Fei Liu and Pengfei Liu and Dong Yu},
  journal={ArXiv},
  year={2024},
  volume={abs/2401.03601},
  url={https://api.semanticscholar.org/CorpusID:266844311}
}

@article{Muennighoff2024GenerativeRI,
  title={Generative Representational Instruction Tuning},
  author={Niklas Muennighoff and Hongjin Su and Liang Wang and Nan Yang and Furu Wei and Tao Yu and Amanpreet Singh and Douwe Kiela},
  journal={ArXiv},
  year={2024},
  volume={abs/2402.09906},
  url={https://api.semanticscholar.org/CorpusID:267681873}
}

@article{Salemi2025LaMPQAAB,
  title={LaMP-QA: A Benchmark for Personalized Long-form Question Answering},
  author={Alireza Salemi and Hamed Zamani},
  journal={ArXiv},
  year={2025},
  volume={abs/2506.00137},
  url={https://api.semanticscholar.org/CorpusID:279075447}
}

@article{Salemi2023LaMPWL,
  title={LaMP: When Large Language Models Meet Personalization},
  author={Alireza Salemi and Sheshera Mysore and Michael Bendersky and Hamed Zamani},
  journal={ArXiv},
  year={2023},
  volume={abs/2304.11406},
  url={https://api.semanticscholar.org/CorpusID:258298303}
}

@article{Lin2024PaperCA,
  title={Paper Copilot: A Self-Evolving and Efficient LLM System for Personalized Academic Assistance},
  author={Guanyu Lin and Tao Feng and Pengrui Han and Ge Liu and Jiaxuan You},
  journal={ArXiv},
  year={2024},
  volume={abs/2409.04593},
  url={https://api.semanticscholar.org/CorpusID:272524265}
}

@article{Kim2025IntentFlowIS,
  title={IntentFlow: Interactive Support for Communicating Intent with LLMs in Writing Tasks},
  author={Yoonsu Kim and Brandon Chin and Kihoon Son and Seoyoung Kim and Juho Kim},
  journal={ArXiv},
  year={2025},
  volume={abs/2507.22134},
  url={https://api.semanticscholar.org/CorpusID:280391826}
}

@inproceedings{
zhang2025modeling,
title={Modeling Future Conversation Turns to Teach {LLM}s to Ask Clarifying Questions},
author={Michael JQ Zhang and W. Bradley Knox and Eunsol Choi},
booktitle={The Thirteenth International Conference on Learning Representations},
year={2025},
url={https://openreview.net/forum?id=cwuSAR7EKd}
}

@article{Gao2024EndtoendTF,
  title={End-to-end Training for Recommendation with Language-based User Profiles},
  author={Zhaolin Gao and Joyce Zhou and Yijia Dai and Thorsten Joachims},
  journal={ArXiv},
  year={2024},
  volume={abs/2410.18870},
  url={https://api.semanticscholar.org/CorpusID:273549728}
}

@article{Shaikh2025CreatingGU,
  title={Creating General User Models from Computer Use},
  author={Omar Shaikh and Shardul Sapkota and Shan Rizvi and Eric Horvitz and Joon Sung Park and Diyi Yang and Michael S. Bernstein},
  journal={ArXiv},
  year={2025},
  volume={abs/2505.10831},
  url={https://api.semanticscholar.org/CorpusID:278715109}
}

@article{Garbacea2025HyPerAlignIP,
  title={HyPerAlign: Interpretable Personalized LLM Alignment via Hypothesis Generation},
  author={Cristina Garbacea and Chenhao Tan},
  journal={ArXiv},
  year={2025},
  volume={abs/2505.00038},
  url={https://api.semanticscholar.org/CorpusID:278769706}
}

@article{Tang2024StepBackPD,
  title={Step-Back Profiling: Distilling User History for Personalized Scientific Writing},
  author={Xiangru Tang and Xingyao Zhang and Yanjun Shao and Jie Wu and Yilun Zhao and Arman Cohan and Ming Gong and Dongmei Zhang and Mark Gerstein},
  journal={ArXiv},
  year={2024},
  volume={abs/2406.14275},
  url={https://api.semanticscholar.org/CorpusID:270619816}
}

@article{Liu2024FilteringDR,
  title={Filtering Discomforting Recommendations with Large Language Models},
  author={Jiahao Liu and Yiyang Shao and Peng Zhang and Dongsheng Li and Hansu Gu and Chao Chen and Longzhi Du and Tun Lu and Ning Gu},
  journal={Proceedings of the ACM on Web Conference 2025},
  year={2024},
  url={https://api.semanticscholar.org/CorpusID:273228188}
}

@article{Saxon2024BenchmarksAM,
  title={Benchmarks as Microscopes: A Call for Model Metrology},
  author={Michael Stephen Saxon and Ari Holtzman and Peter West and William Yang Wang and Naomi Saphra},
  journal={ArXiv},
  year={2024},
  volume={abs/2407.16711},
  url={https://api.semanticscholar.org/CorpusID:271404269}
}

@article{august2023paper,
  title={Paper plain: Making medical research papers approachable to healthcare consumers with natural language processing},
  author={August, Tal and Wang, Lucy Lu and Bragg, Jonathan and Hearst, Marti A and Head, Andrew and Lo, Kyle},
  journal={ACM Transactions on Computer-Human Interaction},
  volume={30},
  number={5},
  pages={1--38},
  year={2023},
  publisher={ACM New York, NY}
}

@article{Jiang2025KnowMR,
  title={Know Me, Respond to Me: Benchmarking LLMs for Dynamic User Profiling and Personalized Responses at Scale},
  author={Bowen Jiang and Zhuoqun Hao and Young-Min Cho and Bryan Li and Yuan Yuan and Sihao Chen and Lyle Ungar and Camillo Jose Taylor and Dan Roth},
  journal={ArXiv},
  year={2025},
  volume={abs/2504.14225},
  url={https://api.semanticscholar.org/CorpusID:277955197}
}

@article{liao2024llms,
  title={LLMs as Research Tools: A Large Scale Survey of Researchers' Usage and Perceptions},
  author={Liao, Zhehui and Antoniak, Maria and Cheong, Inyoung and Cheng, Evie Yu-Yen and Lee, Ai-Heng and Lo, Kyle and Chang, Joseph Chee and Zhang, Amy X},
  journal={arXiv preprint arXiv:2411.05025},
  year={2024}
}

@inproceedings{Hutchinson2003TechnologyPI,
  title={Technology probes: inspiring design for and with families},
  author={Hilary Browne Hutchinson and Wendy E. Mackay and Bo Westerlund and Benjamin B. Bederson and Allison Druin and Catherine Plaisant and Michel Beaudouin-Lafon and St{\'e}phane Conversy and Helen Evans and Heiko Hansen and Nicolas Roussel and Bj{\"o}rn Eiderb{\"a}ck},
  booktitle={International Conference on Human Factors in Computing Systems},
  year={2003},
  url={https://api.semanticscholar.org/CorpusID:1356747}
}

@inproceedings{Boyatzis1998TransformingQI,
  title={Transforming Qualitative Information: Thematic Analysis and Code Development},
  author={Richard E. Boyatzis},
  year={1998}
}

@inproceedings{voorhees1999trec,
  title={The trec-8 question answering track report.},
  author={Voorhees, Ellen M and others},
  booktitle={Trec},
  volume={99},
  pages={77--82},
  year={1999}
}

@inproceedings{zamfirescu2023johnny,
  title={Why Johnny can’t prompt: how non-AI experts try (and fail) to design LLM prompts},
  author={Zamfirescu-Pereira, J Diego and Wong, Richmond Y and Hartmann, Bjoern and Yang, Qian},
  booktitle={Proceedings of the 2023 CHI conference on human factors in computing systems},
  pages={1--21},
  year={2023}
}
\bibliographystyle{acl_natbib}

\appendix \label{section:appendix}

\clearpage

\section{Appendix}

\subsection{Offline Dataset Details} \label{appendix:dataset}

The \textsc{AstaBench}, ScholarQA-CS2 \cite{bragg2025astabench} and CS-PaperSum \cite{Liu2025CSPaperSumAL} datasets we build off of on are publicly accessible and used within their intended use.
To retrieve the full text of papers for offline experiments, we use an internal database to our organization that already collected them; in our interface to handle new papers, we use the Semantic Scholar Snippets API\footnote{https://www.semanticscholar.org/product/api}.
Our dataset is in English and has no PII that is not already publicly accessible (i.e. information present in a research paper).
We summarize our dataset in Table~\ref{appendix:table:full_dataset}.

\subsection{Offline Metric Human Agreement} \label{appendix:metrics}

To develop offline metrics (\cref{subsection:metrics}) we created an initial version of our metrics and ran it on a subset of our dev set; one author then blindly gave ratings on 30--50 examples---half where the model predicted $1$ and half where the model predicted $0$---for agreement.
For specificity, which uses a 1--5 Likert rating, we sample $10$ examples over each predicted rating. 
If the metric had agreement $<0.7$, we modify our prompts based on model errors, and repeat the process until we reach substantial agreement; this process typically took two rounds per metric.

For profiles, our agreements are: $88.3\%$ for inference accuracy, $88.8\%$ for relevance, $96.7\%$ for category accuracy, and $0.66$ Person's correlation for specificity.
For actions, our agreements are: $93.3\%$ for personalization win rate and $82.0\%$ for relevance.
For reports, our agreement is: $86\%$ for action adherence; the other report metrics were already tested by \citet{bragg2025astabench}.
Appendix~\ref{appendix:prompt_outputs} contains the prompts used for metrics.

\subsection{Offline Experiment Setup Details} \label{appendix:setup}

To create profiles and actions, we use $1.0$ temperature and a max token length of $40960$. All commercial \mm{}s are accessed via their original providers; we access DeepSeek models from TogetherAI.\footnote{https://www.together.ai/}

For generating reports, we use the original hyperparemeters for each model; \model{} uses \scholarqa{}'s hyperparameters \cite{singh-etal-2025-ai2}.
For a fairer comparison, we let \textsc{OpenScholar} \cite{Asai2024OpenScholarSS} use Claude over its trained model, just like \model{}, and let \storm{} \cite{Shao2024AssistingIW} use Semantic Scholar as its retriever versus Google.
Perplexity and OpenAI do not directly provide snippets of the sources in the web pages that they use, so we prompt these models to extract them from the web pages; these may be fabricated,\footnote{A manual spot check showed models were pulling quotes from real web pages.} but we~take them in good faith so models have the best possible chance.
Report evaluation uses InspectAI\footnote{https://github.com/UKGovernmentBEIS/inspect\_ai} from \textsc{AstaBench} which needs a specific JSON format.
We prompt Gemini to parse OpenAI's/Perplexity' reports into said format, like \citet{bragg2025astabench}.

We allocate $72$ CPU hours for each experiment run. All results are from a single run.

\subsection{Queries versus Personalized Actions} \label{appendix:tension} 

In \cref{subsection:offline_results}, we show personalized actions are less relevant to the query than generic ones.
To learn why, we run the metric segmented by the similarity of the author to the query (i.e. low, medium, high).
We consistently find personalized actions~have more conflicts for low author similarities; for example, DeepSeek's relevance is $0.765$, $0.72$, $0.69$ for high, medium, and low similarity authors, respectively.
We believe such tension in queries and personalization is natural; for example, an author working on reasoning \mm{}s may see less benefit from personalization when asking about \nlp{}+education.

Overall, as query conflicts are infrequent and our main goal was to see if this was truly a problem via online evaluations (\cref{section:online}), we proceeded anyway.

\subsection{Model Ablations} \label{appendix:ablations} 

We now ablate our design choices in \model{}.
We first consider the best place to show \model{} the actions $\mathcal{A}$ in the execution prompts (Table~\ref{appendix:ablation:strategy}): at all execution prompts, just the individual actions that are relevant (e.g. ``Search for papers'' actions only get shown in retrieval), the individual actions that are relevant and all prior actions, or ignoring multi-step execution entirely and generating the report in a single prompt; the former achieves the highest report quality and action adherence.

We then test \mm{} to power \model{} (Table~\ref{appendix:ablation:llm}): Claude-4 Sonnet, Gemini 2.5 Flash, and GPT-4.1; all have similar scores, so we use Claude-4 to match \citet{singh-etal-2025-ai2} in \scholarqa{}.
Finally, we test how many actions \model{} can use (Table~\ref{appendix:ablation:steps}); on 4--30 steps, \model{} maintains report quality with only minor drops in action adherence scores.

\subsection{Simulation Prompt Variations} \label{appendix:simulation} 

\begin{figure*}[ht]
    \centering
\includegraphics[width=\linewidth]{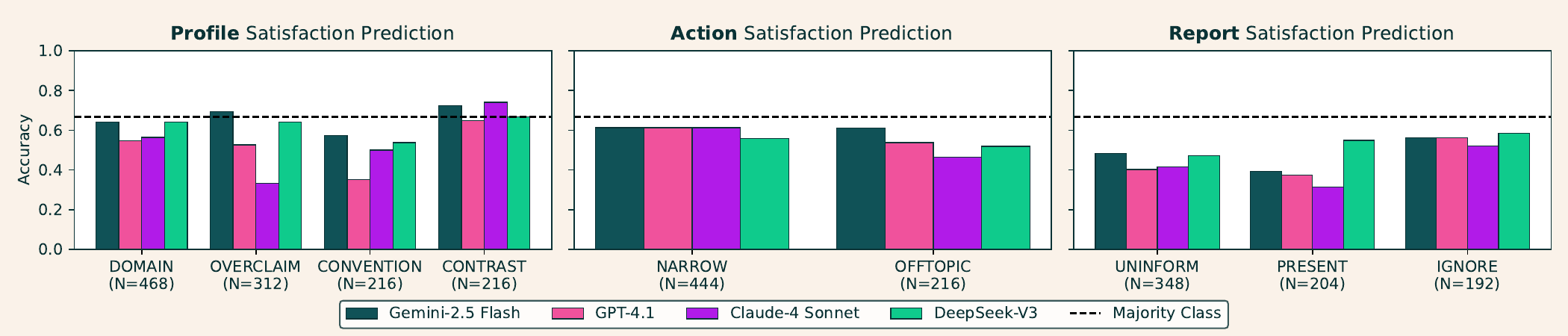}
    \caption{\label{fig:simulation_one_example} Accuracy of \mm{} judges for predicting user satisfaction in personalization aspects over profiles, actions, and reports with \textbf{one few-shot example}. \mm{} judges  are still worse than the majority class baseline.}
\end{figure*}

\begin{figure*}[ht]
    \centering
\includegraphics[width=\linewidth]{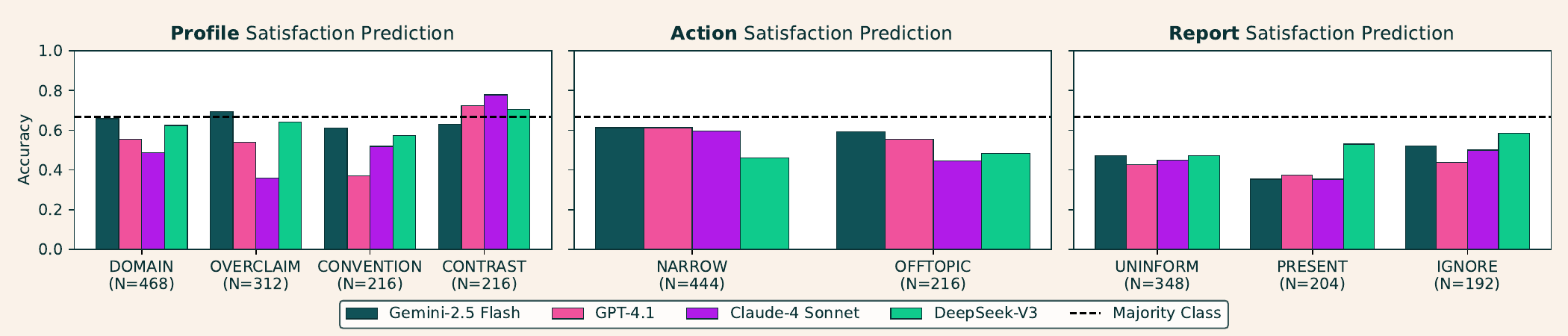}
    \caption{\label{fig:simulation_no_examples} Accuracy of \mm{} judges for predicting user satisfaction in personalization aspects over profiles, actions, and reports with \textbf{no examples}. \mm{} judges  are still worse than the majority class baseline.}
\end{figure*}

To further study how \mm{} judges predict user satisfaction, we test several prompt variations.
We first decrease the number of few-shot examples from six (Figure~\ref{fig:simulation}) to three (Figure~\ref{fig:simulation_one_example}) and zero (Figure~\ref{fig:simulation_no_examples}).
\mm{} judges are still unreliable and adding exemplars does not boost accuracy, suggesting off-the-shelf \mm{} judges struggle with this task regardless of the number of few-shot examples; we suggest future work to train custom reward models for user satisfaction with real user data if they want to better simulate this offline \cite{lambert-etal-2025-rewardbench}.

\begin{figure*}[ht]
    \centering
\includegraphics[width=\linewidth]{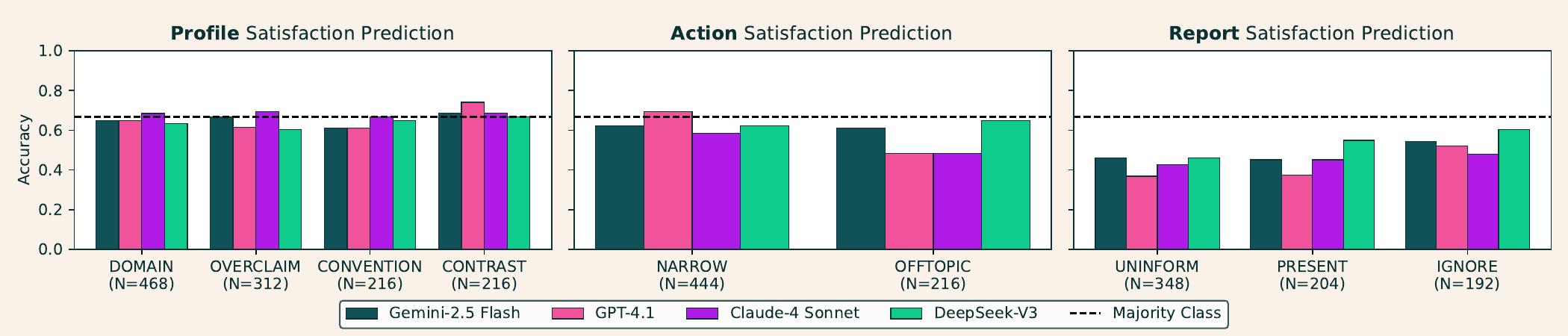}
    \caption{\label{fig:simulation_no_metrics} Accuracy of \mm{} judges for predicting user satisfaction in personalization aspects over profiles, actions, and reports with \textbf{no metric definitions}. \mm{} judges are still worse than the majority class baseline.}
\end{figure*}

Lastly, we test if our metric definitions distract \mm{} judges (e.g. ``Would the user be satisfied by the terms in this profile inference?'') and would~be much more accurate if they directly predict user~satisfaction (i.e. ``Would the user be satisfied with this profile inference?'').
Figure~\ref{fig:simulation_no_metrics} shows removing metric definitions does not boost \mm{} judge accuracy.
Appendix~\ref{appendix:prompt_outputs} has prompts used in the experiment.

\subsection{Formative Study Details} \label{appendix:formative}

\begin{figure*}[ht]
    \centering
    \fbox{\includegraphics[width=\linewidth]{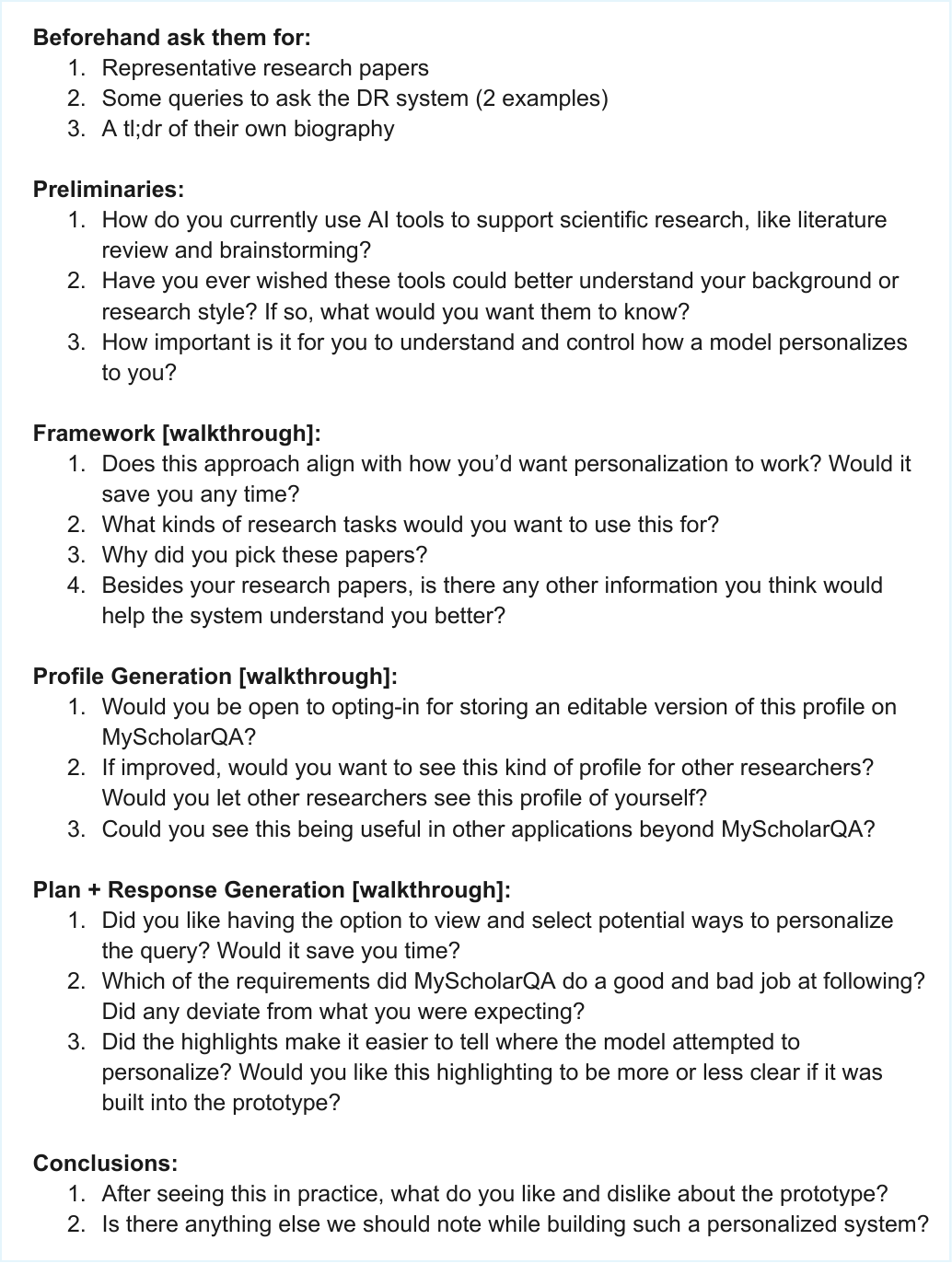}}
    \caption{List of questions we ask participants in the formative. Sections marked with "[walkthrough]" indicates that participants also gave feedback on model outputs or high-level design in addition to answering the questions. During the study, we refer to the list of actions as ``plans'', which users found clear. \label{fig:formative_questions} }
\end{figure*}

We reach out to candidates for our formative study via email from an internal mailing list of \dr{} users (more de-anonymized details will be released post-acceptance).
We confirm they are active \dr{} users over email.
Three participants were M.S. students and two were Ph.D. students in North America, Europe, and Asia.
Each interview was conducted in English and all participants had fluency in English.
In Figure~\ref{fig:formative_questions}, we provide all questions we ask participants in the formative study (\cref{subsection:formative}).

\subsection{Interview Recruitment Protocol} \label{appendix:recruitment}

This section provides more details on the protocol for our 90-minute usability studies, including annotator instructions (Appendix~\ref{appendix:interview_instructions}), Recruitment (Appendix~\ref{appendix:interview_recruitment}), Consent (Appendix~\ref{appendix:interview_consent}), and Demographics (Appendix~\ref{appendix:interview_demographics}).

\subsubsection{Annotator Instructions} \label{appendix:interview_instructions}

We first asked each participant to fill out a Google Form to gain insights on their knowledge and proficiency with deep research systems (\cref{appendix:survey}). After reviewing their responses, we scheduled a 90 minute meeting with each participant to walk them through example annotations and understand their thought process. We asked participants to provide the following in advance of the meeting:
\begin{enumerate}
    \item A set of 5 research papers on semantic scholar (the paper URLs) that you feel best represent your research interests. These are likely papers you have written, wish you had written, are relevant to a project you are working on, or have been highly influential to you as a researcher. Please make sure these are open-access PDFs! (i.e. the semantic scholar link will have the button "[PDF] Semantic Scholar" if it's valid). The should be in the form "{url}/paper/{title}/{id}" on Semantic Scholar
    \item Three research queries that you would be interested in asking our Deep Research System. These do not have to be about the papers you have selected.
    \item Create a new Gmail account with the following credentials: a) Username: [provided email], b) Password: [ anything ], c) Name: [ anything but your own name]. You do not need to use this email to join the Google Meet, but you will need it to log into our interface. You will use this account for both the pilot and any future tasks, and this will ensure no personally identifiable information is shared. You do not need to share your password, but please ensure that you use the username assigned to you.
\end{enumerate}

We did not collect any PII from participants (all information collected is publicly available), and the study does not include any risks.
Upon recruitment, we provided additional guidelines/task instructions for query selection and annotation (plan selection, rating responses, feedback, etc.).

\subsubsection{Recruitment} \label{appendix:interview_recruitment}

We recruited participants via a job post on Upwork.\footnote{https://www.upwork.com/} In the job post, we provided a description of the job, screening/pilot process, target domain (CS), estimated weekly time commitment (less than 30 hr/week for 1-3 months), compensation range (USD \$30-40/hr), project agreement terms, and screening questions. The compensation range was determined based on previous projects involving annotators with similar backgrounds and which in turn is based on the typical hourly rate for annotation tasks requiring SME (>\$30). We did not explicitly state our target education level (PhDs and PhD candidates) in the job posting, but asked for this as a screening question. Candidates submitted proposals (cover letter, answers to screening questions, and desired compensation) which we reviewed, used to shortlist them, and send them an offer based on their proposal. After review, we scheduled paid 90-minute interviews for shortlisted candidates. Compensation was determined by the hours logged by each participant on a weekly basis --- regardless of whether we decided to move forward with a candidate’s application, we asked them to log hours for their time (form and interview) so that we could compensate them.

\subsubsection{Consent} \label{appendix:interview_consent}

Our Upwork job description/offer provided a clear description of the task and what kind of data would be collected. Furthermore, the offer contained a project participation agreement and along with a statement that by accepting the offer, participants also implicitly accepted the terms of the attached agreement. The participation agreement is used for all of the organization’s Upwork annotation projects and explains in detail what kind of data may be collected and how it may be used.

\subsubsection{Demographics} \label{appendix:interview_demographics}

Participants in our interviews were from varied countries in North America, Europe, and Asia with varying educational backgrounds---some were enrolled in MS/Ph.D. programs, others had already obtained their Ph.D., and some were already working in research-focused industry positions.
All had expertise in various areas of computer science, including machine learning, computer vision, natural language processing, security, and AI+Society.
Each interview was conducted in English and all participants had fluency in English.
Our data will not release any protected demographic information.

\subsection{Participant Survey Results} \label{appendix:survey}

To ensure participants are active \dr{} users, we first survey their \dr{} usage; $56.6\%$ use \dr{} daily, $26.1\%$ a few times a week, and others a few times a month. 
Most participants use OpenAI, Gemini, or Perplexity; only one used \scholarqa{}, reducing the risk of overly-positive feedback on \model{}. 
All but one participant valued \dr{} having more knowledge about them, which they felt could best be uncovered via their authored/read papers and documentation for projects. 
Most wanted the system to adapt over time and tailor knowledge per query.

Participants wanted to control all execution steps of \dr (paper search, section planning, generation), preferring picking among actions and flagging low-quality queries as forms of control.
Less than half wanted to answer follow-up questions to control the system, despite their prominence in OpenAI/Gemini. 
Most \dr{} use cases spanned learning, writing, experimentation, literature review, and ideation.

After acceptance, we will provide a link to the survey questions and responses (de-anonymized).

\subsection{Full Interface Screenshots} \label{appendix:interface}

\begin{figure*}[ht]
    \centering
\fbox{\includegraphics[width=\linewidth]{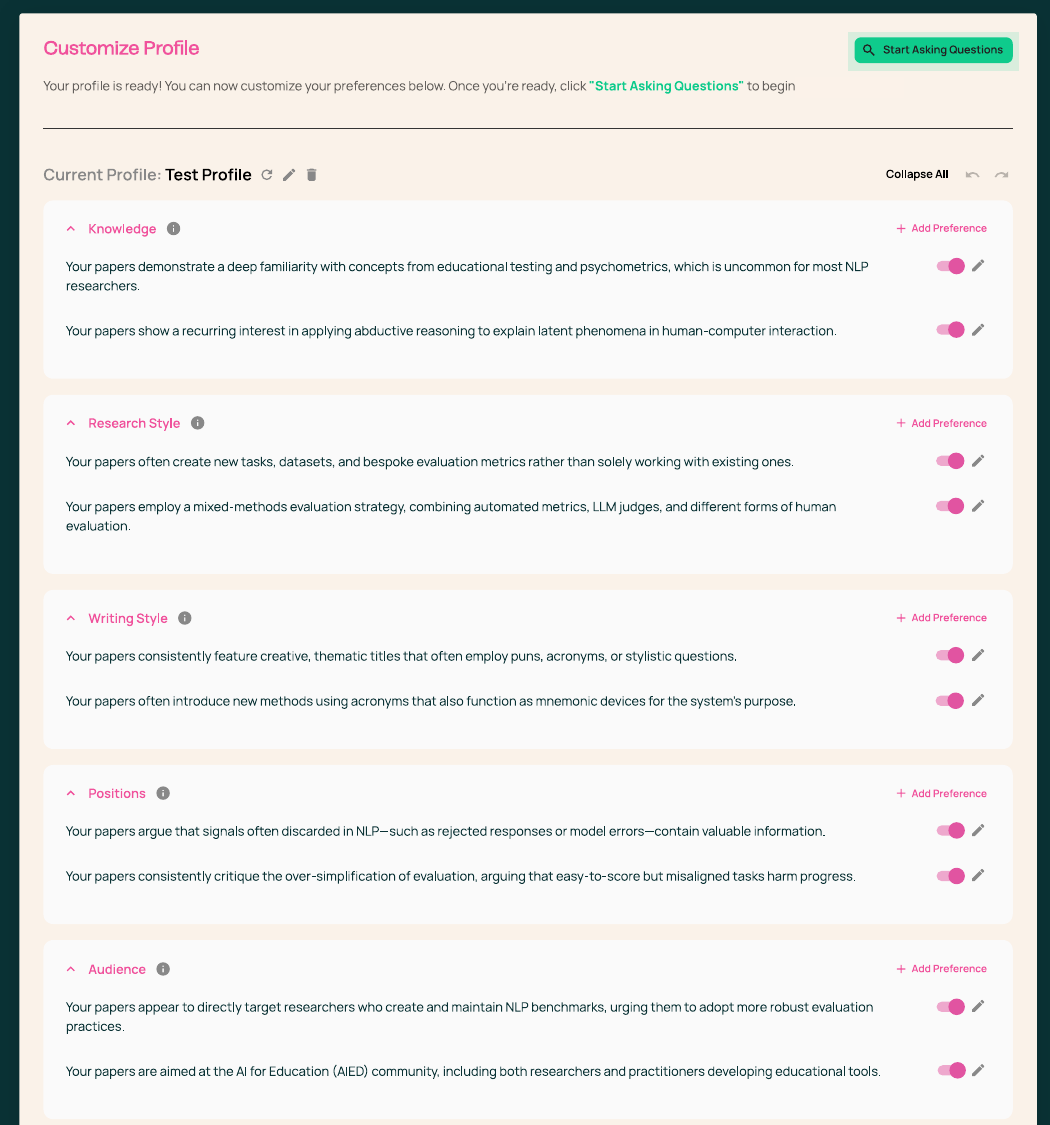}}
    \caption{\label{appendix:fig:full_profile} Full example of one of the author's profiles with two inferences per section in \model{}. We remove the paper citations after each profile inference for anonymity. Users can view, toggle, and edit their profile inferences.}
\end{figure*}

\begin{figure*}[ht]
    \centering
\fbox{\includegraphics[width=\linewidth]{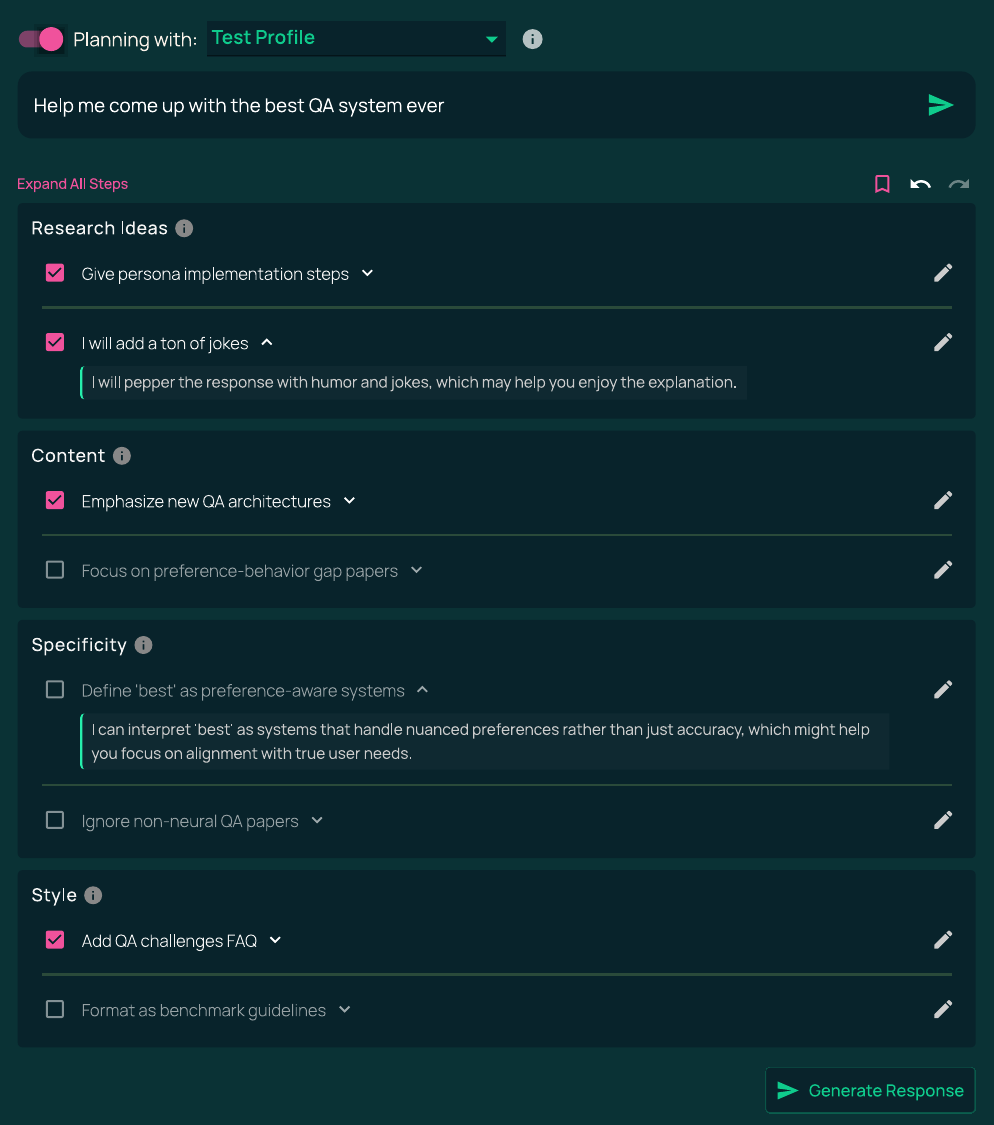}}
    \caption{\label{appendix:fig:full_plan} Full example of actions for a query in \model{}. Users can view, toggle, and edit actions in the list. Clicking on the dropdown arrow gives a more elaborate description of each action.}
\end{figure*}

\begin{figure*}[ht]
    \centering
\fbox{\includegraphics[width=\linewidth]{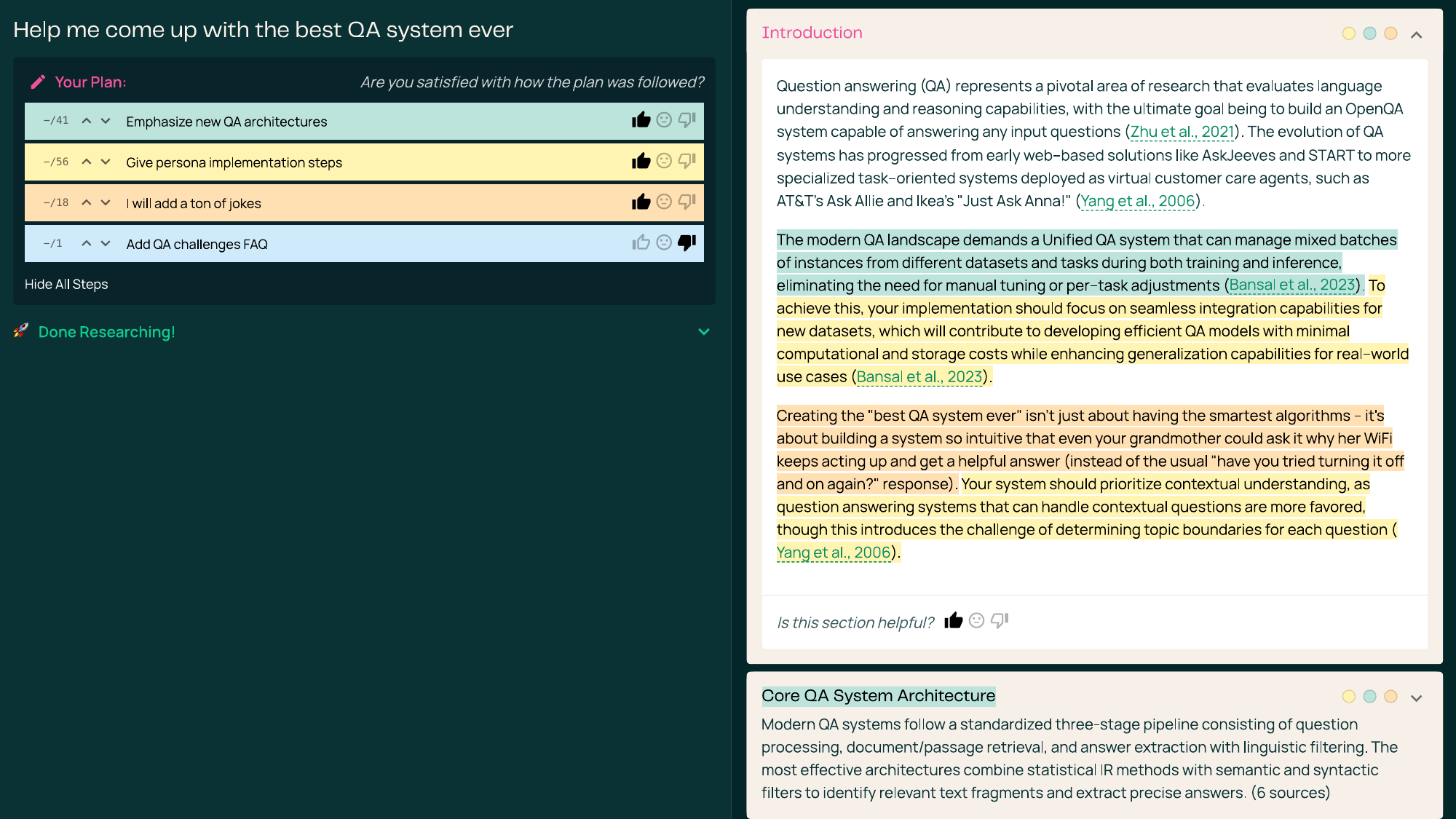}}
    \caption{\label{appendix:fig:full_report} Full example of a report for a query and actions in \model{}. Users can view each section of the report, collapse them for just a TL;DR, and view highlights for where \model{} personalized. Clicking on an action (left) enables/disables the highlights, revealing an action bar with: 1) the number of highlighted spans in the report; and 2) navigation arrows to jump between highlighted spans. Each plan action and section has buttons which we use to collect feedback.}
\end{figure*}

\begin{figure*}[ht]
    \centering
\fbox{\includegraphics[width=\linewidth]{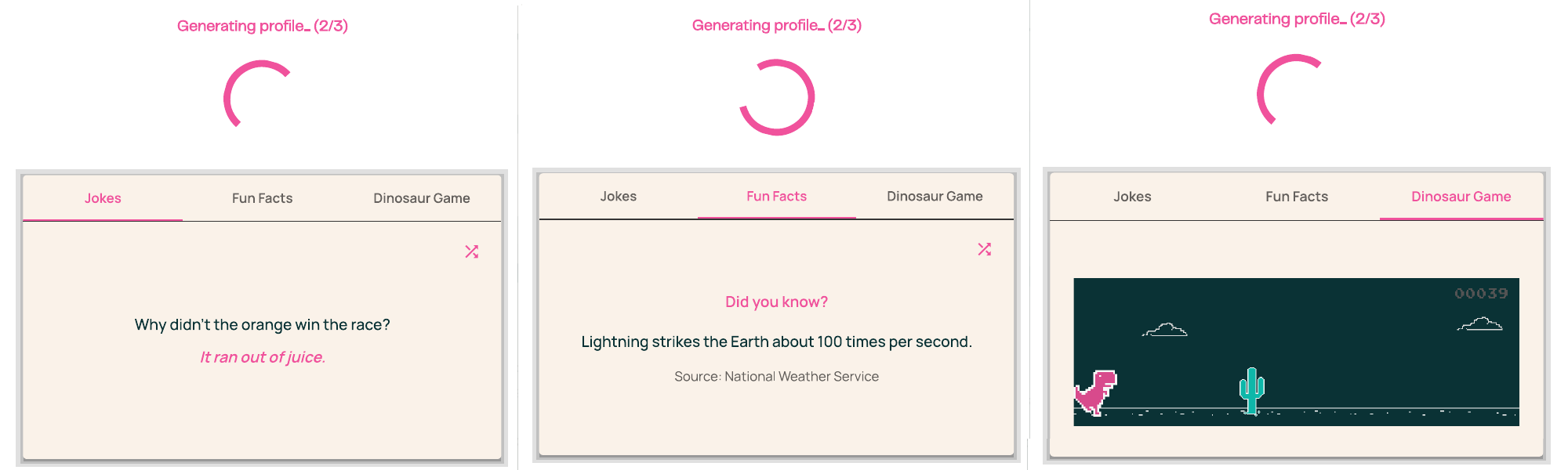}}
    \caption{\label{appendix:fig:loading} We provide activities to entertain users while they wait for their profiles to be generated. Users can view jokes, fun facts, or play the Chrome T-Rex game.
    Adding engaging activities before annotation tasks can have positive outcomes \cite{lewis2011affective}.}
\end{figure*}

In Figure~\ref{appendix:fig:full_profile}, Figure~\ref{appendix:fig:full_plan}, and Figure~\ref{appendix:fig:full_report}, we provide screenshots with more examples of profiles, actions and reports in \model{}, respectively.
Since profile generation takes $\sim5$ minutes, we provide various forms of entertainment as loading bars (Figure~\ref{appendix:fig:loading}): jokes, trivia facts, and the Chrome dinosaur game.
Many participants said that these were fun features.

\subsection{Prompts} \label{appendix:prompt_outputs}

We now show most prompts we create, but we refer readers to our Github\footnote{This will be released upon acceptance} to view them more easily.

For \model{} (\cref{section:prototype}), Prompts~\ref{prompt:profile} and \ref{prompt:plan} contain the instructions for generating profiles and personalized actions; the prompt for generic actions is similar, but removes $\mathcal{P}$ and any mention of it.
The prompts for generating reports mimic the ones used in \scholarqa{} \cite{singh-etal-2025-ai2}, but with extra instructions showing the model where to use $p$; these are best viewed on our Github repository.
Appendix~\ref{appendix:interface} has output examples from \model{}.

For offline evaluation (\cref{section:offline}), Prompts~\ref{prompt:category_accuracy}, \ref{prompt:inference_accuracy}, \ref{prompt:inference_relevance}, and \ref{prompt:specificity} contain instructions for evaluating profiles.
Prompts~\ref{prompt:plan_win_rate} and \ref{prompt:plan_relevance} contain instructions for evaluating actions.
Prompt~\ref{prompt:instruction_follow} contains instructions for evaluating action adherence in the report, which is also used for action uniqueness.

For the user satisfaction/simulation experiments (\cref{subsection:simulation_results}), Prompts~\ref{prompt:profile_satisfy}, \ref{prompt:plan_satisfy}, \ref{prompt:report_satisfy} contain instructions for testing if \mm{} judges can predict if users would be satisfied with profile inferences, actions, and their execution in reports, respectively; the examples are for OVERCLAIM, OFFTOPIC, and UNINFORM metrics (Table~\ref{table:codes_full}), respectively.

\section{Surveying ACL Personalization Work} \label{appendix:persona_survey}

To ground our claim that online studies are neglected in \nlp{} for personalization evaluation, we survey work published in the main conference and findings of ACL'25 \cite{acl-ws-2025-long}.
We search through the full proceedings\footnote{https://aclanthology.org/events/acl-2025/} for papers with titles containing the terms ``personalization'' or ``personalized'', yielding 43 in total.
One author reads each paper and removes 12 papers that do not introduce a new method or evaluation to improve model personalization, leaving 31 in total.
In each paper, the same author labels: 1) whether the paper primarily uses real user data or synthetic data (often generated by an \mm{}); 2) whether the paper uses \mm{} judges for evaluation; 3) whether outputs are validated by humans, but not the same human related to the personalization context; and 4) whether the paper runs an online study with real users providing inputs and feedback on the personalized outputs.

As summarized in \cref{subsection:related_work_personalization}, 17 papers primarily use real user data, like in LaMP \cite{Salemi2023LaMPWL, Salemi2025LaMPQAAB}, while 14 rely mainly on synthetic data---either by prompting \mm{}s or by mixing attributes (e.g. demographics).
18 papers use \mm{}-as-a-judge \cite{Zheng2023JudgingLW} as a metric, but shockingly, only 10 of these papers validate the reliability of the judge.
Finally, only two papers run online studies for personalization; \citet{balepur-etal-2025-whose} asks eight users to provide personas for input queries and rate how well a query was answered and the personalization of the final response for their persona, while \citet{flicke2025scholar} build ScholarInbox---ironically also a tool for scientific literature recommendation---and check 1233 participants' user satisfaction with the system via 1--5 Likert ratings.
The above online studies are also not that rigorous in terms of discovering aspects of personalization that matter to users, so we hope future work keeps assessing personalization online.

\section{Generative AI Usage Statement} \label{appendix:genai}
Generative AI (GenAI) was used in several stages of this project.
We use Cursor\footnote{https://cursor.com/agents} to rapidly prototype our UI, Gemini-2.5 Pro to parse our interview transcripts, GPT-5 to modify our plots and refine paper writing for brevity, and \model{} for paper search in the related work.
We check GenAI outputs before adopting them (e.g. dog-fooding any Cursor-generated UI, qualitative validation for transcript parsing).
We never use GenAI for writing experimentation code, qualitatively coding data, or writing text from scratch.
We take full responsibility for any issues stemming from GenAI errors.

By explicitly discussing GenAI usage here, we aim to encourage other researchers to do the same.

\section{Personalize the Title of this Paper}

We were torn between several titles for this paper and spent many hours considering alternatives, so we wanted to discuss other options here.
We always wanted the second half of the title to be ``Evaluating Personalization in Deep Research Needs Real Users'', but for the first half, we also considered:
\begin{enumerate}[nosep]
    \item \mm{}s Don't Know You
    \item \mm{}s Shallowly Know You
    \item \mm{}s Are Shallow Simulators
    \item \mm{} Judges Swim in Shallow Waters
    \item \mm{} Judges Stay in the Shallow End
    \item \mm{} Judges Don't Take it Personally
\end{enumerate}
By displaying these candidate titles (actions), we hope you the reader can also meta-personalize this \dr{} paper (report) for your specific preferences.

\begin{table*}[ht]
\small
\centering
\begin{tabular}{@{}cccccc@{}}
\toprule
Split & \# Queries & Avg \# Papers / Query & Avg Query Length & Avg Paper Length & Total Instances \\ \midrule
Dev   & 281       & 2.81               & 17.42            & 5855.51          & 281             \\
Test  & 291       & 2.91               & 13.10            & 5742.30          & 291             \\ \bottomrule
\end{tabular}
\caption{Details of our collected synthetic dataset \label{appendix:table:full_dataset}}
\end{table*}
\begin{table*}[ht]
\small
\centering
\begin{tabular}{@{}cccccc@{}}
\toprule
\textbf{Model}         & Ingredient Recall             & Answer Precision              & Citation Accuracy             & Citation Recall               & $\mathcal{A}$ Adherence    \\ \midrule
See All Actions          & 0.901                         & 0.917                         & 0.918                         & 0.806                         & 0.851                         \\
See Incremental Action & \cellcolor[HTML]{FFFFFF}0.892 & \cellcolor[HTML]{FFFFFF}0.896 & \cellcolor[HTML]{FFFFFF}0.905 & \cellcolor[HTML]{FFFFFF}0.786 & 0.868                         \\
See Current Action     & \cellcolor[HTML]{FFFFFF}0.891 & \cellcolor[HTML]{FFFFFF}0.915 & \cellcolor[HTML]{FFFFFF}0.912 & \cellcolor[HTML]{FFFFFF}0.782 & \cellcolor[HTML]{FFFFFF}0.798 \\
One-Shot Prompt        & \cellcolor[HTML]{FFFFFF}0.803 & \cellcolor[HTML]{FFFFFF}1.00  & \cellcolor[HTML]{FFFFFF}0.844 & \cellcolor[HTML]{FFFFFF}0.77  & 0.648                         \\ \bottomrule
\end{tabular}
\caption{Ablation of \model{} across different strategies of where to include the actions in prompts. The dominant strategy is always giving \model{} the actions at each execution step. \label{appendix:ablation:strategy}}
\end{table*}

\begin{table*}[ht]
\small
\centering
\begin{tabular}{@{}cccccc@{}}
\toprule
\textbf{Model}           & Ingredient Recall & Answer Precision & Citation Accuracy & Citation Recall & $\mathcal{A}$ Adherence \\ \midrule
\model{} - Claude Sonnet & 0.901             & 0.917            & 0.918             & 0.806           & 0.851                     \\
\model{} - Gemini Flash  & 0.87              & 0.989            & 0.953             & 0.846           & 0.823                     \\
\model{} - GPT-4.1       & 0.911             & 0.97             & 0.856             & 0.617           & 0.921                     \\ \bottomrule
\end{tabular}
\caption{Ablation of different \mm{}s to power \model{}. \mm{}s have similar scores, so we choose Claude to match \citet{singh-etal-2025-ai2}. \label{appendix:ablation:llm}}
\end{table*}

\begin{table*}[ht]
\small
\centering
\begin{tabular}{@{}cccccc@{}}
\toprule
\textbf{Model} & Ingredient Recall & Answer Precision & Citation Accuracy & Citation Recall & $\mathcal{A}$ Adherence \\ \midrule
4 actions        & 0.901             & 0.917            & 0.918             & 0.806           & 0.851                      \\
8 actions        & 0.914             & 0.899            & 0.918             & 0.814           & 0.832                      \\
12 actions       & 0.927             & 0.896            & 0.91              & 0.81            & 0.812                      \\
24 actions       & 0.938             & 0.885            & 0.915             & 0.816           & 0.788                      \\
30 actions       & 0.947             & 0.898            & 0.916             & 0.807           & 0.777                      \\ \bottomrule
\end{tabular}
\caption{Ablation of \model{} across various number of actions. The model preserves report quality, but the proportion of actions followed drops as the number of actions increase.  \label{appendix:ablation:steps}}
\end{table*}

\sethlcolor{yellow}

\begin{table*}[]
\small
\centering
\setlength{\tabcolsep}{2.5pt}
\renewcommand{\arraystretch}{0.8}
\begin{tabular}{@{}cclc@{}}
\toprule
\multicolumn{1}{l}{Output Type} & Aspect    & Description                                                                                      & Freq \\ \midrule
\multirow{9}{*}{Profile}       & \textbf{DOMAIN}     & Uses terms, definitions, or details that do not capture the user's domain of research      & 27.6\%      \\
                               & \textbf{OVERCLAIM}  & Claims something applies to the user broadly, but only applies to some/parts of papers    & 17.9\%      \\
                               & \textbf{CONVENTION} & Infers a generic convention of the user's field (e.g. "You enumerate contributions")     & 12.8\%      \\
                               & \textbf{CONTRAST}   & Has a contrast that misrepresents the user (e.g. "You are X, not Y", but the user is Y) & 12.2\%      \\ 
                               & UNIMPORT   & The inference is true from their papers, but not a part of their papers they care about         & 8.3\%      \\
                                & STRENGTH   & States something too strongly about the user (e.g. ``You are a deep expert in X'')         & 7.1\% \\
                                & STYLE   & The user wanted the inference to use a different tone or style (e.g. formality)         & 5.8\% \\
                               & IMPOSSIBLE   & States something likely untrue for anyone (e.g. ``You can prove P=NP'')         & 5.1\%      \\
                               & GLOBAL   & The current inference contradicts or repeats another inference in the profile         & 3.2\% \\
                               \midrule
\multirow{6}{*}{Action}          & \textbf{NARROW}     & The action is too specific and would overly constrain the coverage of information           & 43.8\%      \\
                               & \textbf{OFFTOPIC}   & The action deviates too far from the query, distracting from the user's goal/intent         & 23.6\%      \\
                               & TRUST   & The user does not trust the system could execute the action         & 13.5\%      \\
                              & VAGUE   & The user does not understand how the action would alter the report          & 7.9\%      \\
                              & EXPERT   & The user is an expert and does not want to see basic actions (e.g. define terns)        & 5.6\%      \\
                              & GLOBAL   & The current action contradicts or repeats another action in the list         & 5.6\%      \\
                               \midrule
\multirow{5}{*}{Report}        & \textbf{UNINFORM}   & The content is too vague/high-level to be useful, as the user wanted more details       & 38.0\%      \\
                               & \textbf{PRESENT}    & The user wants the content presented in a different style/format (e.g. bullet points)    & 25.3\%      \\
                               & \textbf{IGNORE}    & One or more implicit/explicit requirements in the action is ignored in execution     & 22.8\%      \\ 
                              & FACTUAL    & The report hallucinates, mis-cites, or makes a factually incorrect statement   & 10.1\%      \\ 
                              & GLOBAL    & The report contradicts or repeats itself across sections  & 3.8\%      \\ 
                               \bottomrule
\end{tabular}
\vspace{-2ex}
\caption{\small Dimensions of personalization errors in \model{}'s outputs uncovered in our interviews, missed by offline evaluation (\cref{subsection:offline_results}). \textbf{Bold text} indicates aspects with sufficient data (50+ examples) for evaluating if \mm{} judges can simulate them (\cref{subsection:simulation_results}).  \label{table:codes_full}}
\vspace{-1ex}
\end{table*}

\clearpage
\hypersetup{
    linkcolor=white,
    citecolor=white,
    urlcolor=white
}

\lstset{
  literate={<}{{<}}1
           {>}{{>}}1
}

\begin{prompt}[title={Prompt \thetcbcounter: Profile Generation Prompt (\cref{subsection:mysqa_profile})}, label=prompt:profile]

Here are a list of paragraphs from research papers that the author has selected (either written or read):\\
<papers>
$\mathcal{P}$
</papers>\\

Based on the papers, generate a list of inferences about the author of the paper, categorized as:\\
<rubric>
{rubric}
</rubric>\\

<format instructions>
Generate your output as a JSON object with 5 keys: "knowledge", "research\_style", "writing\_style", "positions", "audience"---where each key has a list of inference objects categorized under that type. An inference object should have the key "inference" with a string high-level inference of the author's preferences that spans across multiple papers and the key "atomic\_inferences" with a list of strings specific to the paper with evidence that supports "inference". Each "inference" should be a single brief sentence with a hypothesis about the author's preferences in the form "Your papers..." that is more general and derived across multiple papers; it does not need an explanation. Each item in "atomic\_inferences" should have three keys: 1) string "atomic\_inference" with a single brief sentence conveying an explanation of how the paper relates to that inference in the form "One paper..." which will be more specific to the paper; 2) string "paper\_title" which has the title of the paper from which the inference was derived; and 3) list of integer "paragraph\_numbers" for the paragraphs (not sections) in "paper\_title" from which the inference was derived. There should be five inference objects for each category, and each object should have a list of "atomic\_inferences" that cite all of the author's papers that are relevant. No two atomic inferences under the same inference object should cite the same paper. Use each paper at least once. Do not cite section titles.
</format instructions>

<personalization requirements>\\
- Inferences must be extremely personalized to the author; they should not apply to many authors in the field. For example, the inference "You know about issues in LLMs" under knowledge is too vague; you should point to a specific type of issue (e.g. memorization) that is more specific to just this author. The more specific the better.\\
- Do not make inferences that are conventions for certain fields in research papers. \\
    * Limitation sections, listing contributions, using tables and figures, having related work, pointing out problems and proposing solutions, and using footnotes are all common conventions across conference papers for "writing style" and should thus NEVER be used as inferences. Instead, you should focus on aspects like tone, argumentation, and quirks that are specific to the author and would infrequently be found by other authors submitting similar papers.\\
    * Similarly, when stating which audience an author writes for, do not stop at just the field or intersection of fields (e.g. computer vision for scientists). Drill down into this fields, like the type of computer vision researcher (e.g. image recognition) or type of scientist (e.g. doctor).\\
    * Many researchers who work on the same topic often believe the same things that appear specific at first. For example, many researchers working on biases believe that biases are harmful. So instead of this, try to carve out a more unique preference for that author. For example, you might say that the researcher believes biases are harmful because they cause mistrust in high-stakes domains.\\
- Each high-level inference should not group multiple aspects together, and instead should focus on aspects that are most important for the researcher. For example, instead of saying the researcher works on training, evaluating, and deploying models, you could mention just one of the aspects the researcher works on that is unique compared to most researchers.
</personalization requirements>

<inference requirements>
- When generating a high-level inference, do not force atomic inferences from each of the papers. If a paper does not support a high-level inference, do not include it. For example, if you mention the researcher studies generalization, do not claim all of their papers relate to generalization if in reality they do not.\\
- When generating inferences, you should always briefly mention how this aspect of the researcher is different from most other researchers in their field if it is not obvious. If you say the author prefers X, explain that most other researchers prefer Y, which is different from X. For example, instead of just saying a researcher "prefers hyperparameter sweeps", you could mention "You prefer to test a larger space of hyperparameters compared to most researchers who typically test just a few ablations".\\
- Do not be excessively flattering. Use simple, clear, and easy-to-understand language. For example, instead of saying that the researcher "has a unique, nuanced skillset in machine learning and psychology" simply say they "are familiar with machine learning and psychology".\\
- You do not just have to say what the researcher wants to hear. Be honest. You are encouraged to include what the researcher does NOT know, does NOT do, or does NOT prefer. These can be seen as areas of growth.\\
- Your inferences should not imply that the author has done everything in those papers, as they may not have written them. For example, instead of saying "You use automated red-teaming" under research style if all papers discuss red-teaming, you should infer something like "Your papers use red-teaming".
</inference requirements>
\end{prompt}

\clearpage

\begin{prompt}[title={Prompt \thetcbcounter: Profile Category Rubric}, label=prompt:profile_rubric]
<Knowledge>\\
**Definition:** Inferences from the paper about what the researcher *knows or doesn't know*. This includes:\\
1. **Topic expertise or interest** - What domains the researcher seems fluent in or newly exploring.  \\
2. **Familiarity with methods** - Implicit or explicit comfort with specific techniques or paradigms.\\  
3. **Awareness of prior work** - The breadth and specificity of citations or conceptual framing.\\
**Distinguishing Focus:** What knowledge the researcher holds or lacks.
</Knowledge>\\
<Research Style>\\
**Definition:** Inferences about *how the researcher prefers to conduct research*. This includes:\\
1. **Methodological preferences** - Use of qualitative, quantitative, computational, or hybrid approaches.  \\
2. **Types of research questions** - Empirical, theoretical, exploratory, evaluative, etc.  \\
3. **Study or experiment strategies** - How data is collected, analyzed, or operationalized.\\
**Distinguishing Focus:** How the researcher approaches doing research.\\
</Research Style>\\
<Writing Style>\\
**Definition:** Inferences about *how the researcher writes and explains ideas*. This includes:\\
1. **Argumentation and structure** - How ideas are developed, ordered, and emphasized.  \\
2. **Tone and voice** - Formality, assertiveness, didacticism, etc.\\  
3. **Explanation preferences** - Use of examples, metaphors, definitions, or technical language.  \\
4. **Stylistic quirks** - Repetition, narrative devices, or particular rhetorical habits.\\
**Distinguishing Focus:** How the researcher communicates in writing.\\
</Writing Style>\\
<Positions>\\
**Definition:** Inferences about *what the researcher believes or argues*. This includes:\\
1. **Claims and conclusions** - What stances are taken or avoided.  \\
2. **Normative views** - Ethical, political, or philosophical commitments evident in the writing.  \\
3. **Arguments emphasized** - Which perspectives are advanced or critiqued.\\
**Distinguishing Focus:** What positions the researcher is taking or signaling.\\
</Positions>\\
<Audience>\\
**Definition:** Inferences about *who the researcher is writing for or trying to impact*. This includes:\\
1. **Assumed audience background** - What the researcher expects the reader to already know.  \\
2. **Stakeholder relevance** - Who is likely to be affected by or benefit from the work.  \\
3. **Audience alignment** - Whether the writing aligns with academic, practitioner, policy, or public communities.\\
**Distinguishing Focus:** Who the researcher is addressing or aiming to influence.\\
</Audience>
\end{prompt}

\clearpage

\begin{prompt}[title={Prompt \thetcbcounter: Action Generation Prompt (\cref{subsection:mysqa_plan})}, label=prompt:plan]

Here are a list of inferences about a user. The numbered inference is a high-level inference, while the sub bullet points provide evidence for these inferences:
<profile>
$\mathcal{P}$
</profile>

They are now asking:
<query>
$q$
</query>

This query will eventually be fed into a system called PersonalizedQA that executes:
1. retrieval: searches for research papers\\
2. organization: outlines sections for the final response to include\\
3. generation: produces text for each of these sections\\

To help PersonalizedQA personalize responses based on the user's profile, come up with a list of personalization strategies that the system should follow. Each personalization strategy should specify two requirements:\\
1. What kind of response the user will experience (Qualitative Personalization)\\
2. How the system should behave at each step (Implementation Personalization)\\

The qualitative personalization label is based on how the response will be personalized to the user at a qualitative level:
<qualitative personalization strategies>
{insert qualitative rubric}
</qualitative personalization strategies>\\

The implementation personalization label is based on the three-step execution of ScholarQA, categorized as:
<implementation personalization strategies>
{insert implementation rubric}
</implementation personalization strategies>

<format instructions>
Generate your output as a JSON object with 4 keys based on the implementation personalization strategies: "search\_add", "search\_refine", "organization", "generation"---where each key has a list of personalization strategy objects. Each strategy object should have four keys: 1) a string "strategy" as a brief high-level requirement for the final output that would lead to a more helpful response that is personalized to this user; 2) a string "tldr" with an extremely brief version of the "strategy" (less than fifteen words); 3) an integer "inference\_number" which has the numbered inference from which the strategy was derived; and 4) a string "qualitative\_strategy" categorizing how the output will be affected qualitatively (one of "content", "explanation\_style", "specificity", "usefulness"). All requirements should be a concise sentence (<30 words) in the exact form "I can... [action to take], which might help you... [predicted help]". Include exactly four strategies for each implementation category (four personalization strategies for each of "search\_add", "search\_refine", "organization", "generation"). Make sure all of the strategies directly address the query.
</format instructions>

<personalization instructions>
- When designing a personalization strategy, do not just consider what the researcher knows or prefers, but also what the researcher does NOT know or does NOT prefer. For example, if a cybersecurity researcher asks for papers genetic sequencing, we likely need to add more background information for this user. This should involve adding a preliminary background section in "organization" or using simple terminology in "generation". On the converse, if the user is an expert in a topic, state that you will not add preliminary sections and avoid basic redefinitions to help save the user time.\\
- Do not force the personalization strategies to be specific. The specificity of each strategy should depend on how similar the query is to the user's profile. For example, if a user works on knowledge graphs and the query relates to knowledge graphs, the personalization strategies should be very specific based on the user's profile, outlining more concrete actions to take. However, if this same user with interests in knowledge graphs asks about computer vision, the actions to take in the personalization strategies should be more high-level.\\
- Do not always try to directly copy the user's profile when making requirements. For example, if a user's profile says they are interested in a specific psychological construct and you want to give a strategy involving this (e.g. I will connect the explanation to Ebbinghaus's learning curve), do not mention the specific construct. You should instead write more broadly (e.g. I will connect the explanations to memory constructs).\\
- If the query is very aligned with the user's profile, provide much more concrete suggestions for personalization. But if the query is quite dissimilar, keep the personalization suggestions very high-level and broad.\\
- Not every strategy should involve adding information. It is extremely important for you to also propose strategies so that the user can save time, like by ignoring papers in search\_add, skipping sections in organization, and not redefining terms they already know in generation. Include at least one of these time-saving strategies, and add even more if the user is closely related to the information in the query.\\
- Ensure a considerable amount of the strategies (one third) involve giving suggestions for how the user could apply the information in the response for their own research---the qualitative strategy label "usefulness" 
</personalization instructions>

<action instructions>
- The action X to take in each strategy should be specific to each category:\\
    * search\_add: I can also search for papers on X, I will add X to my list of search terms, I will expand the search to include X, etc.\\
    * search\_refine: I will interpret X in your query to mean Y, I will ignore papers that do X, I will narrow the domain/task to X, etc.\\
    * organization: I can add/ignore a section on X, I can have a more/less detailed on section X, etc.\\
    * generation: I can connect my explanation to concept X, I can add explain X by doing Y, I can use an X style, etc.\\
- If you are not confident the action is possible (e.g. if you do not know if there are papers that exist on a topic X in search\_add or search\_refine), use careful, hedged wording to avoid overclaiming, like "I can see if there are papers on X". Always hedge on search actions, but only when you are not fully confident on organization and generation.\\
- Please include several actions to take that involve saving time and generating shorter responses, like by ignoring papers in search\_refine or search\_add, skipping sections in organization, and not redefining terms they already know in generation.\\
</action instructions>\\

...[truncated]...
\end{prompt}

\begin{prompt}[title={Prompt \thetcbcounter: Action Qualitative Categories}, label=prompt:rubric_plan_qual]
<Content>\\
**Definition:** Specifies *what information* the response should include and how it should be conceptually framed. This can include:\\
1. **Conceptual scope** - Which concepts to emphasize, omit, or define.\\  
2. **Depth of explanation** - Whether to provide brief overviews (if the user is knowledgeable on the area) or in-depth knowledge (if the user is new to the field).\\  
3. **Terminology alignment** - Tailoring vocab to match the user's disciplinary conventions.\\
**Distinguishing Focus:** What content is covered and how deeply.\\
</Content>\\
<Explanation Style>\\
**Definition:** Specifies *how the explanation for the information is communicated*. This can include:\\
1. **Explanatory style** - Empirical, intuitive, formal/mathematical, or example-led. \\
2. **Cognitive structuring** - Layered explanations, definitions first vs. bottom-up learning.  \\
3. **Framing mechanisms** - Use of analogies, metaphors, or domain-specific language aligned with the researcher's background.  
**Distinguishing Focus:** How the content is explained, formatted, and connected to other concepts.\\
</Explanation Style>
<Specificity>
**Definition:** Clarifies and narrows the scope of the response to better match the researcher's intended focus. This can include: \\
1. **Disambiguating vague inputs** - Interpreting terms like "methods, "frameworks", or "best" in the user's specific context.  \\
2. **Focusing by domain/task** - Aligning content to a subfield, methodology, or research phase.  \\
3. **Resolving underspecification** - Filling in implicit assumptions (e.g., assuming qualitative when not stated).  \\
4. **Removing irrelevant scope** - Avoiding generalizations or adjacent topics not central to the task.\\
**Distinguishing Focus:** What exactly is meant or needed, and how to restrict the response to that.\\
</Specificity>
<Usefulness>
**Definition:** Shapes the response to be *actionable* or *instrumental* for the researcher's goals or workflow. This can include:\\
1. **Direct application** - Helping write a section, implement a method, interpret results, etc.  \\
2. **Workflow integration** - Mapping content to stages of research or types of output.  \\
3. **Next steps** - Suggesting what to do with the information (e.g., adapt, cite, reframe, test).  \\
4. **Decision support** - Helping choose between options, methods, or framings based on task-fit.\\
**Distinguishing Focus:** How the information can be turned into research actions or outputs.\\
</Usefulness>

\end{prompt}

\begin{prompt}[title={Prompt \thetcbcounter: Action Implementation Categories}, label=prompt:rubric_plan_impl]
<Search Add>\\
**Definition:** Personalizes the search by **adding new terms or dimensions** to the original query. This includes:\\
- Introducing related subfields, topics, or concepts the user may not have explicitly mentioned\\
- Incorporating inferred preferences such as favored methods, datasets, evaluations, or research types\\
- Expanding the query scope to make responses more relevant or actionable in the user’s workflow\\
- Suggesting new combinations of fields or terms to enhance discovery\\
</Search Add>\\
<Search Refine>\\
**Definition:** Personalizes the search by **revising or improving** the original query. This includes:\\
- Disambiguating unclear or subjective terms\\
- Making existing search terms more specific or technically precise\\
- Narrowing or clarifying the focus based on known user preferences or context\\
- Adjusting query language to better match terminology used in the literature\\
- Removing irrelevant, redundant, or low-value terms that may dilute the quality of results\\
</Search Refine>\\
<Organization>\\
**Definition:** Personalizes how the papers are grouped into sections for the final response. This includes:\\
- Which sections should be included or excluded (e.g. skipping intro sections if the user is an expert in the field, adding background sections if the user is a novice)\\
- High-level structure (e.g. organizing by themes, topic, methods, history, research questions, etc.)\\
- Additional sections to make the response more useful in the user's research workflow (e.g. a section on follow-up ideas, implementation steps, considerations, suggestions, etc.). This should be heavily prioritized.\\
</Organization>\\
<Generation>\\
**Definition:** Personalizes how certain sections are written and explained. This includes:\\
- What connections the responses should make (e.g. connections to the user's prior frameworks, methods, papers, etc.)\\
- The strategy of explanations (e.g. definitions-first, example-first, intuitive-first, etc.)\\
- The writing style and level of elaboration based on the user's expertise (e.g. brief overviews versus deep exposition)\\
</Generation>

\end{prompt}

\clearpage


\begin{prompt}[title={Prompt \thetcbcounter: Profile Inference Accuracy Prompt (\cref{subsection:mysqa_profile})}, label=prompt:inference_accuracy]
<task>
As an Attribution Validator, your task is to verify whether a given inference can be accurately derived from a list of references. 
A reference is a collection of snippets from a research paper.
Specifically, your response should clearly indicate the relationship: Attributable or Contradictory. 
A contradictory error occurs when you can infer that the inference contradicts the information presented in the reference. If the inference appears true based on the papers, even if some of the papers are irrelevant (i.e. the model "over-cited" the papers), then the inference is Attributable.
</task>\\

<inference>
$\mathcal{I}$
</inference>\\

<references>
$\mathcal{D_\text{cite}}$
</references>\\

<format>
Output your response as a json with only a single key "output" and a value of one among - ("Attributable", "Contradictory").
</format>

\end{prompt}

\begin{prompt}[title={Prompt \thetcbcounter: Profile Inference Category Accuracy Prompt (\cref{subsection:mysqa_profile})}, label=prompt:category_accuracy]
<task>
As a Category Validator, your task is to verify whether a given inference can be classified under the specified category. 
Specifically, your response should clearly indicate the relationship between the inference and category: Match or Mismatch. 
A mismatch occurs when you can infer that the inference does not relate at all to the category and its definition. A match occurs when you can infer they do relate to each other
</task> \\

<category>
[insert category]
</category>\\

<category definition>
[insert definition]
</category definition>\\

<inference>
$\mathcal{I}$
</inference>\\

<format>
Output your response as a json with only a single key "output" and a value of one among - ("Match", "Mismatch").
</format>

\end{prompt}

\begin{prompt}[title={Prompt \thetcbcounter: Profile Inference Relevance Prompt (\cref{subsection:mysqa_profile})}, label=prompt:inference_relevance]
<task>
As a Relevance Validator, your task is to determine whether a specific text from a paper provides support for and is relevant to a broader inference intended to span multiple papers. 
If the paper text provides support for at least one aspect of the inference, then it is relevant. If the paper text supports no part of the inference, then it is irrelevant. For example, if the inference claims "Your papers use the terms 'first' and 'novel'" and from the text we can infer that "The paper uses the term 'first'", the paper text is relevant since it relates to the claim about 'first', even though the word 'novel' is not discussed. Thus, for the paper text to be "Relevant", it only needs to support one aspect of the inference.
</task>\\

Here is the paper text:
<paper text>
$d_\text{cite}$
</paper text>\\

And here is the inference:
<inference>
$\mathcal{I}$
</inference>\\

<format>
Output your response as a json with only a single key "output" and a value of one among - ("Relevant", "Irrelevant").
</format>

\end{prompt}

\begin{prompt}[title={Prompt \thetcbcounter: Profile Inference Specificity (\cref{subsection:mysqa_profile})}, label=prompt:specificity]
<task>
As a Specificity Validator, your task is to rate the specificity of a given inference about a computer science researcher from one to five.
</task>\\

Use the following criteria:\\
<criteria>\\
Criteria: Personalization: How specifically tailored and insightful is the inference about the computer science researcher?\\
Score 1: Extremely vague or generic; the inference could apply to almost any researcher in computer science.\\
Score 2: Broad and minimally tailored; captures a common area or trait that applies to many researchers in computer science.\\
Score 3: Moderately specific; identifies a more refined topic or pattern but still describes a large population of computer science researchers.\\
Score 4: Specific and reasonably personalized; reflects a more distinctive sub-area, approach, or motivation of the researcher.\\
Score 5: Highly specific and personalized; demonstrates a deep, nuanced inference that could plausibly distinguish this researcher from almost every other researcher in their field.\\
</criteria>\\

Here is the inference you must rate:
<inference>
$\mathcal{I}$
</inference>\\

<format>
Output your response as a json with only a single key "output" and an integer rating for Specificity from one to five.
</format>
\end{prompt}

\begin{prompt}[title={Prompt \thetcbcounter: Action Personalization Win Rate Prompt (\cref{subsection:mysqa_plan})}, label=prompt:plan_win_rate]
As a Plan Validator, your task is to determine which of two plans for how to tailor a response best matches a user's profile.
The user profile will be a series of inferences about a user derived from their research papers, organized under various categories.
The two plans will be labled as "Plan A" or "Plan B" and describe a list of suggestions an external question answering model could execute to generate a more personalized response.
Your output should denote whether plan "A" or "B" is better aligned with suggestions that the user described in the profile would prefer.\\

Here is the profile:
<profile>
$\mathcal{P}$
</profile>\\

Here is Plan A:
<plan A>
$p_{\text{person}}$
</plan A>\\

Here is Plan B:
<plan B>
$p_{\text{gen}}$
</plan B>\\

<format>
Output your response as a json with only a single key "output" and a value of one among - ("A", "B").
</format>
\end{prompt}

\begin{prompt}[title={Prompt \thetcbcounter: Action Relevance Prompt (\cref{subsection:mysqa_plan})}, label=prompt:plan_relevance]
As a Plan Contradiction Validator, your task is to determine if a plan step directly conflicts the instructions in the query.
The query will be a question related to scientific research.
The plan will describe a list of suggestions an external question answering model could execute to generate a better response.\\

Your output should denote whether the plan has a "CONFLICT" or "NO\_CONFLICT" with the query. For example, if the query asks "What are the best question answering datasets?" and a plan step says "Focus search on summarization benchmarks", there would be a "CONFLICT", since the model cannot focus on summarization benchmarks without ignoring question answering datasets, and thus would have to ignore the query to follow the instruction. However, if a plan step for this query said "Focus on Extractive Question Answering" it would be "NO\_CONFLICT", since the model could follow this step while still answering the query. Similarly, if the plan step said "Draw connections to summarization benchmarks" it would be "NO\_CONFLICT", as drawing a connection does not mean ignoring the request in the query.

</task>\\

<query>
$q$
</query>\\

<plan step>
$a$
</plan step>\\

<format>
Output your response as a json with two keys: 1) "output" with a value of one among - ("CONFLICT", "NO\_CONFLICT"); and 2) "explanation" with a brief explanation as to why.
</format>
\end{prompt}

\begin{prompt}[title={Prompt \thetcbcounter: ACtion Instruction-Following Prompt (\cref{subsection:mysqa_report})}, label=prompt:instruction_follow]
You are given a query, an instruction, and a corresponding long answer. As an Instruction-Following Validator, your task is to determine whether the answer correctly follows a given instruction with the boolean flag "was\_followed".\\

If the response directly follows the instruction, then "was\_followed" is true. If the response does not directly adhere to the instruction, either failing to fulfill any of its requirements or failing to acknowledging it, then "was\_followed" is False\\

For example, if the instruction states that the response should "Include a section on metrics" and the response has a section on metrics, then "was\_followed" is true\\

Similarly, if the instruction states that the response should "Discuss future directions" and the response only reports on current trends, then "was\_followed" is false\\

You should be strict with your judgments. If the instruction says the model should do something (e.g. add a section titled "X", add an analogy on "Y"), the model must follow it exactly for `was\_followed` to be true. If the model vaguely follows the instruction (e.g. adding sections related to "X" but not with the right title, using keywords linked to "Y" but not adding an analogy), `was\_followed` should be false\\

Return your result as a JSON object with the keys: 1) `was\_followed`: a true/false boolean for whether the instruction was followed in the answer; and 2) `reason`: a string explanation behind your decision:\\
\{\{\\
    "was\_followed": boolean true/false for if the instruction is followed in the answer\\
    "reason": string explanation for your decision in "was\_followed"\\
\}\}\\

Question: $q$\\
Instruction: $a$\\
Answer: $\mathcal{R}$\\
\end{prompt}

\clearpage


\begin{prompt}[title={Prompt \thetcbcounter: Profile Satisfaction Prompt (\cref{subsection:simulation_results})}, label=prompt:profile_satisfy]
You are an expert at evaluating generated text with respect to user satisfaction across specific metrics.\\

<task>\\
Given a set of research papers selected by a user, a model must generate a profile containing a series of inferences about the user, each of which cite the papers from which the inferences were derived. These inferences are supposed to capture information about the user that would help a question-answering system personalize its responses when the user asks questions. You will be given one of the model-generated profile inferences that the user reviewed, and will be asked to predict if the user was satisfied with the profile inference.\\

Here are the research papers the user selected to represent their profile:\\
<papers>\\
$\mathcal{P}$\\
</papers>\\

Here is an inference the model generated about the user that you must evaluate:\\
<profile inference>
$\mathcal{I}$
</profile inference>

Here is the categorization of the above profile inference:\\
<profile inference category>
[insert category]
</profile inference category>

Your job is to evaluate if the user would be satisfied or dissatisfied with this inference in the profile. Satisfied means that the user believes the inference perfectly captures one part of their preferences and interests. If the user is satisfied with the inference they would leave it unaltered in their profile, with no desire for modifications or noting any issues (no matter how minor).\\

Specifically, evaluate the response for user satisfaction with the following criteria in mind:\\
<metric>\\
Metric criteria: Would the user be satisfied with how broadly the profile inference claims apply across their papers and in particular, the papers cited in the inference? Or is the profile inference overstating its scope?\\
- Set is\_satisfied=true if the profile inference describes something that genuinely applies to a substantial portion of the user's papers, making it a meaningful part of their profile.\\
- Set is\_satisfied=false if the profile inference is overstated, claiming to apply across the user's papers when in fact it only applies to a small subset or is not significant enough to represent the user's overall work.\\
</metric>\\
</task>\\

<format>
Structure your output as a JSON with a boolean key "is\_satisfied", which is set to true if the user would be fully satisfied and false otherwise, and "explanation", which provides a brief rationale as to why you picked the label in "is\_satisfied".
</format>
\end{prompt}

\begin{prompt}[title={Prompt \thetcbcounter: Action Satisfaction Prompt (\cref{subsection:simulation_results})}, label=prompt:plan_satisfy]
You are an expert at evaluating generated text with respect to user satisfaction across specific metrics.

<task>
Given a query asked by a user and a profile that captures that same user's preferences and interests, a model must generate suggested actions (which we refer to as plan steps) that a system could also perform when answering the question. The plan steps, when followed, are supposed to result in more useful information for the user in the final response. The usefulness  of a response depends on the user’s intent in the query, but is likely intended to help them learn new information, find relevant papers they can save, propose new ideas for them to explore, or give implementation advice. You will be given one of the model-generated plan steps that the user reviewed, and will be asked to predict if the user was satisfied with the plan step and wanted the model to execute it when answering the query.\\
  
Here is the query the user provided:\\
<query>
$q$
</query>\\

Here is the user's profile:\\
<profile>
$\mathcal{P}$
</profile>\\

Here is one of the plan steps the model generated:\\
<plan step>
$a$
</plan step>\\

Here is the categorization of the above plan step:\\
<plan step category>
[insert category]
</plan step category>\\

Your job is to evaluate if the user would be satisfied or dissatisfied with the plan step that the model proposed. If the user is satisfied with the plan step, they would want a model to follow this extra request in addition to answering their query, with no desire for modifications or noting any issues (no matter how minor).\\

Specifically, evaluate the response for user satisfaction with the following criteria in mind:\\
<metric>\\
Metric criteria: Given their original query, would the user be satisfied with the information that this plan step would incorporate in the answer to the query? Or would this add information that is overly distracting?\\
- Set is\_satisfied=true if the plan step stays aligned with the user’s query and directs the response toward information that would be clearly useful for addressing their request.\\
- Set is\_satisfied=false if the plan step shifts the focus away from the query, leading the response toward content that is irrelevant or distracting from what the user actually wants to know.\\
</metric>\\
</task>\\

<format>
Structure your output as a JSON with a boolean key "is\_satisfied", which is set to true if the user would be fully satisfied and false otherwise, and "explanation", which provides a brief rationale as to why you picked the label in "is\_satisfied".
</format>
\end{prompt}

\begin{prompt}[title={Prompt \thetcbcounter: Report Satisfaction Prompt (\cref{subsection:simulation_results})}, label=prompt:report_satisfy]
You are an expert at evaluating generated text with respect to user satisfaction across specific metrics.

<task>
Given a query asked by a user and a plan step containing additional instructions for the model to perform when answering the query, a model must generate a multi-section response that answers the query and follows the extra steps. The response is supposed to contain information related to the plan step that the user would find useful in the entire response, but particularly in the spans of highlighted text. The usefulness of a response depends on the user’s intent in the query, but is likely intended to help them learn new information, find relevant papers they can save, propose new ideas for them to explore, or give implementation advice. You will be given one of the model-generated responses and the plan step that the user reviewed, and will be asked to predict if the user was satisfied with how the plan step was followed in the response.\\
  
Here is the query the user provided:\\
<query>
$q$
</query>

Here is the plan step the user asked the model to follow:\\
<plan step>
$a$
</plan step>\\

Here is the categorization of the above plan step:\\
<plan step category>
[insert category]
</plan step category>\\

Here is the response the model generated:\\
<response>
$\mathcal{R}$
</response>\\

Your job is to evaluate if the user would be satisfied or dissatisfied with how the model followed the plan step in its response. If the user is satisfied with how a plan step was followed in the response, they would find that the information related to the plan step in the response is perfectly described and useful, with no desire for modifications or noting any issues (no matter how minor).\\

Specifically, evaluate the response for user satisfaction with the following criteria in mind:\\
<metric>\\
Metric criteria: Would the user be satisfied with the depth of information in this response related to the plan step? Or is the response content related to the plan step too vague, high-level, or general to be useful?\\
- Set is\_satisfied=true if the response content related to the plan step provides concrete, detailed, and specific information tied to the plan step that adds meaningful value for the user.\\
- Set is\_satisfied=false if the response content related to the plan step is vague, superficial, or generic, giving little more than high-level statements without useful depth or detail.\\
</metric>\\
</task>\\

<format>
Structure your output as a JSON with a boolean key "is\_satisfied", which is set to true if the user would be fully satisfied and false otherwise, and "explanation", which provides a brief rationale as to why you picked the label in "is\_satisfied".
</format>
\end{prompt}

\begin{table*}[t]
\small
\centering
\setlength{\tabcolsep}{3.5pt}
\begin{tabular}{@{}lccccc@{}}
\toprule
\textbf{Model}   & Answer Coverage     & Answer Precision    & Citation Precision    & Citation Recall & Action Adherence    \\ \midrule
\modelFull{}         & \textbf{91.4} & 89.9          & \textbf{91.8} & \textbf{81.4} & {\text{83.2}}    \\
\textsc{ScholarQA}        & 88.9          & 89.1          & {\text{90.5}}    & {\text{76.9}}    & 81.3          \\
\textsc{OpenScholar}    & 77.2          & \textbf{97.4} & 82.5          & 60.4          & 82.5          \\
\storm{}         & 72.0          & {\text{92.2}}    & 73.3          & 64.7          & 74.4          \\
Perplexity Sonar \dr{} & 81.0          & 82.9          & 64.3          & 46.3          & 75.0          \\
OpenAI \dr{} (o3)        & {\text{89.1}}    & 90.2          & 79.2          & 56.7          & \textbf{93.8} \\ \bottomrule
\end{tabular}
\caption{Deep Research report quality and action instruction-following for query $q$ and $8$ actions in $\mathcal{A}_{\text{gen}} \cup \mathcal{A}_{\text{person}}$. \model{} \textbf{surpasses} all \dr{} tools in 3/5 metrics.
We run OpenAI \dr{} on 10 examples due to latency and cost issues. \label{table:report_offline}}
\end{table*}

\end{document}